\documentclass[11pt]{article}

\usepackage[]{acl}

\usepackage{times}
\usepackage{latexsym}

\usepackage[T1]{fontenc}

\usepackage[utf8]{inputenc}

\usepackage{microtype}

\usepackage{inconsolata}

\usepackage{graphicx}

\usepackage{graphicx}               
\usepackage{tabularx}               
\usepackage{booktabs}
\usepackage{amsmath}
\usepackage{stmaryrd}
\usepackage{xcolor}
\usepackage{soul}

\newcolumntype{C}{>{\centering\arraybackslash}X}
\usepackage{multirow}               
\usepackage{diagbox}                
\usepackage{hhline}                 
\usepackage{color}                  
\usepackage{amsmath}                
\usepackage{amssymb}                
\usepackage{mathtools}              
\usepackage{enumitem}               
\usepackage[most]{tcolorbox}

\usepackage{booktabs}
\usepackage{extarrows}
\usepackage{makecell}
\usepackage{subcaption}
\usepackage{soul}

\definecolor{RoseQuartzBg}{HTML}{F7CAC9}
\definecolor{RoseQuartz}{HTML}{F5A798}
\definecolor{Serenity}{HTML}{92A8D1}
\definecolor{OrangeRed}{rgb}{1.0, 0.27, 0.0}
\definecolor{Red}{rgb}{1.0, 0.0, 0.0}
\definecolor{Turquoise}{HTML}{0F4C81}
\definecolor{RoyalBlue}{cmyk}{1, 0.50, 0, 0}
\definecolor{Green}{rgb}{0.0, 1.0, 0.0}
\definecolor{Orchid}{rgb}{0.85, 0.44, 0.84}
\definecolor{Orange}{rgb}{1.0, 0.5, 0.0}
\usepackage{xparse}
\NewDocumentCommand{\lifu}{ mO{} }{\textcolor{Red}{\textsuperscript{\textit{Lifu}}\textsf{\textbf{\small[#1]}}}}
\NewDocumentCommand{\barry}{ mO{} }{\textcolor{blue}{\textsuperscript{\textit{Barry}}\textsf{\textbf{\small[#1]}}}}
\NewDocumentCommand{\hugo}{ mO{} }{\textcolor{Serenity}{\textsuperscript{\textit{Hugo}}\textsf{\textbf{\small[#1]}}}}
\NewDocumentCommand{\qf}{ mO{} }{\textcolor{RoseQuartzBg}{\textsuperscript{\textit{Qifan}}\textsf{\textbf{\small[#1]}}}}
\NewDocumentCommand{\sijia}{ mO{} }{\textcolor{RoyalBlue}{\textsuperscript{\textit{sijia}}\textsf{\textbf{\small[#1]}}}}
\NewDocumentCommand{\minqian}{ mO{} }{\textcolor{teal}{\textsuperscript{\textit{Minqian}}\textsf{\textbf{\small[#1]}}}}
\NewDocumentCommand{\zhiyang}{ mO{} }{\textcolor{OrangeRed}{\textsuperscript{\textit{Zhiyang}}\textsf{\textbf{\small[#1]}}}}

\usepackage{amssymb}
\usepackage{pifont}

 \setlist[itemize]{leftmargin=*}
\setlist[enumerate]{leftmargin=*}

\newpage
\usepackage{appendix}  
\usepackage{diagbox}                

\usepackage{natbib} 
\usepackage{algorithm}
\usepackage{algpseudocode}

\usepackage{amsthm}

\newtheorem{definition}{Definition}

\usepackage{enumitem}
\newtheorem{finding}{Finding}

\newcommand{\dataset}{\textsc{Fico}}
\newtcolorbox{findingbox}{
  colback=gray!10,
  colframe=black!25,
  boxrule=0.9pt,
  arc=3pt,
  left=2pt,
  right=2pt,
  top=2pt,
  bottom=2pt
}
\usepackage{amssymb}

\usepackage{wrapfig}
\let\cite\citep
\vspace{-2em}
\title{How Do Large Language Models Learn Concepts \\ During Continual Pre-Training?}

\author{
  Barry Menglong Yao$^{\spadesuit}$ \quad
  Sha Li$^{\heartsuit}$ \quad
  Yunzhi Yao$^{\clubsuit}$ \quad
  Minqian Liu$^{\heartsuit}$ \\
  \textbf{Zaishuo Xia}$^{\spadesuit}$ \quad
  \textbf{Qifan Wang}$^{\diamondsuit}$ \quad
  \textbf{Lifu Huang}$^{\spadesuit}$ \\
  $^{\spadesuit}$UC Davis \quad
  $^{\heartsuit}$Virginia Tech \quad
  $^{\clubsuit}$UCLA \quad
  $^{\diamondsuit}$Meta AI \\
  \texttt{\{bmyao,zsxia,lfuhuang\}@ucdavis.edu} \\
  \texttt{\{shal,minqianliu\}@vt.edu} \\
  \texttt{\{wqfcr\}@meta.com}
}

\begin{document}
\maketitle

\vspace{-3em}
\begin{abstract}

Human beings primarily understand the world through concepts (e.g., \textit{dog}), abstract mental representations that structure perception, reasoning, and learning. However, how large language models (LLMs) acquire, retain, and forget such concepts during continual pretraining remains poorly understood. In this work, we study how individual concepts are acquired and forgotten, as well as how multiple concepts interact through interference and synergy. We link these behavioral dynamics to LLMs' internal \textbf{concept circuits}, computational subgraphs associated with specific concepts, and incorporate graph metrics to characterize \textbf{circuit topology}. Our analysis reveals: (1) LLMs' concept circuits provide a non-trivial, consistent signal of concept learning and forgetting; (2) concept circuit topology evolves stage-wise, rising during concept acquisition, and transiently rising before declining during concept forgetting; (3) semantically similar concepts induce stronger interference than weakly related ones; (4) conceptual knowledge differs in their transferability, with some significantly facilitating the learning of others. Together, our findings provide a circuit-level view of concept learning dynamics and motivate concept-aware training strategies, such as Circuit-aware Experience Replay, which uses circuit topology to prioritize concepts vulnerable to forgetting\footnote{The code is publicly available at \url{https://github.com/PLUM-Lab/llm_circuit_for_learning_dynamics}.}.
\end{abstract}
\vspace{-1pt}

\vspace{-1pt}
\section{Introduction}

\begin{figure*}[t]
\begin{center}
    \includegraphics[width=\textwidth]{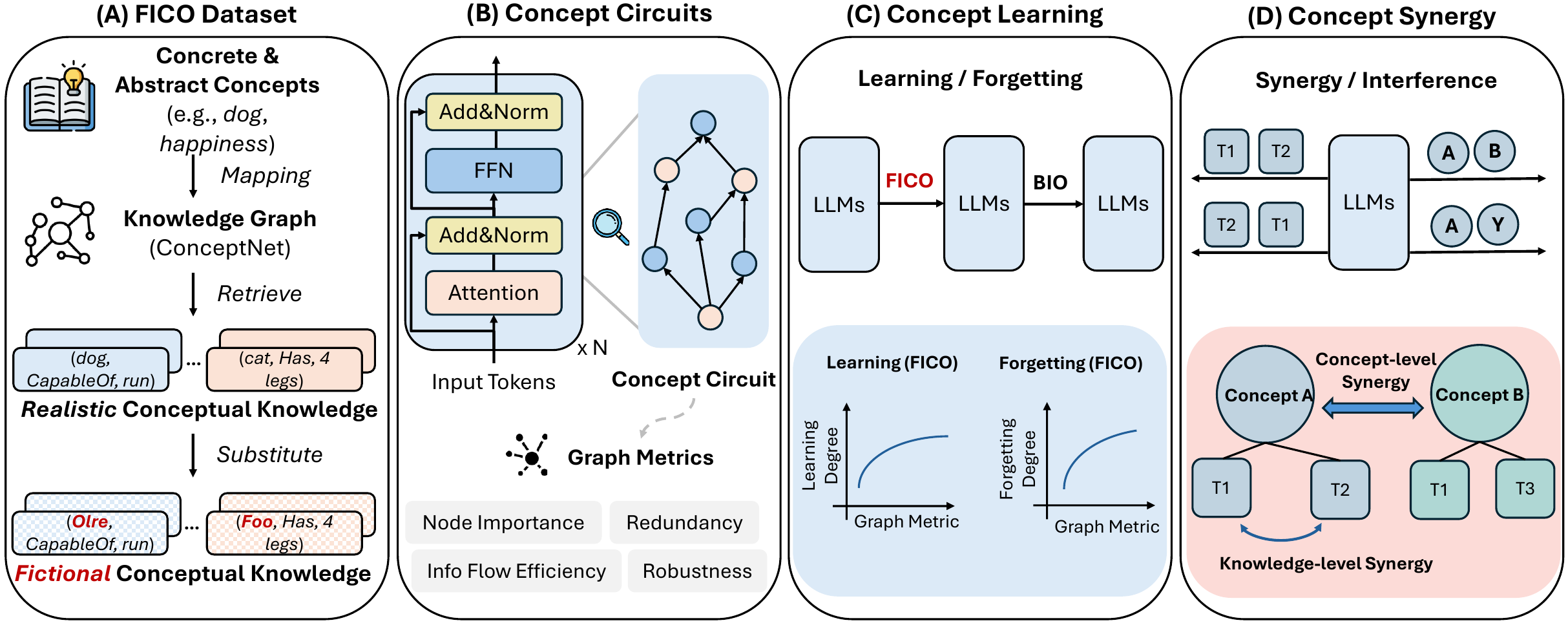}
    \caption{ (A) Construct the \textsc{Fico} dataset based on ConceptNet. (B) Extract \textbf{Concept Circuits}, LLM computational subgraph associated with individual concepts, and characterize their structure using graph metrics. (C) Analyze \textbf{concept learning and forgetting dynamics} in two-stage continual pretraining. (D) Study \textbf{synergy and interference} across concepts (e.g., $A$, $B$, $Y$) and knowledge type (e.g., $T1$ and $T2$). }
    \label{fig:overall}
\end{center}
\end{figure*}

Humans organize perception and reasoning around \textbf{concepts}, abstract mental categories (e.g., \textit{dog}) that enable generalization from individual observations to shared properties, relations, and actions~\cite{harpaintner2018semantic}. Similarly, a defining characteristic of large language models (LLMs) is their ability to abstract conceptual knowledge from vast corpora, encoding it within internal \emph{concept-level} representations that facilitate downstream reasoning and generation\cite{aljaafari2024mechanics,wang-etal-2024-editing,park2023linear}. As new concepts continually emerge and models are updated through continual pre-training, it becomes essential to understand how reliably new concepts are acquired and prior concepts are retained, and how to make concept learning more efficient in the presence of interactions among the many concepts and types of conceptual knowledge.

Prior studies have examined concept learning in LLMs from two angles. One line of work uses prompt-based \textbf{knowledge probing} to test whether specific conceptual properties or relations (e.g., \textit{commonsense facts}) can be elicited from a pretrained model~\cite{gu-etal-2023-language,liao2023concept,shani-etal-2023-towards,zheng2024concept,peng-etal-2022-copen,xu2024tip}, while another leverages \textbf{mechanistic interpretability} tools to localize internal structures associated with conceptual information~\cite{aljaafari2024mechanics,wang-etal-2024-editing}. However, these efforts provide only a partial view of concept learning: they typically probe isolated pieces of conceptual knowledge that models may already possess, rather than systematically characterizing how concepts are acquired, consolidated, and forgotten during continual pre-training.\looseness=-1

In this work, we take a step toward filling this gap by asking two more research questions: (1) \textit{how internal concept representations correlate with concept acquisition and forgetting}, and (2) \textit{how they relate to interference and synergy among multiple concepts and types of knowledge during joint training}. Answering these questions matters not only for interpretability, but also for practice: they can inform concept-aware continual pre-training decisions, such as how much training to allocate to previously learned concepts to mitigate forgetting and to newly introduced concepts to facilitate efficient acquisition, how to schedule and reorder training data, and how to mitigate destructive interference among semantically related concepts.\looseness=-1


To enable a controlled study, we introduce the \dataset{} (\textbf{FI}ctional \textbf{CO}ncept) dataset, built from ConceptNet-derived conceptual relations but mapped to synthetic, non-existent concept names, preserving realistic knowledge structure while making the concepts novel to the model. We then conduct controlled continual pre-training experiments and link behavioral changes to internal mechanisms by extracting \textbf{Concept Circuits}, computational subgraphs associated with each concept, using circuit identification methods and characterizing their topology over time with graph metrics. This joint analysis allows us to trace how concept-level behavior (e.g., learning and forgetting measures) co-evolves with circuit topology, and how cross-concept relatedness and knowledge-type ordering shape interference and transfer (Figure~\ref{fig:overall}).


Our results reveal several meaningful and systematic patterns in concept learning dynamics in LLMs: \textbf{(1)} circuit topology is non-trivially correlated with both concept learning and forgetting, suggesting a trade-off between acquisition strength and subsequent stability: more centralized and redundant circuit structures are associated with stronger acquisition, but also with greater forgetting under continual training. This trade-off is also reflected behaviorally, as concepts with larger learning gains tend to exhibit greater forgetting during subsequent training. \textbf{(2)} Concept circuits follow stable stage-wise temporal trajectories across multiple graph metrics in both training stages: their topology rises and then stabilizes as concepts are acquired, whereas during continued training on new domains it transiently rises before gradually declining and stabilizing, suggesting distinct phases of concept consolidation and subsequent reorganization. \textbf{(3)} At the level of multi-concept learning, semantically similar concepts interfere more strongly than weakly related concepts, and different types of conceptual knowledge differ substantially in their transferability; e.g., pretraining on \textit{Hyponym \& Hypernym} knowledge improves subsequent learning performance on \textit{Synonym \& Antonym} knowledge by 63.74\%. Together, these findings offer a circuit-level view of concept learning dynamics and show that the resulting circuit-based signals can be translated into actionable training strategies, such as circuit-aware replay, to improve continual pre-training stability.

Our contributions are summarized as follows:
\vspace{-3pt}
\begin{itemize}
    \item We present the first systematic study of concept acquisition and forgetting in LLMs during continual pretraining, revealing consistent correlations between concept circuits, learning dynamics, and forgetting behavior, suggesting these signals as indicators of future learning and forgetting.
    \item We present a relational perspective on knowledge learning in LLMs, revealing interference and synergy patterns across concepts and knowledge types during joint training, motivating interference-aware data scheduling.
    \item Our analysis informs actionable concept-aware training strategies for continual pre-training. As a proof of value, we instantiate a circuit-aware replay strategy to improve knowledge retention.
    
\end{itemize}

\vspace{-4pt}
\section{Dataset Construction}~\label{sec:fictional_concept}

\vspace{-0.7cm}
We follow previous studies~\cite{wang-etal-2024-editing} and define a \textbf{concept} (e.g., \textit{dog}) as an abstraction over the world that captures the shared features and essential characteristics of a class of entities. \textbf{Conceptual knowledge} (e.g., \textit{a dog has four legs}) refers to factual or relational information associated with a concept, reflecting familiarity with and understanding of its properties and relations~\cite{wang-etal-2024-knowledge-mechanisms}. Each concept is therefore linked to a set of conceptual knowledge that collectively describe its semantic content. To study concept learning across different levels of concreteness, we consider both \textbf{concrete concepts} that are grounded in direct sensory experience (e.g., \textit{dog}), and \textbf{abstract concepts} which are not directly perceptible (e.g., \textit{love}). We sample 500 concrete concepts from the THINGS dataset~\cite{hebart2023things} and 500 abstract concepts from the Concreteness Ratings dataset~\cite{brysbaert2014concreteness}. For each concept, we retrieve associated conceptual knowledge from ConceptNet~\cite{speer2017conceptnet}, a large-scale knowledge graph of millions of concepts and 34 typed relations. 

To mitigate influence of pre-existing knowledge encoded in LLMs, we replace each real concept name with a fictional name generated by GPT-5~\cite{hurst2024gpt} and perform relation filtering and canonicalization, as detailed in Appendix~\ref{sec:knowledge_filtering}. This process forms our \dataset{}, \textbf{FI}ctional \textbf{CO}ncept dataset, illustrated in Figure~\ref{fig:overall}(A), with statistics in Appendix~\ref{sec:statistics}. Following prior work~\cite{how2025dynamics}, we use GPT-5 to generate multiple natural-language templates for each relation type (e.g., ``\textit{\{concept\} has the ability to}'' for \textit{CapableOf}), converting each knowledge triple (e.g., $(\text{dog}, \text{CapableOf}, \text{run})$) into prefix-target examples. For evaluation, we sample 500 concepts and use a \emph{disjoint} template pool to construct test instances with unseen surface forms, enabling assessment of whether LLMs learn underlying relations rather than memorizing specific templates.

\vspace{-1pt}

\section{How Internal Concept Representations Correlate with LLM Learning and Forgetting Dynamics?}

\subsection{Concept Circuit as Internal Concept Representation}   
Given a knowledge triple $k_{ij}=(c_i, r_{ij}, o_{ij})$, where $c_i$ is a subject concept (e.g., \textit{dog}), $r_{ij}$ is a relation type (e.g., \textit{HasProperty}), and $o_{ij}$ is an object which can be another concept or a descriptive phrase (e.g., \textit{fleas} or \textit{four legs}), a \textbf{Knowledge Circuit} is defined as the minimal computational subgraph in a pretrained LLM that can faithfully predict the target object $o_{ij}$ conditioned on a textual prefix converted from the subject--relation pair $(c_i,r_{ij})$~\cite{yao2025knowledgecircuitspretrainedtransformers}. We adapt this notion to concept-level analysis and define a \textbf{Concept Circuit} for concept $c_i$ as the computational subgraph that jointly and faithfully predict all conceptual knowledge $\mathcal{K}_i=\{k_{i0},k_{i1},\ldots,k_{ij},\ldots\}$ associated with $c_i$. Unlike a union of independently extracted circuits for individual facts, a Concept Circuit is identified at the level of $\mathcal{K}_i$ and captures the shared computation underlying the conceptual knowledge.



To extract concept circuits at different checkpoints during continual training, we use EAP-IG~\cite{hanna2024have} as our circuit identification method. EAP-IG assigns an importance score to each edge while balancing computational efficiency with attribution faithfulness\footnote{More details for the implementation of EAP-IG can be found in Appendix~\ref{sec:circuit_discovery}.}. Given the edge importance scores, we construct a circuit by selecting the top-scoring edges such that the resulting subgraph preserves at least 70\%\footnote{Following prior circuit-level interpretability work~\cite{ou-etal-2025-llms,yao2025knowledgecircuitspretrainedtransformers}, we use 70\% as a heuristic preservation threshold and provide an 80\% ablation in Section~\ref{sec:rq1_finding}. Thresholds $\geq 90\%$ often yield overly large, less interpretable circuits.} of the full model's performance on the corresponding concept.

\paragraph{Topology of Circuit Graph}
Previous work has shown that the geometric and topological properties of LLM circuits and features can provide insight into model behaviors such as grokking, reasoning, and learning. \cite{humayun2024grokking,zhou2025geometry,raj2026hyperbolic,li2025geometry,ou-etal-2025-llms}. Motivated by these findings, we compute four families of standard graph metrics to characterize the topology of concept circuits and their relationship to concept learning dynamics in LLMs: \textbf{(1) Density}, defined as the ratio of existing edges to the maximum possible number of edges. Higher density reflects more redundant connections~\cite{humayun2024grokking,newman2010networks}. \textbf{(2) Eigenvector Centrality}, whose standard deviation quantifies how unevenly structural influence is distributed across nodes. Higher variance indicates a more concentrated hub structure~\cite{newman2010networks}. \textbf{(3) Global Efficiency}, i.e., the average inverse shortest-path distance between node pairs. Higher global efficiency indicates that signals can propagate more efficiently across the graph~\cite{latora2001efficient}.
\textbf{(4) k-core}, whose average number captures the depth of a graph's densely connected core and is used as a proxy for resilience to disruption in graph~\cite{laishram2018measuring,seidman1983network}. We select these four metrics because, in the graph analysis literature, they are standard measures of graph redundancy, node importance, information flow efficiency, and graph robustness, respectively.



\subsection{Experiment Design }~\label{sec:experiment_design}

\vspace{-3pt}


To quantify how internal concept representations (i.e., concept circuits) relate to concept acquisition and forgetting, given a pre-trained LLM $\pi_0$, we conduct a two-stage~\footnote{We further evaluate longer continual updates and observe consistent trends; see Appendix~\ref{sec:longer_continual_training}.} continual pretraining setup. In \textbf{Stage 1 (Concept Acquisition)}, we continually train on the \dataset{} training set to learn novel concepts, yielding $\pi_1$, which is used to analyze concept learning dynamics. In \textbf{Stage 2 (Forgetting Induction)}, we further train $\pi_1$ on the BIO dataset~\cite{allen2023physics}, a standard pretraining dataset used in previous LLMs knowledge acquisition studies~\cite{allen2023physics,how2025dynamics}, yielding a new model $\pi_2$, which is used to analyze concept forgetting dynamics. We experiment with GPT-2 Large~\cite{radford2019language}, LLaMA-3.2-1B, LLaMA-3.1-8B~\cite{dubey2024llama}, and Qwen3-4B-Base~\cite{yang2025qwen3}.\footnote{We observe similar trends across models. Due to space constraints, we present GPT-2 results in the main paper and defer LLaMA and Qwen results to Appendix~\ref{sec:llama}.} In both stages, we concatenate the prefix and target phrase, as described in Section~\ref{sec:fictional_concept}, and optimize the model using next-token prediction. 

For evaluation, we provide only the prefix on the \dataset{} test set and compute concept learning/forgetting degrees from the model's gold-target logit, as defined below. We use \textbf{Spearman's rank correlation coefficient} ($\rho$)~\cite{spearman1961proof} to analyze the correlation between concept learning/forgetting dynamics and concept circuit topology.\footnote{See Appendix~\ref{sec:experiment_setting} for additional experiment details.}

\begin{definition}[Knowledge Learning/Forgetting Degree]
For a knowledge triple $k=(c,r,o)$ with prefix $x=T(c,r)$, let $s_{\pi}(k)=\mathrm{logit}_{\pi}(o\mid x)$ denote the gold-target logit.\footnote{Following prior work~\cite{heimersheim2024use,hanna2024have}, we use logits as a fine-grained indicator of LLM learning and report log-probability results in Appendix~\ref{sec:log_probability}.} The \textbf{\emph{knowledge learning degree}} and \textbf{\emph{knowledge forgetting degree}} of $k$ are defined as $\Delta_{\mathrm{learn}}(k)=s_{\pi_1}(k)-s_{\pi_0}(k)$ and $\Delta_{\mathrm{forget}}(k)=s_{\pi_1}(k)-s_{\pi_2}(k)$, respectively.
\end{definition}
\vspace{-3pt}

\begin{definition}[Concept Learning/Forgetting Degree]\label{def:learned_degree}
For a concept $c$ with associated triples $\mathcal{K}_c$, its \textbf{\emph{concept learning degree}} and \textbf{\emph{concept forgetting degree}} are defined as $\Delta_{\mathrm{learn}}(c)=\frac{1}{|\mathcal{K}_c|}\sum_{k\in\mathcal{K}_c}\Delta_{\mathrm{learn}}(k)$ and $\Delta_{\mathrm{forget}}(c)=\frac{1}{|\mathcal{K}_c|}\sum_{k\in\mathcal{K}_c}\Delta_{\mathrm{forget}}(k)$, respectively.
\end{definition}
\vspace{-3pt}

\subsection{Experiment Finding}~\label{sec:rq1_finding}

\begin{findingbox}
\begin{finding}
Concepts exhibit substantial cross-concept variation in both learning degree and forgetting degree, indicating that LLMs acquire and forget different concepts to markedly different extents under the same training regime.
\end{finding}
\end{findingbox}

 \begin{figure}[htbp] 
	\includegraphics[width=\linewidth]{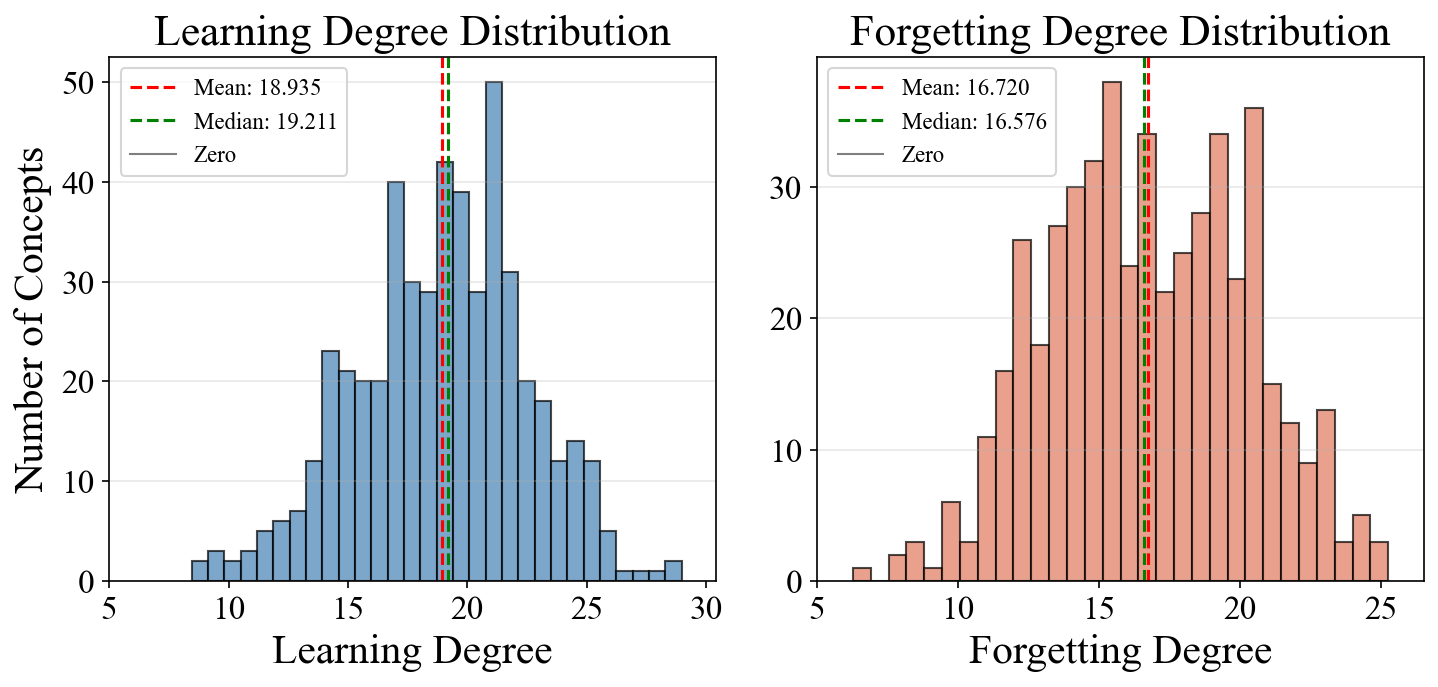}
	\caption{Distribution of learning and forgetting degree across concepts.}
    \label{fig:concept_learn_forget_hist}
    \vspace{-1pt}
\end{figure}

\begin{sloppypar}
\paragraph{Concept Learning/Forgetting Degree Distribution.} 
Figure~\ref{fig:concept_learn_forget_hist} shows the distribution of concept learning (logit increase) and forgetting (logit decrease) degrees across 500 concepts on the FICO test set. The distribution is unimodal but widely spread, indicating substantial variability in how concepts are learned and forgotten. Practically, some concepts are learned more readily and robustly, while others require greater effort to learn and retain. This observation motivates our subsequent analysis, which seeks to identify indicators that can predict the differing learning and forgetting behaviors across concepts.

\end{sloppypar}

 \vspace{-1pt}

\begin{findingbox} 
\begin{finding} \label{finding:finding_circuit_allocation}
Concept learning degree and forgetting degree shows non-trivial, consistent correlations with multiple circuit graph metrics, suggesting that circuit topology can serve as an informative indicator of concept learning and forgetting dynamics.
\end{finding}
\end{findingbox}


\begin{table}[t]
\centering
\resizebox{\linewidth}{!}{
\setlength{\tabcolsep}{6pt}
\renewcommand{\arraystretch}{0.92}
\begin{tabular}{l l c c}
\toprule
Type & Metric & Learning & Forgetting \\
\midrule
Circuit Topology (70\%) & Density & 0.4474 & 0.4585 \\
Circuit Topology (70\%) & $k$-core & 0.4468 & 0.4587 \\
Circuit Topology (70\%) & Eigenvector Centrality & 0.4403 & 0.4459 \\
Circuit Topology (70\%) & Global Efficiency & 0.4085 & 0.4322 \\
\midrule
Circuit Topology (80\%) & Density &  0.4723 &  0.4785 \\
Circuit Topology (80\%) & $k$-core &  0.4475 &  0.4610\\
Circuit Topology (80\%) & Eigenvector Centrality & 0.3235 &0.3314\\
Circuit Topology (80\%) & Global Efficiency & 0.4662 & 0.4747  \\
\midrule
Activation & Mean EAP-IG & 0.2680  
           & 0.2784   \\
Activation & STD EAP-IG  & 0.2398  
           & 0.2494   \\
Activation & Max EAP-IG  & 0.1575  
           & 0.1342  \\
\midrule
Random Topology & Density & -0.0078 & -0.0255 \\
Random Topology & $k$-core & -0.0458 & -0.0437 \\
Random Topology & Eigenvector Centrality & 0.0473 & 0.0737 \\
Random Topology & Global Efficiency & -0.0559 & -0.0769 \\
\bottomrule
\end{tabular}
}
\caption{Correlation between learning/forgetting degree and circuit metrics (Spearman $\rho$) }
\label{tab:baseline}
\end{table}

\paragraph{Correlation between Concept Learning/Forgetting and Concept Circuits.}
As shown in Table~\ref{tab:baseline}, our circuit topology metrics exhibit stronger correlations~\footnote{$p$-values are provided in Appendix~\ref{sec:complete_result}} with concept learning degree ($\rho\approx0.41$--$0.45$) than either edge-level EAP-IG statistics~\cite{hanna2024have} or random topology metrics ($|\rho|<0.08$)~\footnote{See baseline metrics in Appendix~\ref{sec:baseline}.}, indicating that circuit topology may be systematically associated with how concepts are acquired. For example, positive correlations with eigenvector centrality and $k$-core suggest that circuits with centralized bottlenecks and stable structural cores tend to achieve larger logit gains during learning. Notably, these same topological properties are also associated with greater vulnerability under continual training. In particular, concepts whose circuits are dominated by a small set of influential hub nodes, reflected by high eigenvector centrality, tend to exhibit larger forgetting degrees, potentially because these centralized components are more susceptible to perturbation during subsequent training.

This structural trade-off is also reflected behaviorally: concepts that are learned more strongly also tend to be forgotten more strongly under subsequent training, revealing a positive correlation between learning degree and forgetting degree, as detailed in Appendix~\ref{sec:learn_forget_correlation}. Furthermore, these correlations remain comparable when increasing the circuit performance-preservation threshold from 70\% to 80\%, suggesting that the observed circuit-behavior associations are robust to the threshold choice. Taken together, these results indicate that circuit topology captures informative signals of both concept learning and forgetting dynamics in LLMs. Such signals may provide practical guidance for concept-aware continual pretraining decisions, such as how to allocate training effort between newly introduced concepts and previously learned concepts, as further explored in Section~\ref{sec:circuit_experience_replay}.

\subsubsection{Experiment Finding}

\begin{findingbox}
\begin{finding}
LLMs exhibit consistent stage-wise temporal patterns across multiple circuit graph metrics: the metrics increase and then stabilize during concept acquisition, and show a transient increase followed by a gradual decrease and stabilization during concept forgetting.
\end{finding}
\end{findingbox}


\begin{figure}[htbp]
    \centering
    \begin{subfigure}[b]{\linewidth}
        \centering
        \includegraphics[width=\linewidth]{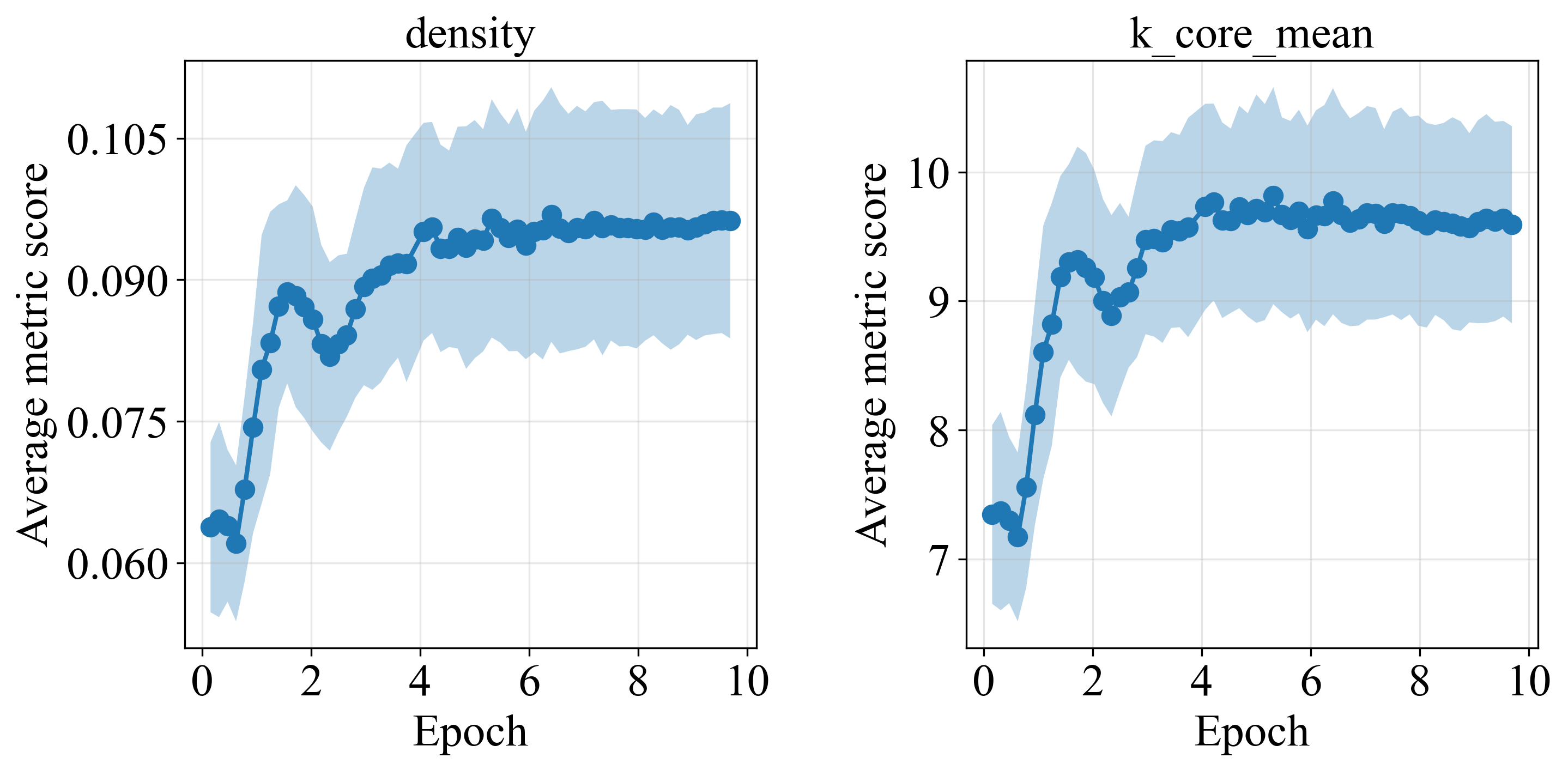}
        \caption{Concept Acquisition.}
        \label{fig:rq1_learn_over_steps_gpt2}
    \end{subfigure}

    \begin{subfigure}[b]{\linewidth}
        \centering
        \includegraphics[width=\linewidth]{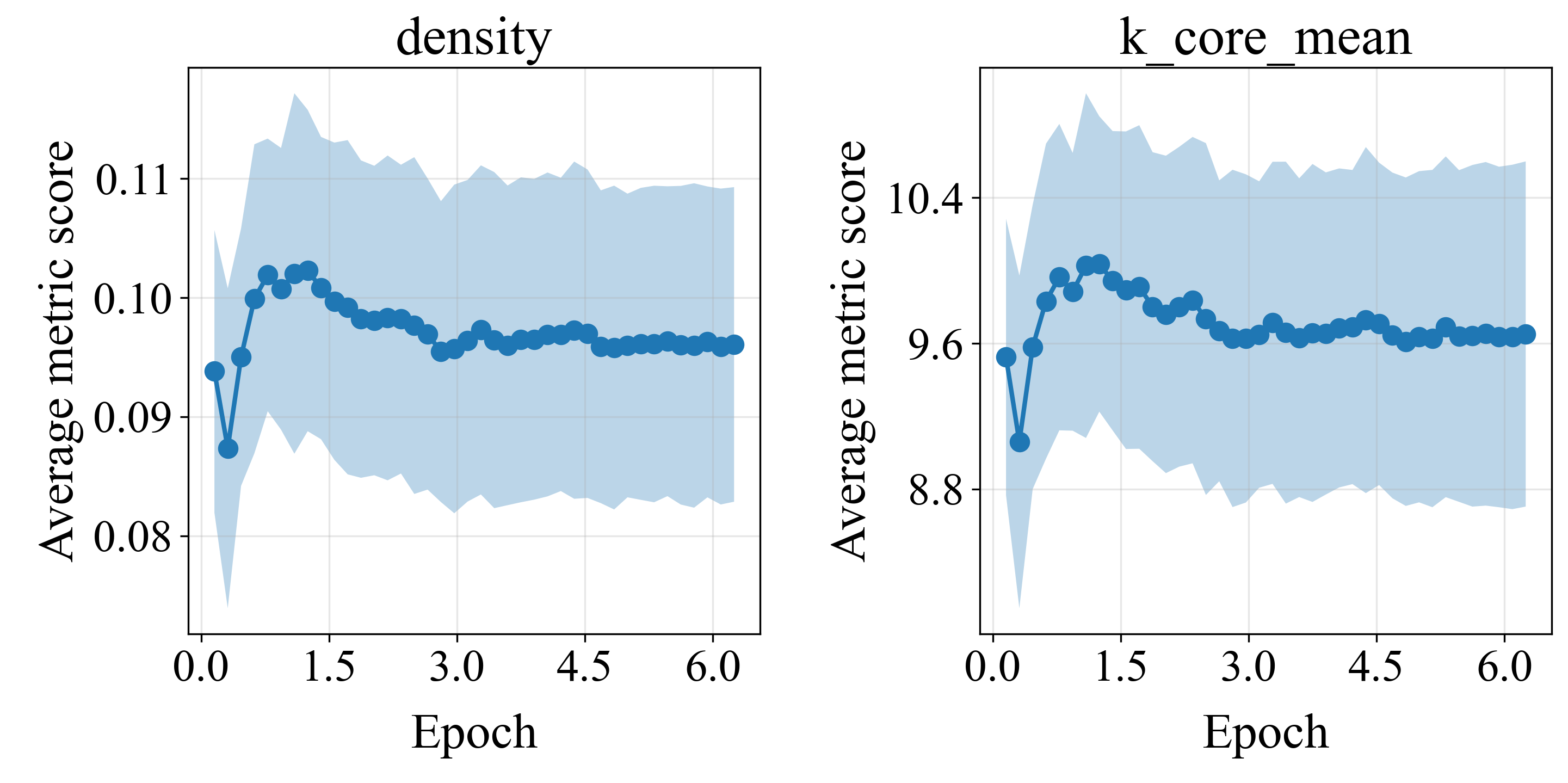}
        \caption{Concept Forgetting.}
        \label{fig:rq2_forget_over_steps_gpt2}
    \end{subfigure}
    \caption{Evolution of circuit graph metrics during the learning and forgetting stages, shown for density and mean $k$-core; all four metrics are in Appendix~\ref{sec:metric_evolution}.}
    \label{fig:circuit_evolution}
\end{figure}


\paragraph{Evolution of Circuit Topology.} We track how circuit topology evolves during both training stages. Figure~\ref{fig:rq1_learn_over_steps_gpt2} shows the learning stage: all graph metrics rise sharply during early learning (within roughly the first 1.5 epochs), continue to increase gradually with minor fluctuations, and stabilize after epoch ${\sim}5$, suggesting that concept acquisition is accompanied by the progressive formation and consolidation of circuit structure. Figure~\ref{fig:rq2_forget_over_steps_gpt2} shows a consistent temporal pattern of circuit graph metrics during the forgetting stage: an initial increase followed by a gradual decrease and eventual stabilization. 
The early increase may indicate transient entanglement of previously learned concept circuits with newly introduced knowledge during continued training, while the subsequent decrease and stabilization may reflect structural relaxation and convergence to a weaker representation under interference. 

\vspace{-4pt}
\section{How Do Interference and Synergy Arise Across Concepts and Conceptual Knowledge During Joint Training?}
The analyses above suggest a trade-off between acquisition strength and vulnerability under continual training. This raises a natural question: is such vulnerability intrinsic to individual concepts, or is it also shaped by interactions among concurrently learned concepts?
\vspace{-1pt}
\subsection{Interference and Synergy across Concepts}\label{sec:concept_interference_exp}
\subsubsection{Experiment Design}


To examine cross-concept interactions during joint training, we construct relatedness-based groups for each target concept. Relatedness is defined as the cosine similarity between embeddings of the concatenated conceptual knowledge associated with each concept, computed using Qwen3-Embedding-4B~\cite{yang2025qwen3}. For each concept $c$, we rank all other concepts by similarity and select the top-$K$, middle-$K$, and bottom-$K$ concepts as the highly, moderately, and weakly related groups, respectively, with $K=100$. We then perform joint training for each concept $c$ under three settings, where $c$ is trained together with concepts from one of the three relatedness groups\footnote{A controlled experiment disentangling semantic relatedness from surface-level overlap shows that the interference persists at zero lexical overlap; see Appendix~\ref{sec:controlled_interference}.}. Performance is evaluated using (1) the average logit assigned to target objects and (2) the corresponding average probability. This procedure is repeated across all concepts in the \dataset{} test set to characterize cross-concept interactions.

\begin{table}[t]
\centering
\small
\resizebox{\linewidth}{!}{
\setlength{\tabcolsep}{5pt}
\begin{tabular}{l ccc ccc}
\toprule
& \multicolumn{3}{c}{Avg.\ probability} & \multicolumn{3}{c}{Avg.\ logit} \\
\cmidrule(lr){2-4} \cmidrule(lr){5-7}
Model & Most & Med. & Least & Most & Med. & Least \\
\midrule
GPT-2 Large & 0.575 & 0.672 & \textbf{0.759} & 14.37 & 14.90 & \textbf{16.24} \\
LLaMA-3.2-1B & 0.754 & 0.811 & \textbf{0.877} & 18.23 & 18.55 & \textbf{20.28} \\
LLaMA-3.1-8B & 0.781 & 0.846 & \textbf{0.889} & 23.65 & 24.18 & \textbf{25.76} \\
Qwen3-4B-Base & 0.807 & 0.828 & \textbf{0.881} & 26.74 & 27.10 & \textbf{28.73} \\
\bottomrule
\end{tabular}
}
\caption{Concept interference across models: average target probability and logit after co-training targets with its most, moderately, or least related concepts. (Appendix~\ref{sec:llama}).}
\label{tab:interference_summary}
\end{table}

\begin{figure}[t]
    \centering
    \includegraphics[width=\linewidth]{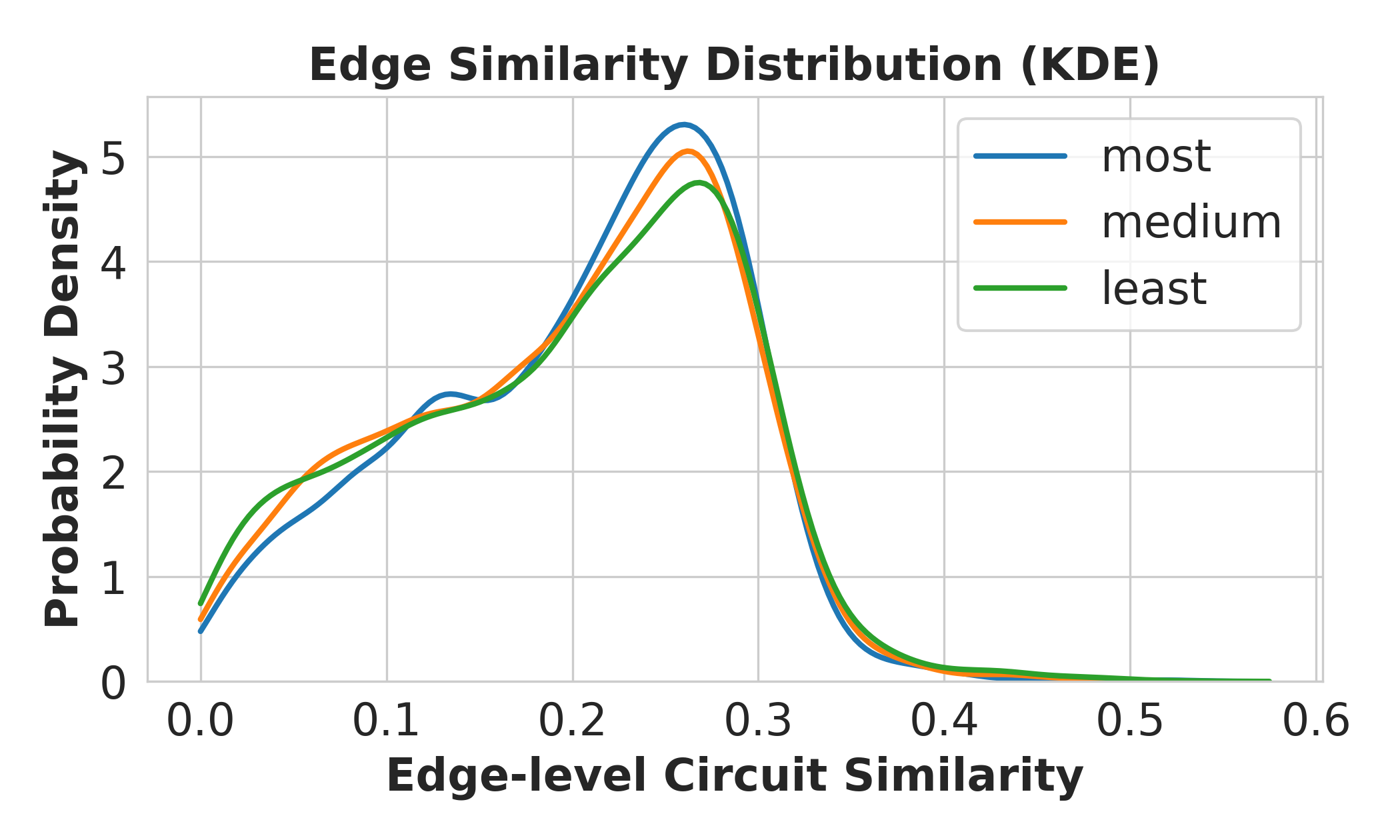}
    \caption{Distribution of Jaccard similarity between circuits of co-trained concept pairs, by relatedness group.}
    \label{fig:concept_interference_circuit_similarity}
    \vspace{-3pt}
\end{figure}



\subsubsection{Experiment Finding}
\begin{findingbox}
\begin{finding}
Training with highly related concepts yields lower performance than training with weakly related concepts, indicating stronger interference among semantically similar concepts during joint learning.
\end{finding}
\end{findingbox}


\begin{sloppypar}
Table~\ref{tab:interference_summary} shows a clear and consistent dependence on semantic relatedness. Across both evaluation metrics (average logit and average probability), training with weakly related concepts achieves the highest performance (75.9\%), substantially outperforming training the highly related ( 57.5\%) and moderately related (67.2\%) concepts. These results demonstrate that cross-concept interactions meaningfully affect how effectively a target concept can be acquired in a multi-concept setting. 
\end{sloppypar}

\begin{sloppypar}
To explore a potential mechanism associated with this effect, we analyze the degree of overlap between the internal representations of co-trained concepts. For each target concept $c$, we pair it with each of its related concepts drawn from three semantic similarity groups: highly related, moderately related, and weakly related. For each pair $(c, c')$, we obtain their concept circuits and compute the Jaccard similarity between their edge sets. Figure~\ref{fig:concept_interference_circuit_similarity} visualizes the resulting similarity distributions using KDE plots. The highly related setting exhibits a sharper, higher-density peak at moderate circuit similarity, indicating consistent reuse of overlapping computational pathways between target and auxiliary concepts, which may induce sustained representational competition during joint training, consistent with the stronger interference we observe. Instead, the moderately and weakly related settings show lower and more dispersed similarity distributions, reflecting reduced circuit overlap, consistent with the weaker interference observed in these settings. This observation motivates interference-aware data scheduling that reduces circuit similarity among co-trained concepts within the same batch.
\end{sloppypar}

\subsection{Interference and Synergy across Knowledge} 

\subsubsection{Experiment Design}


Beyond concepts, we ask whether \emph{different types of conceptual knowledge} can also exhibit interference or synergy, even when they describe the \emph{same} concept. We consider five high-level semantic knowledge categories: (1) Hyponym \& Hypernym (HAH), (2) Synonym \& Antonym (SAA), (3) Meronym \& Holonym (MAH), (4) Property \& Affordance (PAA), and (5) Spatial Relation (SR), as detailed in Appendix~\ref{sec:knowledge_type}. To study interactions across knowledge types, we adopt a pairwise continual-training setup. In Stage~1, the model is trained on either a source knowledge type $S$ or the BIO dataset~\cite{allen2023physics}, which does not contain conceptual knowledge, for the same training steps. In Stage~2, it is continually trained on a target knowledge type $T$, yielding the checkpoints $\pi_{S\rightarrow T}$ and $\pi_{\mathrm{BIO}\rightarrow T}$, respectively. We define \textbf{paired transferability} from $S$ to $T$ as
\vspace{-1pt}
\begin{equation}
\small
T(S \rightarrow T)
= \frac{\operatorname{logit}(T \mid \pi_{S\rightarrow T})-\operatorname{logit}(T \mid \pi_{\mathrm{BIO}\rightarrow T})}
{\left|\operatorname{logit}(T \mid \pi_{\mathrm{BIO}\rightarrow T})\right|}.
\end{equation}
\vspace{-3pt}
Positive $T(S \rightarrow T)$ indicates \emph{synergy}, where learning $S$ facilitates subsequent learning of $T$, while negative values indicate \emph{interference}.

\subsubsection{Experiment Finding}

\begin{findingbox}
\begin{finding}
Substantial and asymmetric transfer effects emerge across knowledge types, where pretraining on one type can facilitate learning of another, with highly directional and uneven benefits across ordered pairs.
\end{finding}
\end{findingbox}


 \begin{figure}[htbp] 
	\centering
	\includegraphics[width=\linewidth]{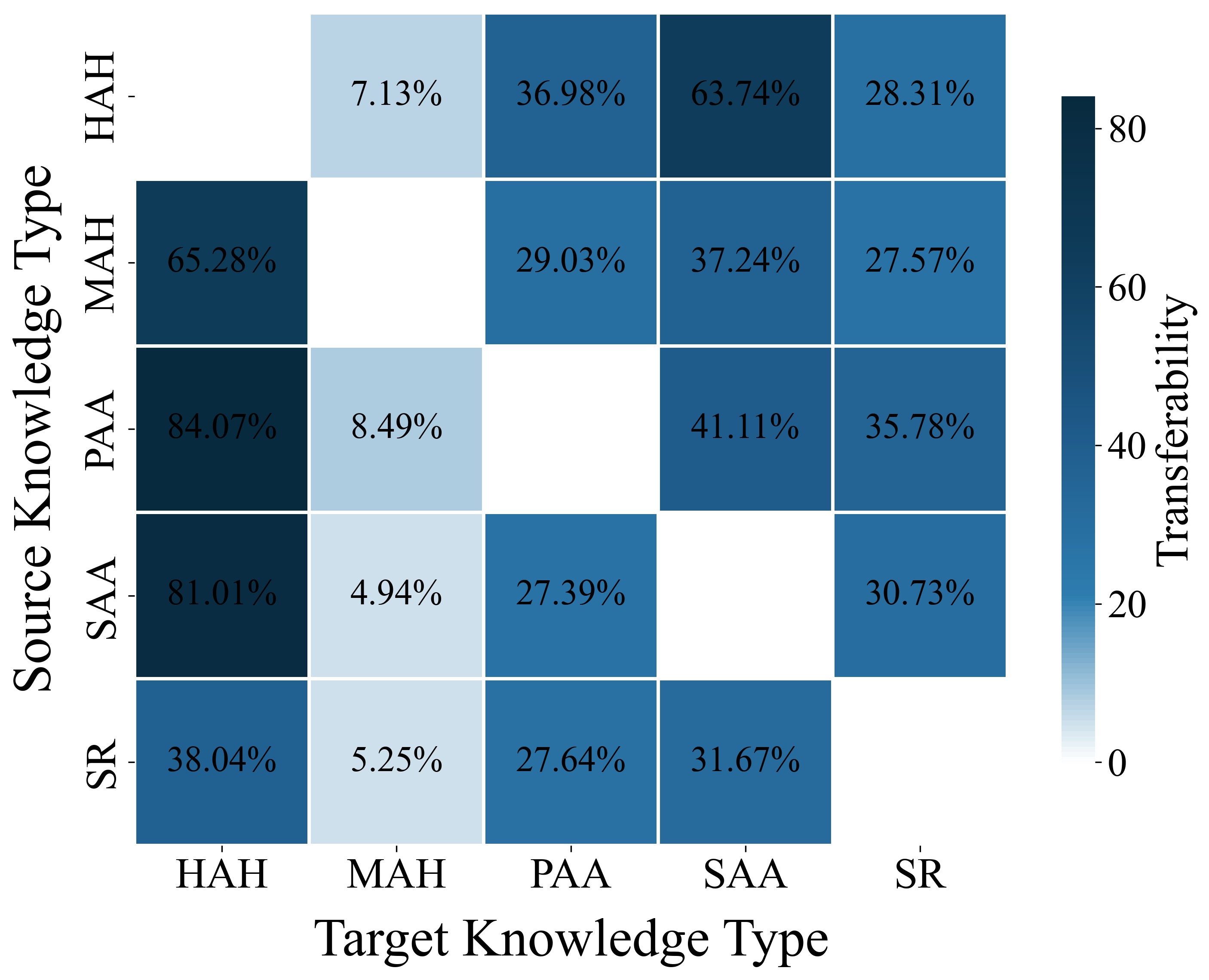}
	\caption{Paired transferability across knowledge}
	\label{fig:knowledge_dependency}
\end{figure}


\begin{sloppypar}
Figure~\ref{fig:knowledge_dependency} provides a fine-grained view of pairwise synergy among the different knowledge types, where each cell shows the paired transferability $T(S \rightarrow T)$ when a \emph{source} type $S$ is trained before a \emph{target} type $T$. While all five categories encode concept-related information, they capture distinct semantic facets, leading to non-redundant and uneven transfer behaviors. The heatmap reveals heterogeneous and directional effects: for example, pretraining on Property \& Affordance (PAA) yields substantial gains when transferring to Hyponym \& Hypernym (HAH) and Synonym \& Antonym (SAA), suggesting that learning functional and attribute-level regularities can effectively scaffold the acquisition of more abstract relational structures. In contrast, the reverse directions exhibit markedly weaker transfer, highlighting the asymmetry of these interactions. Moreover, some target types, such as Meronym \& Holonym (MAH), show relatively small improvements across most sources, indicating higher intrinsic learnability and reduced sensitivity to prior knowledge-type pretraining. Overall, these findings suggest practical guidance for future training curricula, such as reordering training data to place knowledge types with strong positive transfer earlier, thereby encouraging synergy and efficiency of downstream concept learning.

\begin{figure}[htbp]
	\centering
	\includegraphics[width=1\linewidth]{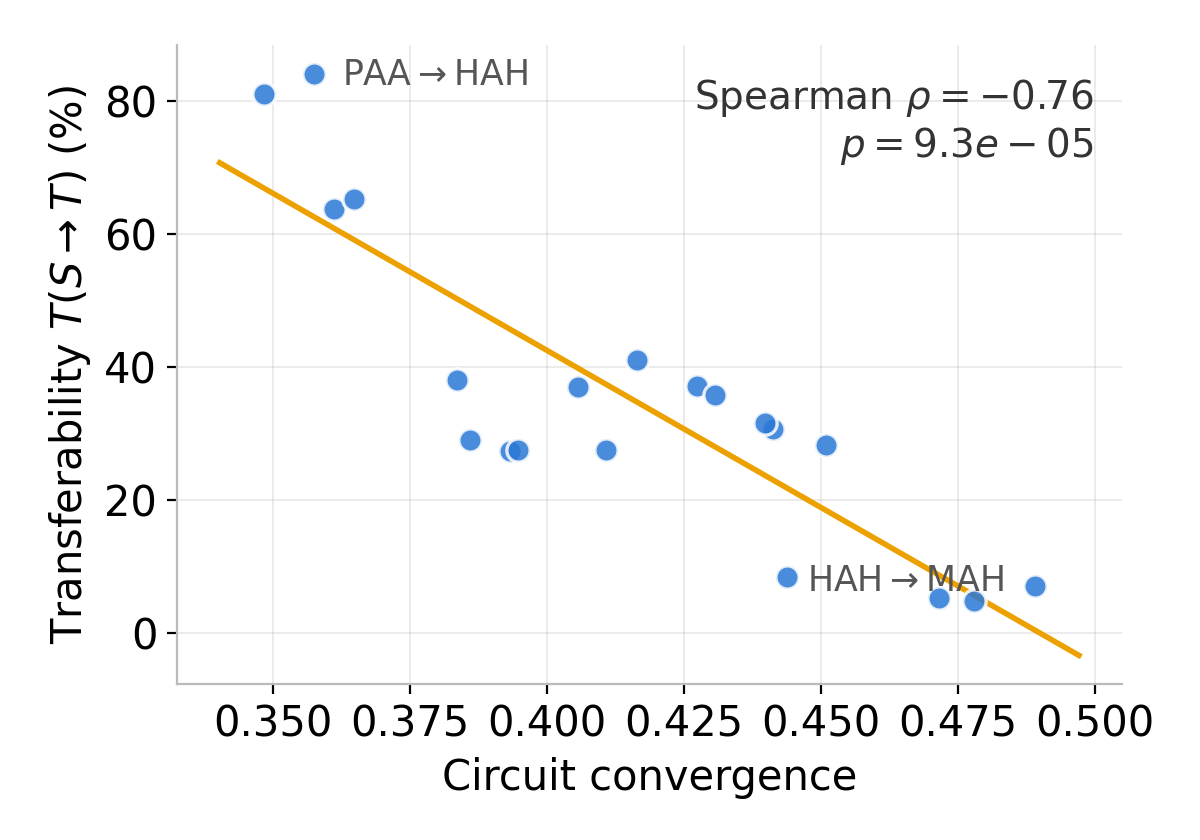}
	\caption{Transferability versus circuit convergence across the 20 ordered knowledge-type directions (GPT-2 Large). Directions whose $\pi_{S\rightarrow T}$ circuit remains least similar to the $\pi_{\mathrm{BIO}\rightarrow T}$ circuit transfer best.}
	\label{fig:type_circuit_convergence}
\end{figure}

To explore a potential mechanism behind this asymmetry, we extract one EAP-IG circuit per knowledge type from both Stage-2 checkpoints. For each direction $S\rightarrow T$, we compute \textbf{convergence}, the Jaccard similarity between the circuit extracted from $\pi_{S\rightarrow T}$ and the one extracted from $\pi_{\mathrm{BIO}\rightarrow T}$, i.e., how closely the transferred circuit resembles the one formed without a related source type. As shown in Figure~\ref{fig:type_circuit_convergence}, convergence is negatively correlated with transferability across all 20 ordered directions (Spearman $\rho=-0.76$, $p=9.3\times10^{-5}$): the best-transferring directions, such as PAA$\rightarrow$HAH, are those whose $\pi_{S\rightarrow T}$ circuits deviate most from the $\pi_{\mathrm{BIO}\rightarrow T}$ circuit.  Directions with stronger transfer thus arrive at target circuits that depart most from the one formed without a relevant source, indicating that the source type reshapes the target computation rather than leaving it to re-form as it would from BIO alone. This connects Findings~4 and~5 at the circuit level: cross-knowledge interactions leave structural traces whether the related knowledge is trained jointly (Finding~4) or beforehand (Finding~5).
\end{sloppypar}

\begin{table}[hbt!]
\small
\centering
 \resizebox{\linewidth}{!}{%
    \begin{tabular}{l|c|c|c|c}
    \toprule
    \textbf{Method} & \textbf{Replay Ratio} & \textbf{Accuracy(\%)} & \textbf{Hit@5(\%)} & \textbf{Logit} \\
    \midrule
    w/o Replay & 0 & 1.93 & 7.23 & 3.93 \\
 \midrule
    Random Replay & 0.01  &11.76   & 19.96  &6.77 \\
    Gradient Replay & 0.01 & 15.17 & 28.35 & 7.96 \\
    Circuit Replay & 0.01  &  22.49 & 38.78  & 8.85  \\
     \midrule
    Random Replay & 0.05 &  40.39 &  61.11 &10.88 \\
    Gradient Replay & 0.05 & 41.96 & 59.76 & 11.17 \\
    Circuit Replay & 0.05  & 51.58  &  71.06 &12.72   \\
         \midrule
    Random Replay & 0.10  & 60.94  & 77.36  & 12.98\\
    Gradient Replay & 0.10 & 59.57 & 74.46 & 12.95 \\
    Circuit Replay & 0.10  & 69.15  &  82.74 & 14.26  \\
         \midrule
    Random Replay & 0.50  &  82.48 & 91.42  & 16.34\\
    Gradient Replay & 0.50 & 81.90 & 90.44 & 16.81 \\
    Circuit Replay & 0.50  & 86.48  &  93.04 & 17.42  \\
    \bottomrule
    \end{tabular}
    }
    \captionof{table}{Retention performance on the AntiLeak dataset under continual pretraining.}
    \label{tab:real_world}
\end{table}

\begin{table}[t]
\centering
\resizebox{\linewidth}{!}{
\setlength{\tabcolsep}{3.5pt}
\renewcommand{\arraystretch}{0.92}
\begin{tabular}{l cc cc cc cc}
\toprule
 & \multicolumn{2}{c}{GPT-2 Large} & \multicolumn{2}{c}{LLaMA-3.2-1B} & \multicolumn{2}{c}{LLaMA-3.1-8B} & \multicolumn{2}{c}{Qwen3-4B-Base} \\
\cmidrule(lr){2-3} \cmidrule(lr){4-5} \cmidrule(lr){6-7} \cmidrule(lr){8-9}
Metric & Learn & Forget & Learn & Forget & Learn & Forget & Learn & Forget \\
\midrule
Density & 0.447 & 0.459 & 0.399 & 0.411 & 0.349 & 0.386 & 0.333 & 0.416 \\
$k$-core & 0.447 & 0.459 & 0.375 & 0.381 & 0.349 & 0.389 & 0.319 & 0.409 \\
Eigenvector Centrality & 0.440 & 0.446 & 0.163 & 0.175 & 0.148 & 0.182 & 0.328 & 0.394 \\
Global Efficiency & 0.409 & 0.432 & 0.332 & 0.372 & 0.201 & 0.207 & 0.285 & 0.304 \\
\bottomrule
\end{tabular}
}
\caption{Correlation between learning (Learn) and forgetting (Forget) degree and circuit topology metrics across model families (Spearman $\rho$). All $p$-values are below $10^{-3}$; full $p$-values are reported in Table~\ref{tab:correlation_pvalues}.}
\label{tab:correlation}
\end{table}

\section{How Circuit Analysis Can Inform Better Continual Pretraining?}
\label{sec:circuit_experience_replay}
As shown in Finding~\ref{finding:finding_circuit_allocation}, circuit topology metrics provide informative signals of concept forgetting during continual pre-training of LLMs. Leveraging this insight, we derive a simple yet actionable training strategy, \textbf{Circuit-aware Experience Replay}, that prioritizes concepts predicted to be more susceptible to forgetting and allocates additional training effort to them (e.g., via experience replay~\cite{isele2018selective}).

\paragraph{Method}
Given a model $\pi_{\text{learn}}$ that has learned a sequence of concepts $\{c_0, c_1, \dots\}$, our goal is to further train the model on a new dataset while preserving performance on previously learned concepts. 
To this end, we extract concept circuits for all learned concepts and compute their corresponding circuit topology metrics. 
Based on topology metric\footnote{As shown in Table~\ref{tab:baseline}, the Density metric exhibits the strongest correlation with forgetting, and is therefore used to rank concepts.}, we select the top $p\%$ of concepts with the largest topology metric values and replay their associated training samples during continual pre-training on the new dataset. \footnote{Circuit extraction computational overhead is in Appendix~\ref{sec:overhead}.}.

\paragraph{Experiment}
To evaluate this strategy in a realistic setting, we use the AntiLeak dataset~\cite{wu2025antileakbench}, which contains newly introduced real-world knowledge not previously learned by existing LLMs. We train GPT-2 Large on AntiLeak to obtain $\pi_{\text{learn}}$, continue training on the BIO dataset~\cite{allen2023physics}, and measure retention on AntiLeak with \textbf{Logit}, \textbf{Accuracy}, and \textbf{Hit@5} (Appendix~\ref{sec:metrics}). We compare \textbf{Circuit-aware Experience Replay} against \textbf{Random Replay} and \textbf{Gradient Replay}, which ranks concepts by per-sample gradient norm at the final stage-1 checkpoint. As shown in Table~\ref{tab:real_world}, circuit-aware replay outperforms both baselines at every replay ratio and metric (e.g., 22.49\% vs.\ 15.17\% accuracy at ratio 0.01). The gap reflects what the two signals measure: gradient norm indicates difficulty rather than forgetting risk, so gradient replay spends its budget on weakly learned facts with little to protect. 


\section{Robustness and Ablations}
\label{sec:ablation}
We assess the stability of our findings along three additional axes: model family and scale, and the continual-training horizon.

\paragraph{Model family and scale}
We repeat RQ1 and RQ2 on LLaMA-3.2-1B, LLaMA-3.1-8B, and Qwen3-4B-Base. As shown in Table~\ref{tab:correlation}, density and $k$-core remain at $\rho\approx0.3$ or above for both learning and forgetting degree on all four models, while eigenvector centrality is as predictive as density on Qwen3-4B-Base but weaker on LLaMA, whose circuits exhibit uniformly small eigenvector centrality across concepts ($\approx$0.027 vs.\ $\approx$0.040 in GPT-2), compressing the metric's range. The interference and transfer patterns also replicate across families. As summarized in Table~\ref{tab:interference_summary}, co-training a target with its least related concepts yields the highest target probability and logit on all four models. Knowledge-type transferability likewise remains heterogeneous and directional (Appendix~\ref{sec:llama}).


\paragraph{Longer training horizon}
Finally, we extend the two-stage FICO$\rightarrow$BIO setup to a four-stage FICO$\rightarrow$BIO$\rightarrow$FICO$\rightarrow$BIO sequence, testing whether our conclusions persist under repeated acquisition and interference phases. Graph metrics correlate even more strongly with learning degree in this setting, reaching $\rho=0.512$ (Appendix~\ref{sec:longer_continual_training}), which may indicate that repeated training encourages the reuse or consolidation of previously learned computational structures.

\section{Related Work}


\paragraph{LLM Knowledge Acquisition.}
A growing body of work studies learning mechanism behinds LLM knowledge acquisition. One line uses synthetic or fictional corpora, e.g., biographies of fabricated individuals~\cite{allen2023physics,how2025dynamics,ou-etal-2025-llms,zhu2025effective,feng2024extractive}, Wikipedia-style entries for fictional entities~\cite{chang2024large}, or \textbf{post-cutoff data}~\cite{huang2024demystifying}, to examine how data properties, training choices, and curricula (e.g., ordering dependencies between facts and implications) affect learning. Another line~\cite{chang2024large,xu2024tracking,leybzon2024learning,im2024understanding,qian2024towards,tigges2024llm} leverages mechanistic interpretability to analyze LLM learning by tracing how internal representations evolve across training stages. The third line~\cite{ren2024learning,chen2023sudden,jain2023mechan} identify phase transitions that reveal discrete, objective-dependent shifts in model behavior during training. In contrast, we target \textbf{concept-level} knowledge rather than isolated factual triples, enabling the study of higher-order phenomena such as concept learning, forgetting, interference, and synergy, while also opening the door to future work on concept generalization, compositional generalization, and hierarchical interactions~\cite{zaeem2021cause}.

\paragraph{LLM Concept Probing and Editing.}
Prior work studies conceptual knowledge in LLMs mainly through: (i) \textbf{prompt-based probing} of concept properties and relations~\cite{gu-etal-2023-language,liao2023concept,shani-etal-2023-towards,zheng2024concept,peng-etal-2022-copen}; (ii) \textbf{definition-name alignment} tests (e.g., reverse-dictionary probes) that assess mapping between descriptions and names~\cite{xu2024tip}; and (iii) \textbf{compositional binding and consistency} evaluations that test correct attribution of concept knowledge to instances and consistency across hierarchies~\cite{he2023language,sosa2024reasoning,sahu2022unpacking}. Complementing these behavioral probes, mechanistic interpretability aims to localize internal components responsible for concept-related behavior~\cite{aljaafari2024mechanics,wang-etal-2024-editing}. Unlike these largely static evaluations, we focus on the \textbf{dynamics of concept learning}: we connect internal \emph{concept circuits} to acquisition, forgetting, and cross-concept interactions, providing a mechanistic, time-resolved complement to existing probing approaches.

\section{Conclusion} 

In this work, we study how large language models acquire, retain, and forget concepts during continual pretraining by linking output behavior with circuit-level interpretability. We introduce the \dataset{} dataset, define \textbf{Concept Circuits} as internal representations of concepts, and use graph metrics to characterize \textbf{Circuit Topology}. Our experiments reveal consistent patterns in concept learning, forgetting, interference, and synergy, showing that circuit topology provides meaningful signals of learning dynamics  that can guide experience replay to mitigate forgetting. Our findings offer a step toward more concept-aware, interpretable, and robust continual pretraining of LLMs.

\section*{Limitations}
Our study has several limitations. (1) \textbf{Limited Model Scale}. Due to computational constraints, our experiments focus on GPT-2 Large (0.7B), with LLaMA-3.2-1B, LLaMA-3.1-8B, and Qwen3-4B-Base used for ablation; extending the analysis to larger LLMs remains important future work. (2) \textbf{Limited Exploration of Actionable Training Strategies}. Our focus is analytical, and we instantiate \textbf{Circuit-aware Experience Replay} only as a proof-of-concept, evaluated on GPT-2 Large and AntiLeak. Future work should validate it more broadly and explore other concept-aware strategies such as adaptive effort allocation, interference-aware data scheduling, or curriculum design. (3) \textbf{Controlled Operationalization of Concept Learning}. \dataset{} operationalizes concept learning as templated prefix-target prediction over fictional concept-relation triples, with disjoint test templates that isolate relational generalization from surface-form memorization (Section~\ref{sec:fictional_concept}). This controlled scope leaves broader conceptual abstraction, hierarchical interactions, and compositional generalization outside our evaluation. (4) \textbf{Pre-training Focus}. We study continual pre-training; since post-training is likewise continued training on new data, topology metrics could predict which knowledge is vulnerable and circuit-aware replay could mitigate the ``alignment tax''~\cite{ortu2024competition,prakash2024finetuning,lee2024mechanistic}, but we leave empirical validation to future work.

\section*{Acknowledgments}

This research is partially supported by the award No. \#2604309 from the Faculty Early Career Development Program (CAREER) and the award No. \#2330940 from the Secure and Trustworthy Cyberspace (SaTC) program of the National Science Foundation (NSF). The views and conclusions contained herein are those of the authors and should not be interpreted as necessarily representing the official policies, either expressed or implied, of the U.S. Government. The U.S. Government is authorized to reproduce and distribute reprints for governmental purposes notwithstanding any copyright annotation therein.

\section*{Ethical considerations}
We carefully follow the ACM Code of Ethics and have not found potential societal impacts or risks so far. To the best of our knowledge, this work has no notable harmful effects and uses, environmental impact, fairness considerations, privacy considerations, security considerations, or other potential risks. We will release the constructed \dataset{} dataset to support reproducibility and future research. Since \dataset{} is derived from ConceptNet, which is distributed under the Creative Commons Attribution-ShareAlike 4.0 license, our released dataset will follow the same license and include appropriate attribution to ConceptNet. Users of the dataset should comply with the corresponding attribution and share-alike terms. The dataset does not contain information that names or uniquely identifies individual people, and we manually check that it does not include offensive content. We have not identified notable societal, fairness, privacy, security, environmental, or other risks beyond those associated with standard use of commonsense knowledge resources. The dataset is in English and focuses on concept-centric commonsense knowledge. It covers broad everyday domains represented in ConceptNet, including objects, animals, locations, properties, affordances, part-whole relations, spatial relations, taxonomic relations, and synonym/antonym relations.

\bibliography{anthology,sections/custom}
 
\newpage
\appendix

\section{Knowledge Type Grouping and Knowledge Filtering}
\label{sec:knowledge_type}

\subsection{Knowledge Type Grouping}
\label{sec:knowledge_type_grouping}

\begin{table*}[t]
\centering
\begin{tabular}{p{4.2cm} p{6.5cm}}
\toprule
\textbf{High-level Knowledge Type} & \textbf{Relation Types} \\
\midrule
Hyponym and Hypernym &
\textit{IsA}, \textit{DefinedAs}, \textit{FormOf}, \textit{InstanceOf} \\

Synonym and Antonym &
\textit{Synonym}, \textit{SimilarTo}, \textit{Antonym}, \textit{DistinctFrom} \\

Meronym and Holonym &
\textit{PartOf}, \textit{HasA}, \textit{MadeOf} \\

Property and Affordance &
\textit{HasProperty}, \textit{UsedFor}, \textit{CapableOf}, \textit{ReceivesAction} \\

Spatial Relation &
\textit{AtLocation}, \textit{LocatedNear} \\
\midrule
\multicolumn{2}{l}{\textit{Excluded Relation Categories}} \\
\midrule
Causality \& Event &
\textit{Causes}, \textit{MotivatedByGoal}, \textit{HasPrerequisite}, \textit{HasSubevent}, \textit{HasFirstSubevent}, \textit{HasLastSubevent}, \textit{CreatedBy} \\

Desire &
\textit{Desires}, \textit{CausesDesire} \\

Lexical / Etymological &
\textit{DerivedFrom}, \textit{EtymologicallyDerivedFrom}, \textit{EtymologicallyRelatedTo} \\

Other &
\textit{RelatedTo}, \textit{HasContext}, \textit{ExternalURL}, \textit{SymbolOf} \\
\bottomrule
\end{tabular}
\caption{Mapping from fine-grained relation types to high-level knowledge categories. We retain five concept-centric categories for experiments and exclude relation types that primarily encode events, lexical form, or noisy contextual associations.}
\label{tab:relation_type_mapping}
\end{table*}

Our dataset is constructed from the ConceptNet knowledge graph, where each knowledge instance is represented as a triplet \(\langle \text{subject}, \text{relation}, \text{object} \rangle\). Since ConceptNet contains a large number of fine-grained relation types, we group them into a smaller set of semantically coherent, high-level knowledge categories to facilitate concept-level analysis. Each relation is assigned to a category according to its primary semantic function.

As shown in Table~\ref{tab:relation_type_mapping}, taxonomic and definitional relations (\textit{IsA}, \textit{DefinedAs}, \textit{FormOf}, \textit{InstanceOf}) are grouped into \textbf{Hyponym and Hypernym}. Relations expressing similarity or contrast (\textit{Synonym}, \textit{SimilarTo}, \textit{Antonym}, \textit{DistinctFrom}) are grouped into \textbf{Synonym and Antonym}. Part-whole and compositional relations (\textit{PartOf}, \textit{HasA}, \textit{MadeOf}) are grouped into \textbf{Meronym and Holonym}. Relations describing properties, affordances, or functional roles (\textit{HasProperty}, \textit{UsedFor}, \textit{CapableOf}, \textit{ReceivesAction}) are grouped into \textbf{Property and Affordance}. Spatial relations (\textit{AtLocation}, \textit{LocatedNear}) are grouped into \textbf{Spatial Relation}.

\subsection{Knowledge Filtering}
\label{sec:knowledge_filtering}

We retain the above five concept-centric knowledge types in our experiments because they capture relatively stable semantic properties of real-world concepts. In contrast, we filter out the \textbf{Causality \& Event}, \textbf{Lexical/Etymological}, and \textbf{Other} categories, since they primarily encode event-level associations, linguistic form, or noisy relations that are less suitable for modeling concept-level knowledge.

Based on these filtered relations, we construct fictional concepts from real concepts. Following previous work~\cite{allen2023physics,how2025dynamics,ou-etal-2025-llms,zhu2025effective,feng2024extractive}, for each concept-relation pair, we enforce a single canonical object and remove competing candidates. For example, if \textit{dog CapableOf} is associated with multiple possible objects such as \textit{run} and \textit{guard your home}, we retain only one canonical object. This procedure yields a controlled and constrained concept-relation mapping, ensuring that fictional concepts are not merely renamed versions of existing real-world concepts.

\section{Dataset Statistics}\label{sec:statistics}

\begin{table}[t]

\centering
\begin{tabular}{l|c|c  }
\toprule
\textbf{Data} & \textbf{\tt Train} &  \textbf{\tt Test}  \\
 \midrule
\# Concepts  &1,000 & 500 \\ 
\# Knowledges  & 3,075 &1,586  \\ 
\# Samples & 92,250 &1,586 \\ 
\# Tokens & 1,039,784 & 13,530 \\ 
\bottomrule
\end{tabular}%
\caption{Dataset statistics of \dataset{}. }
\label{tab:statistic}
\vspace{-5mm}
\end{table}
As shown in Table~\ref{tab:statistic}, following dataset construction, we leverage 1,000 concepts for training and 500 concepts for testing. The training set contains 3,075 knowledge triples, which are instantiated into 92,250 training samples with different templates, comprising approximately 1.04M tokens in total. We train LLMs on our dataset for 10 epochs, leading to total training tokens of 10.4M, similar as previous continual-pretraining work~\cite{ou-etal-2025-llms}. The test set consists of 1,586 knowledge triples and corresponding evaluation samples, totaling 13,530 tokens.

\section{Circuit Discovery}\label{sec:circuit_discovery}
We instantiate concept circuits using EAP-IG~\citet{hanna2024have}, an efficient edge-level circuit discovery method that approximates activation patching with integrated gradients. We choose EAP-IG because our analysis requires extracting circuits for many concepts across multiple continual-pretraining checkpoints, making exhaustive causal intervention methods computationally prohibitive. Following prior circuit-discovery work, we view a circuit as a sparse computational subgraph whose standalone computation preserves the model behavior on a target task.

For each concept $c_i$, we construct a circuit-discovery set from its associated held-out knowledge triples $\{(c_i,r_{ij},o_{ij})\}_j$. Each clean input is the prefix generated from the subject--relation pair $(c_i,r_{ij})$, with the gold object $o_{ij}$ as the correct target. To construct a corrupted counterpart, we replace the gold object with an incorrect object sampled from another concept, yielding clean--corrupted pairs for attribution. EAP-IG assigns an importance score to each edge by integrating the gradient of the task loss along the path from corrupted activations to clean activations. The task metric is the logit difference between the correct and corrupted objects:
\[
M(x)=\mathrm{logit}(o_{ij}\mid x)-\mathrm{logit}(\tilde{o}_{ij}\mid x),
\]
where $\tilde{o}_{ij}$ denotes the corrupted object. Following EAP-IG, we use $L(x)=-M(x)$ as the attribution loss.

After scoring all edges, we rank them by the absolute EAP-IG score and select the smallest set of top-ranked edges whose resulting subgraph preserves at least $\tau$ of the full model's performance on the corresponding concept. Performance preservation is measured by the average logit-difference metric $M(x)$ over the concept's triples. In our main experiments, we set $\tau=70\%$, following prior circuit-level interpretability work~\cite{ou-etal-2025-llms,yao2025knowledgecircuitspretrainedtransformers} as a practical faithfulness--sparsity trade-off. Lower thresholds may omit important edges, while higher thresholds yield denser circuits that are less suitable for topology analysis. To verify that our conclusions are not tied to this heuristic, we additionally evaluate $\tau=80\%$ and observe similar circuit--behavior correlations, as shown in Table~\ref{tab:baseline}.

\section{Robustness Under Longer or Repeated Continual Training}\label{sec:longer_continual_training}

To assess whether our findings persist beyond the two-stage continual pretraining setup in the main paper, we conduct an additional \textbf{repeated continual training} experiment with GPT-2 Large. Specifically, after obtaining checkpoint $\pi_2$ from Stage 2 (Forgetting Induction), as discussed in Section~\ref{sec:experiment_design}, we continue training the model for two more stages using the same protocol: we further train $\pi_2$ on \dataset{} dataset, followed by BIO dataset, resulting in a \textbf{four-stage continual training setup}. Across this extended setting, we observe that the main trends remain qualitatively consistent: concept learning and forgetting continue to show substantial cross-concept variability, maintain non-trivial correlations with circuit graph metrics, and exhibit a positive association between stronger learning and greater forgetting. These results suggest that our findings are robust beyond a single two-stage continual training cycle.

\begin{figure}[htbp]
	\centering
	\includegraphics[width=1\linewidth]{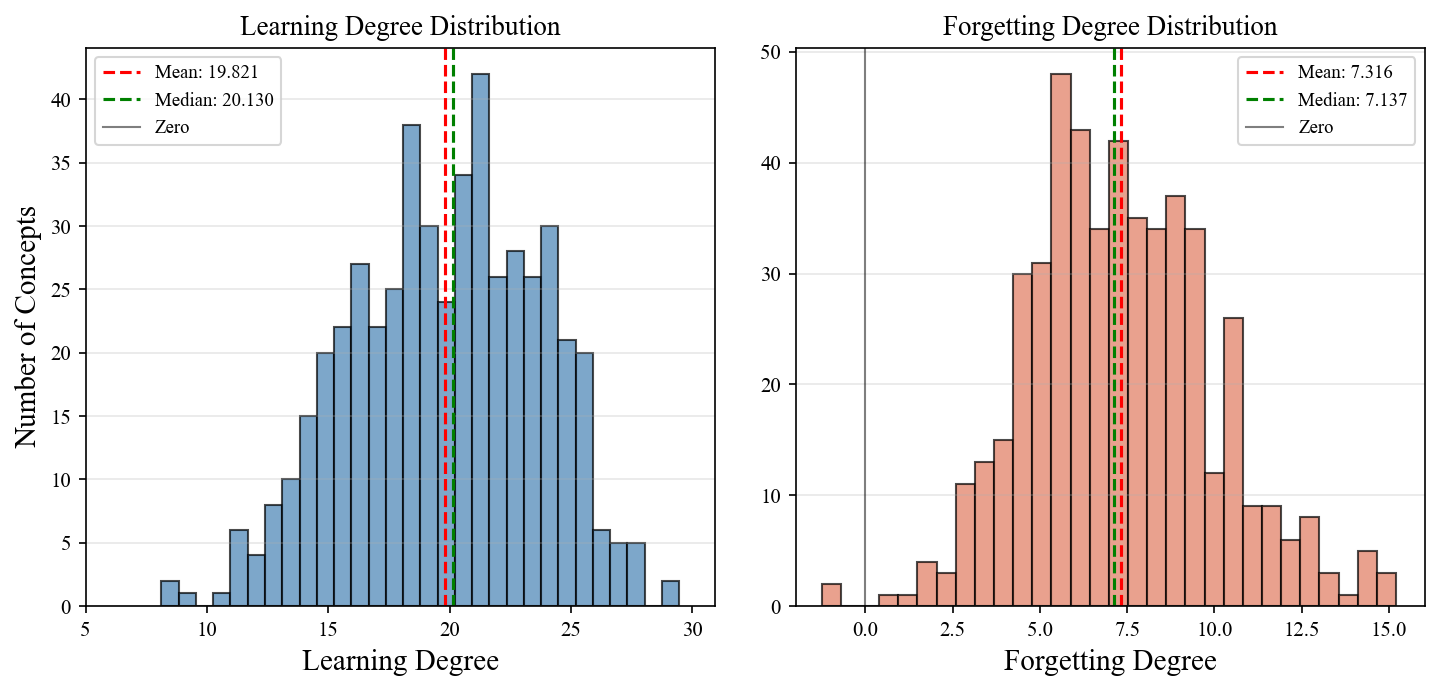}
	\caption{Distribution of learning and forgetting degree across concepts for GPT-2 Large under four-stage continual training.}
	\label{fig:concept_learn_hist_continual_pretrained_gpt2}
\end{figure}

\begin{figure}[t]
    \centering
    \includegraphics[width=1\linewidth]{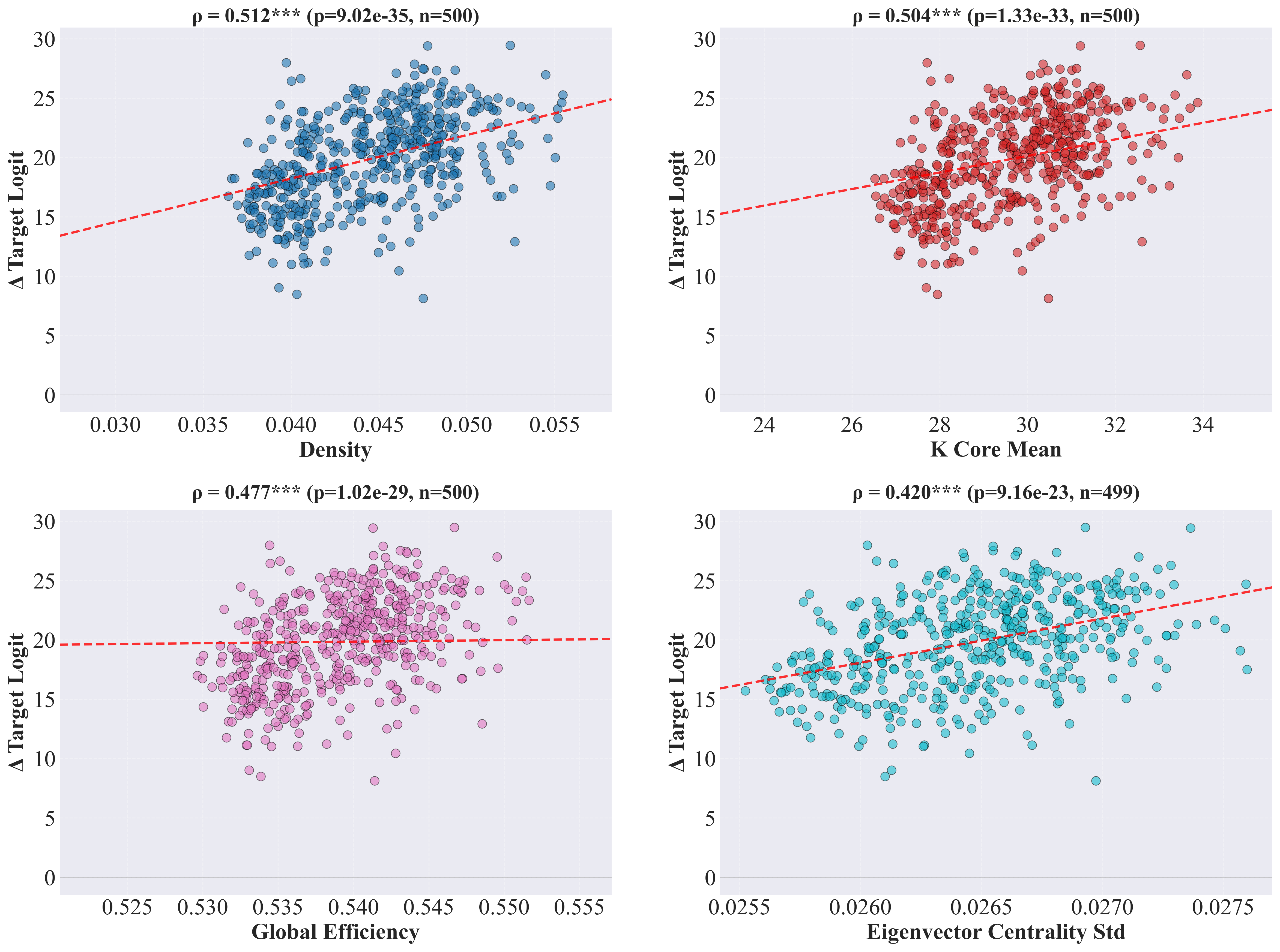}
    \caption{Correlation between learning degree and LLM circuit graph scores. }
    \label{fig:learn_correlation_final_continual_pretrained_gpt2}
\end{figure}

\begin{figure}[htbp]
	\centering
	\includegraphics[width=1\linewidth]{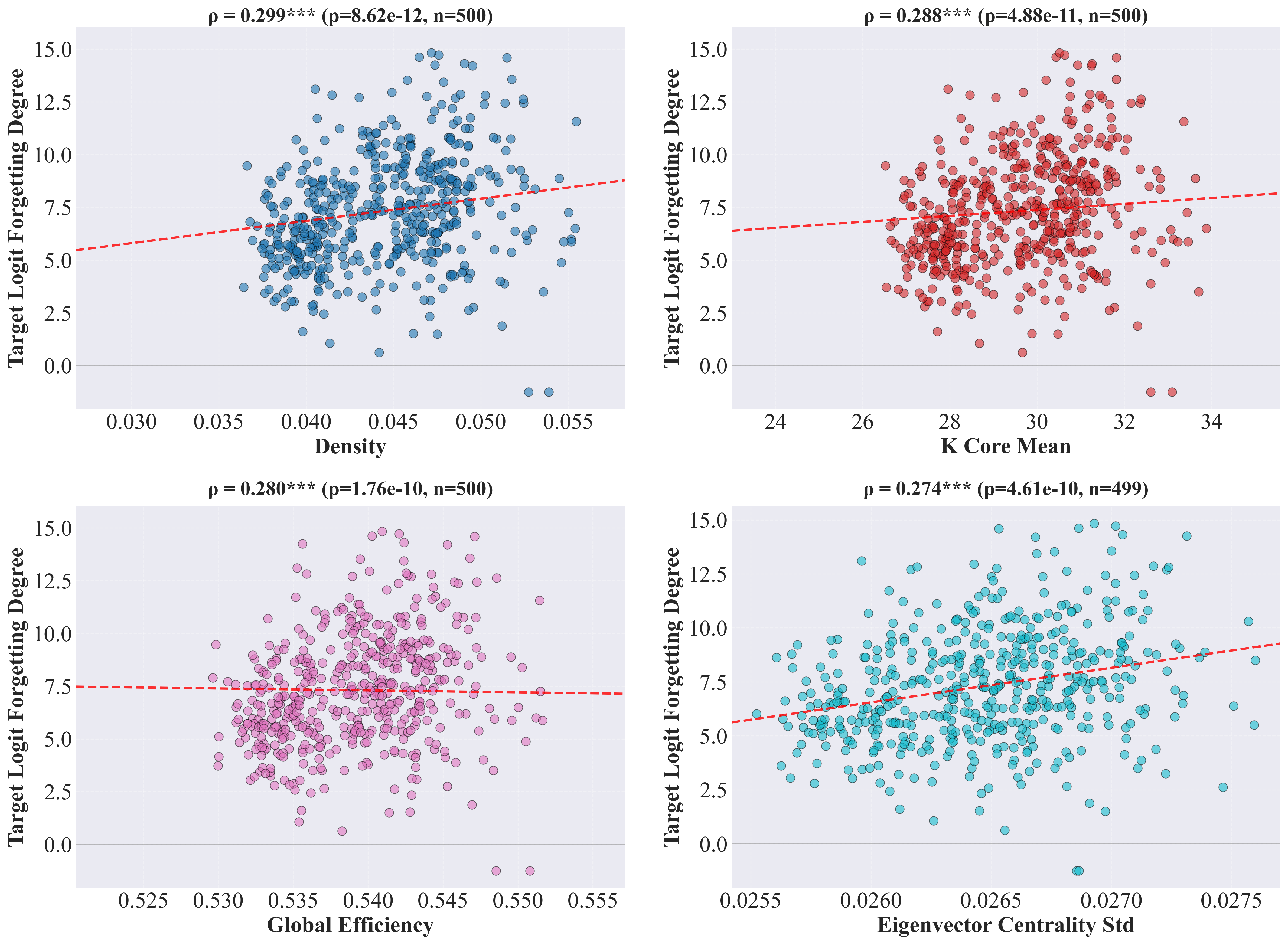}
\caption{Correlation between forgetting degree and LLM circuit graph scores.}
	\label{fig:forget_correlation_continual_pretrained_gpt2}
\end{figure}

\begin{figure}[htbp]
	\centering
	\includegraphics[width=1\linewidth]{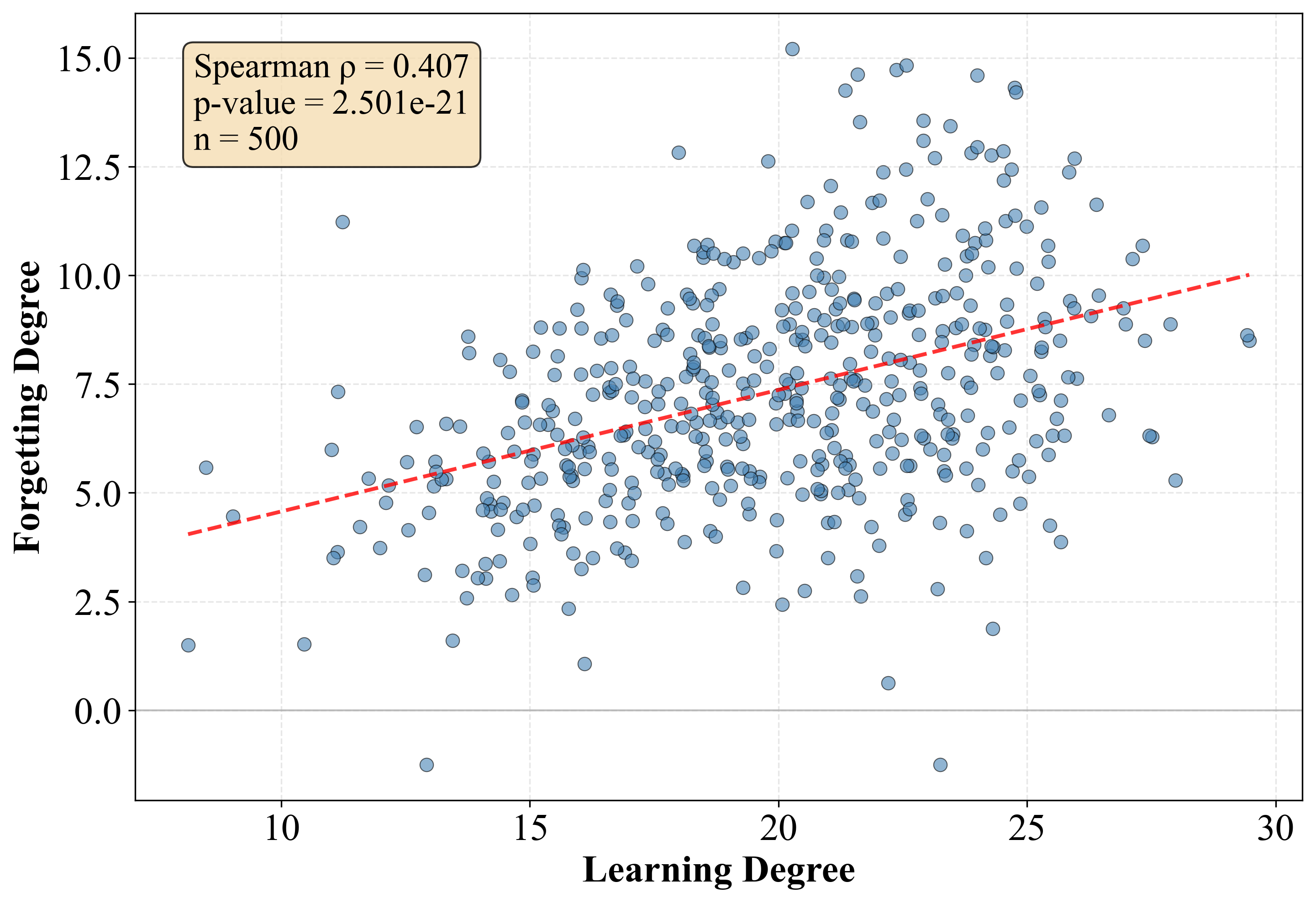}
	\caption{Spearman Correlations between learning and forgetting of concepts}
	\label{fig:concept_corr_continual_pretrained_gpt2}
\end{figure}

We show results in Figure~\ref{fig:concept_learn_hist_continual_pretrained_gpt2}, Figure~\ref{fig:learn_correlation_final_continual_pretrained_gpt2}, Figure~\ref{fig:forget_correlation_continual_pretrained_gpt2}, and Figure~\ref{fig:concept_corr_continual_pretrained_gpt2}.

\section{LLaMA and Qwen Results}\label{sec:llama}

\begin{table}[t]
\centering
\resizebox{\linewidth}{!}{
\setlength{\tabcolsep}{6pt}
\renewcommand{\arraystretch}{0.92}
\begin{tabular}{l l c c c c}
\toprule
 Model & Metric & Learning & Learning $p$& Forgetting& Forgetting $p$ \\
 \midrule
Qwen3-4B-Base & Density & 0.333 & 2.1e-14 & 0.416  &2.5e-22 \\
Qwen3-4B-Base & $k$-core & 0.319 & 2.7e-13 & 0.409   &1.4e-21\\
Qwen3-4B-Base& Eigenvector Centrality & 0.328 &5.2e-14  & 0.394  &4.8e-20 \\
Qwen3-4B-Base& Global Efficiency & 0.285  & 8e-11 & 0.304   &3.7e-12\\
\midrule
 LLaMA-3.2-1B & Density & 0.399 & 2.2e-20 & 0.411 & 1.1e-21 \\
LLaMA-3.2-1B  & $k$-core & 0.375 & 5.2e-18 & 0.381 & 1.5e-18 \\
LLaMA-3.2-1B & Eigenvector Centrality & 0.163 & 2.6e-4 & 0.175 & 9.1e-5  \\
 LLaMA-3.2-1B & Global Efficiency & 0.332 & 3.4e-14 & 0.372 & 9.2e-18  \\
\midrule
 LLaMA-3.1-8B & Density & 0.349 & 1.1e-15 & 0.386 & 4.1e-19 \\
LLaMA-3.1-8B  & $k$-core & 0.349 & 1.1e-15 & 0.389 & 2.1e-19 \\
LLaMA-3.1-8B & Eigenvector Centrality & 0.148 & 9.4e-4 & 0.182 & 4.4e-5  \\
 LLaMA-3.1-8B & Global Efficiency & 0.201 & 6.4e-6 & 0.207 & 3.5e-6  \\

\bottomrule
\end{tabular}
}
\caption{Complete correlations and $p$-values for Table~\ref{tab:correlation}: Spearman correlation between learning/forgetting degree and circuit topology metrics on LLaMA-3.2-1B, LLaMA-3.1-8B, and Qwen3-4B-Base.}
\label{tab:correlation_pvalues}
\vspace{-0.8em}
\end{table}

Table~\ref{tab:correlation_pvalues} reports the complete cross-model correlations with $p$-values corresponding to Table~\ref{tab:correlation} in the main paper.

\subsection{LLaMA-3.2-1B Results}

\subsubsection{RQ1 Results}

\begin{figure}[htbp]
	\includegraphics[width=1\linewidth]{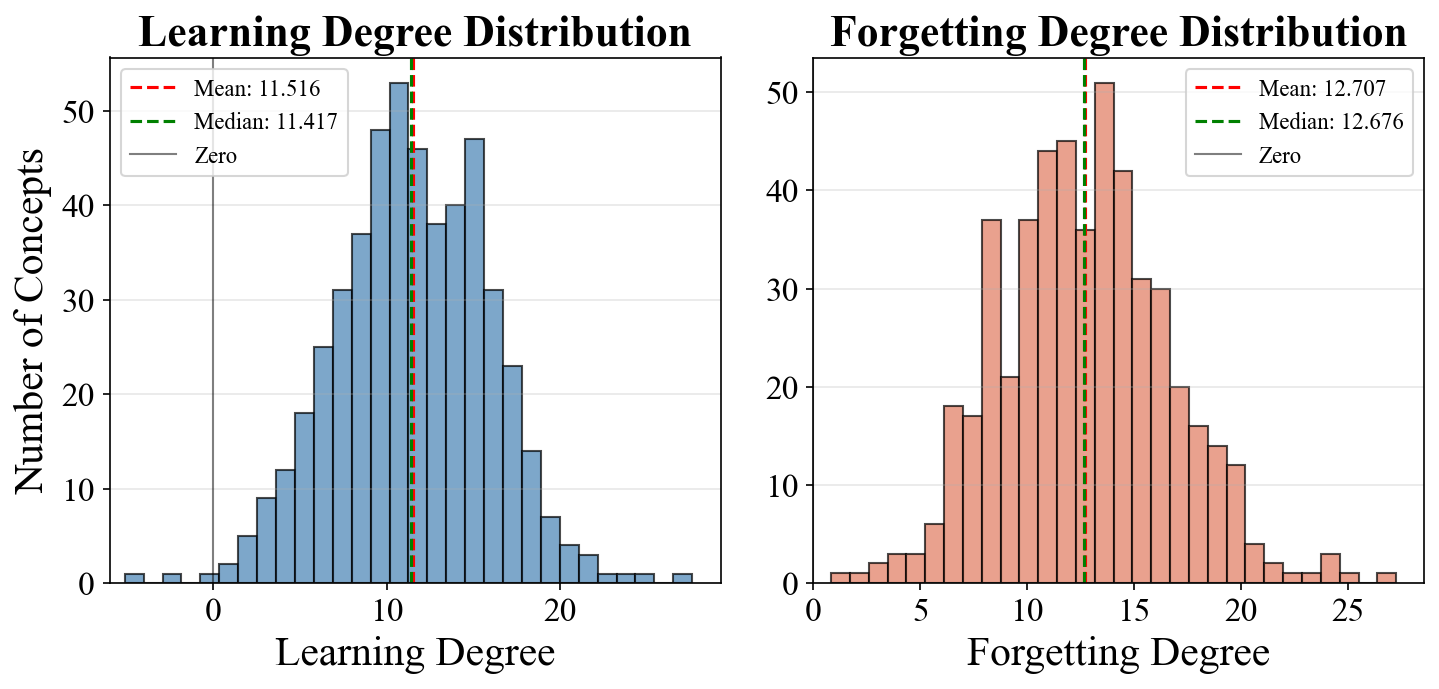}
	\caption{Distribution of learning and forgetting degree across concepts for LLaMA.}
	\label{fig:concept_learn_hist_llama}
\end{figure}

\begin{figure}[t]
    \centering
    \includegraphics[width=1\linewidth]{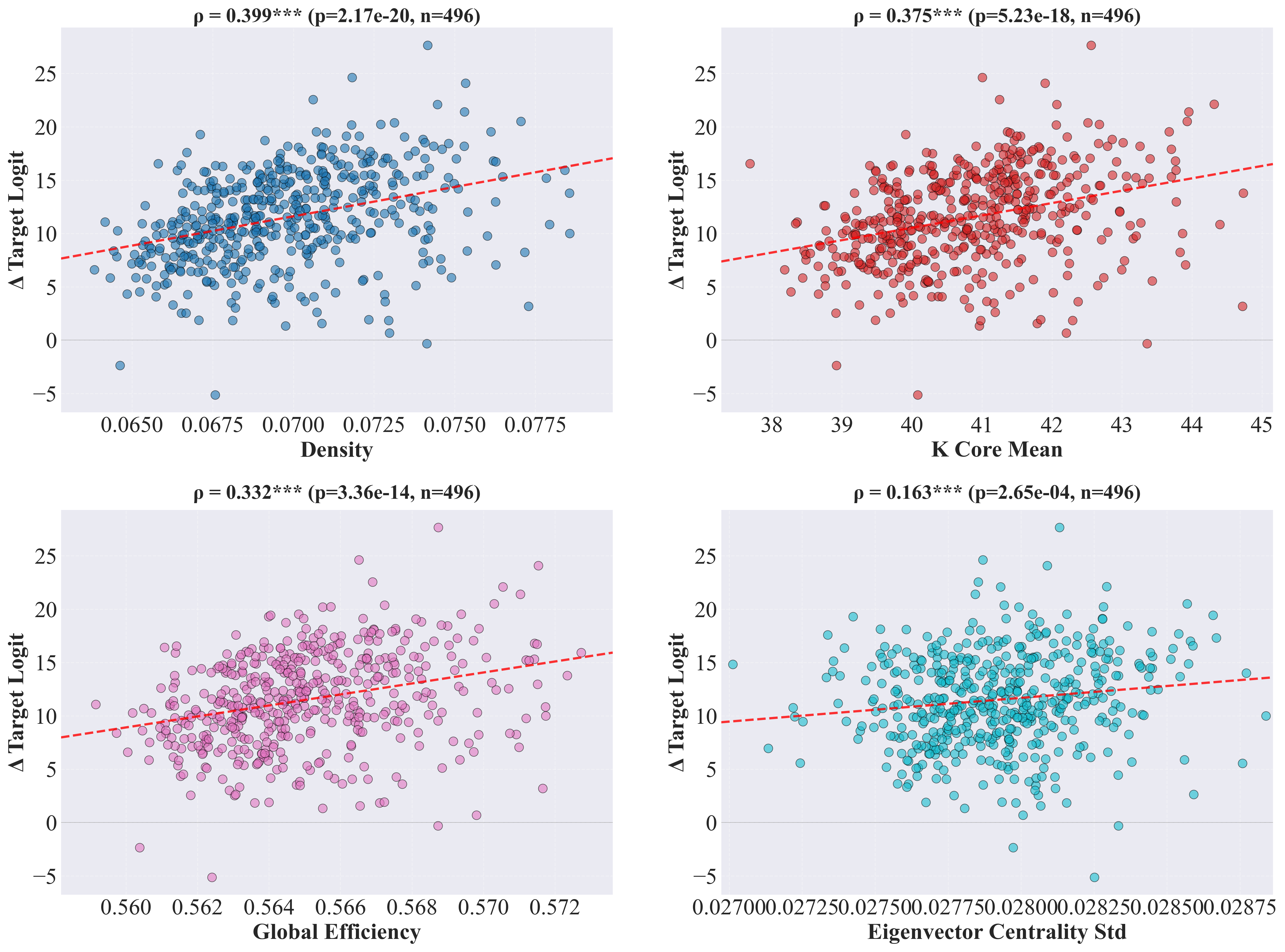}
    \caption{Correlation between learning degree and LLM circuit graph scores. }
    \label{fig:learn_correlation_final_llama}
\end{figure}

\begin{figure}[htbp]
	\centering
	\includegraphics[width=1\linewidth]{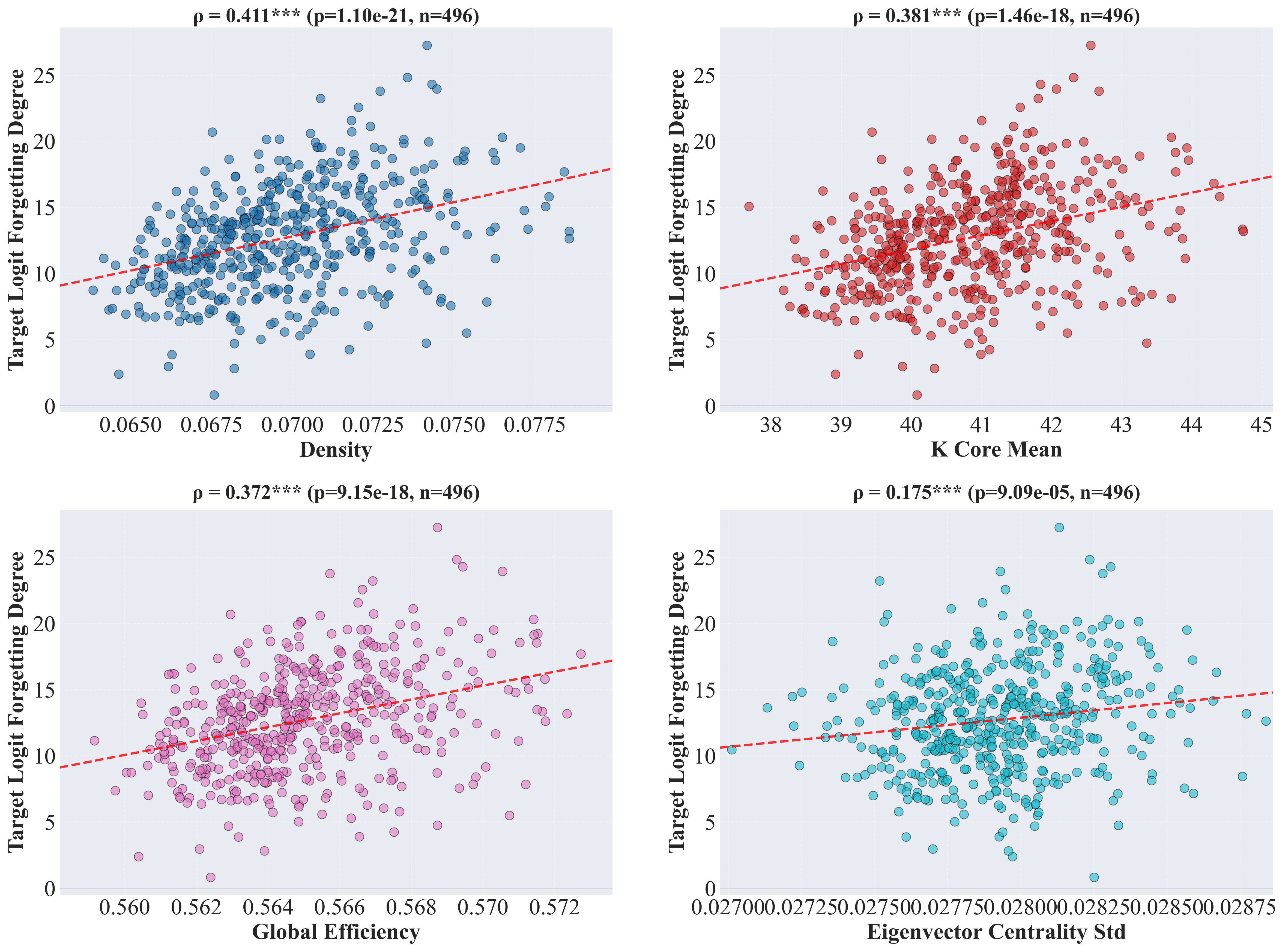}
\caption{Correlation between forgetting degree and LLM circuit graph scores.}
	\label{fig:forget_correlation_llama}
\end{figure}

\begin{figure}[htbp]
	\centering
	\includegraphics[width=1\linewidth]{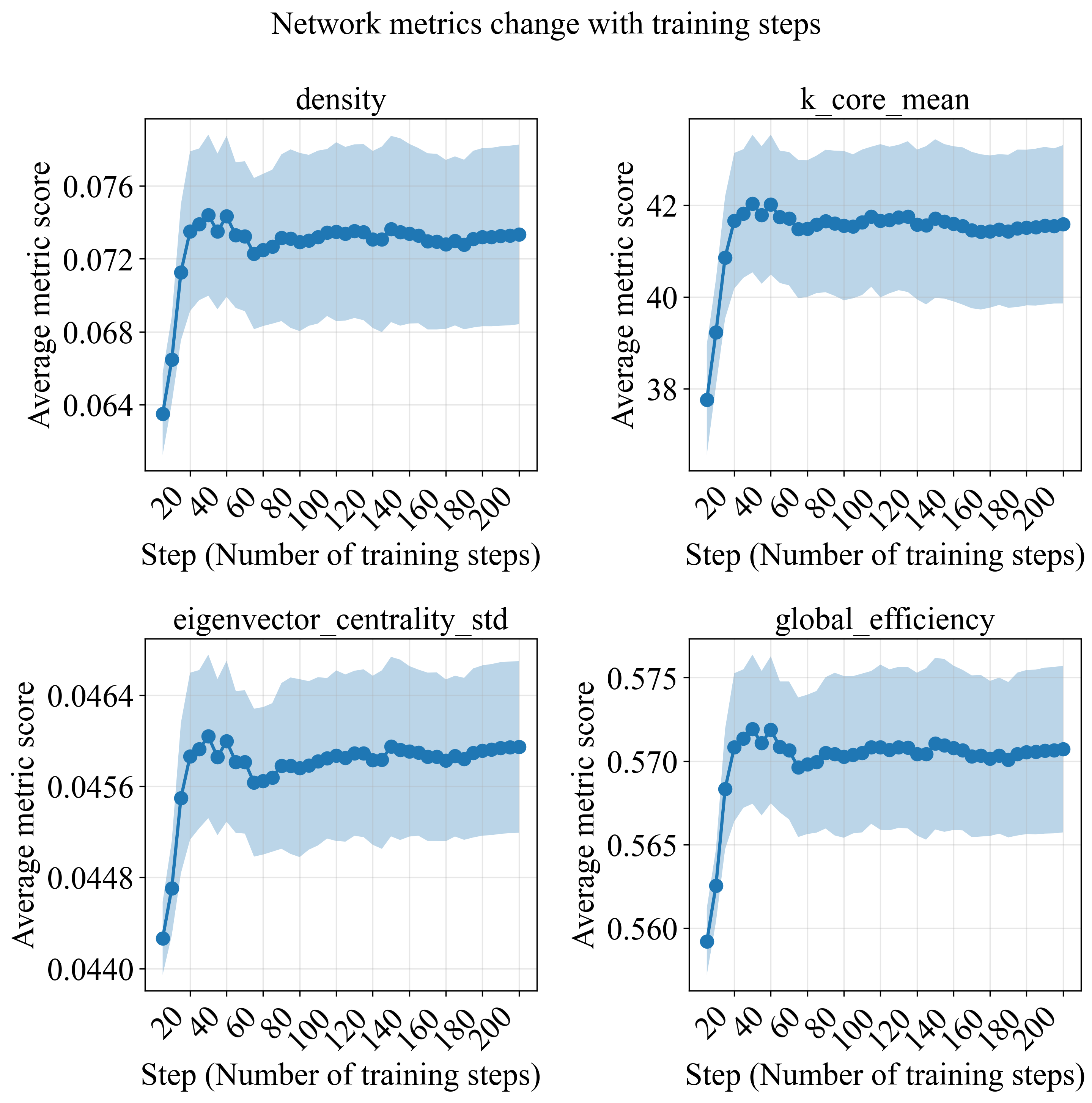}
	\caption{Graph Metric over training steps}
	\label{fig:rq2_forget_over_steps_llama}
\end{figure}

\begin{figure}[htbp]
	\centering
	\includegraphics[width=1\linewidth]{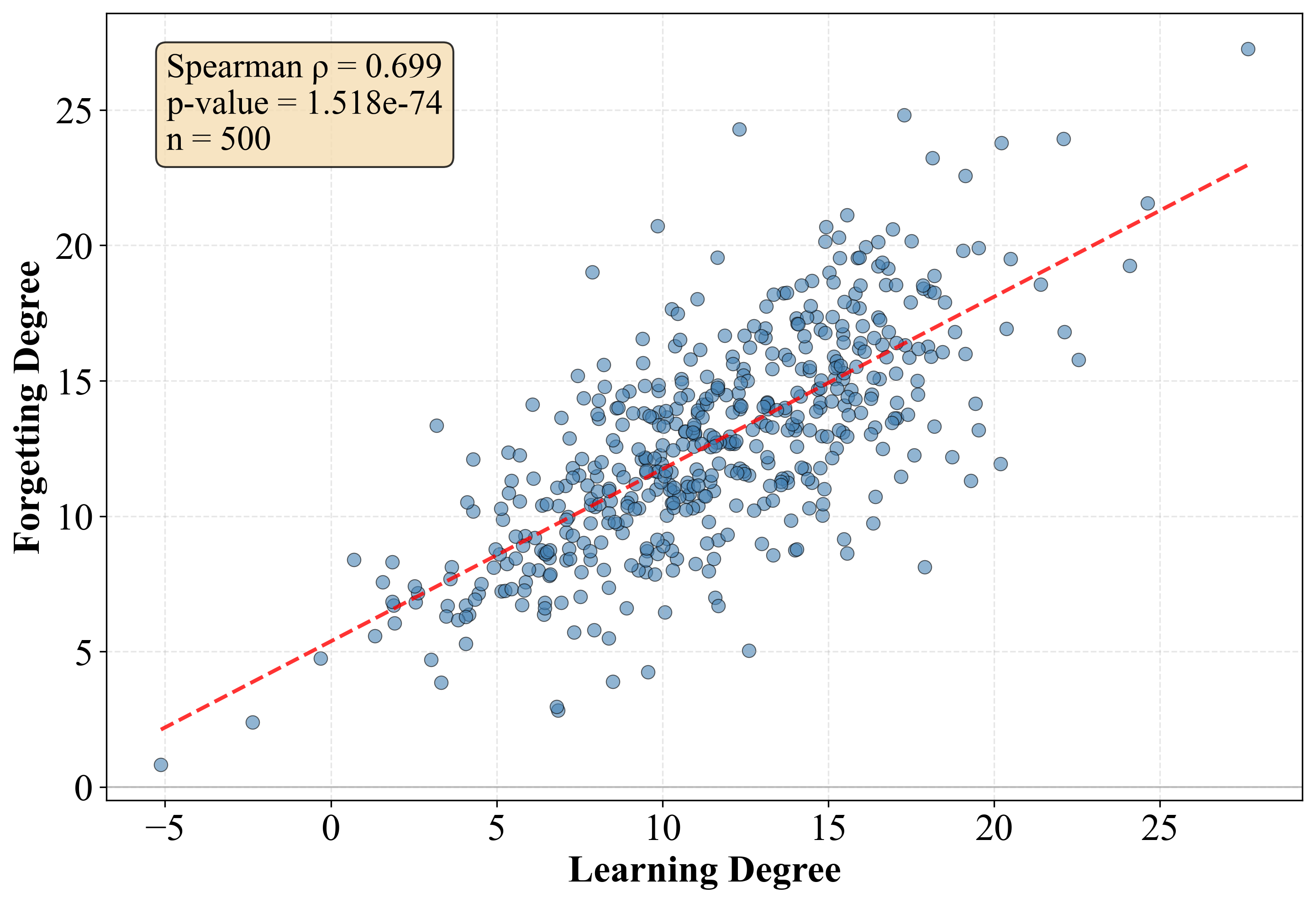}
	\caption{Correlations between learning and forgetting of concepts}
	\label{fig:concept_corr_llama}
\end{figure}

Figure~\ref{fig:concept_learn_hist_llama} shows the distribution of concept learning and forgetting degree for LLaMA-3.2-1B, again revealing substantial variability across concepts. Figure~\ref{fig:learn_correlation_final_llama} and Figure~\ref{fig:forget_correlation_llama} show that circuit topology metrics remain correlated with concept learning and forgetting degree, respectively, reproducing the same qualitative trend observed for GPT-2 Large in the main paper; the strongest metric, density, reaches $\rho\approx0.40$ for learning and $\rho\approx0.41$ for forgetting (Table~\ref{tab:correlation}). Figure~\ref{fig:rq2_forget_over_steps_llama} further shows the temporal evolution of circuit topology during continual forgetting, exhibiting a similar pattern of early increase followed by later decrease and stabilization. Finally, Figure~\ref{fig:concept_corr_llama} shows a positive correlation between learning degree and forgetting degree, confirming that concepts learned more strongly also tend to be forgotten more strongly under subsequent training.

\subsubsection{RQ2 Results}
\begin{figure}[htbp]
	\centering
	\includegraphics[width=1\linewidth]{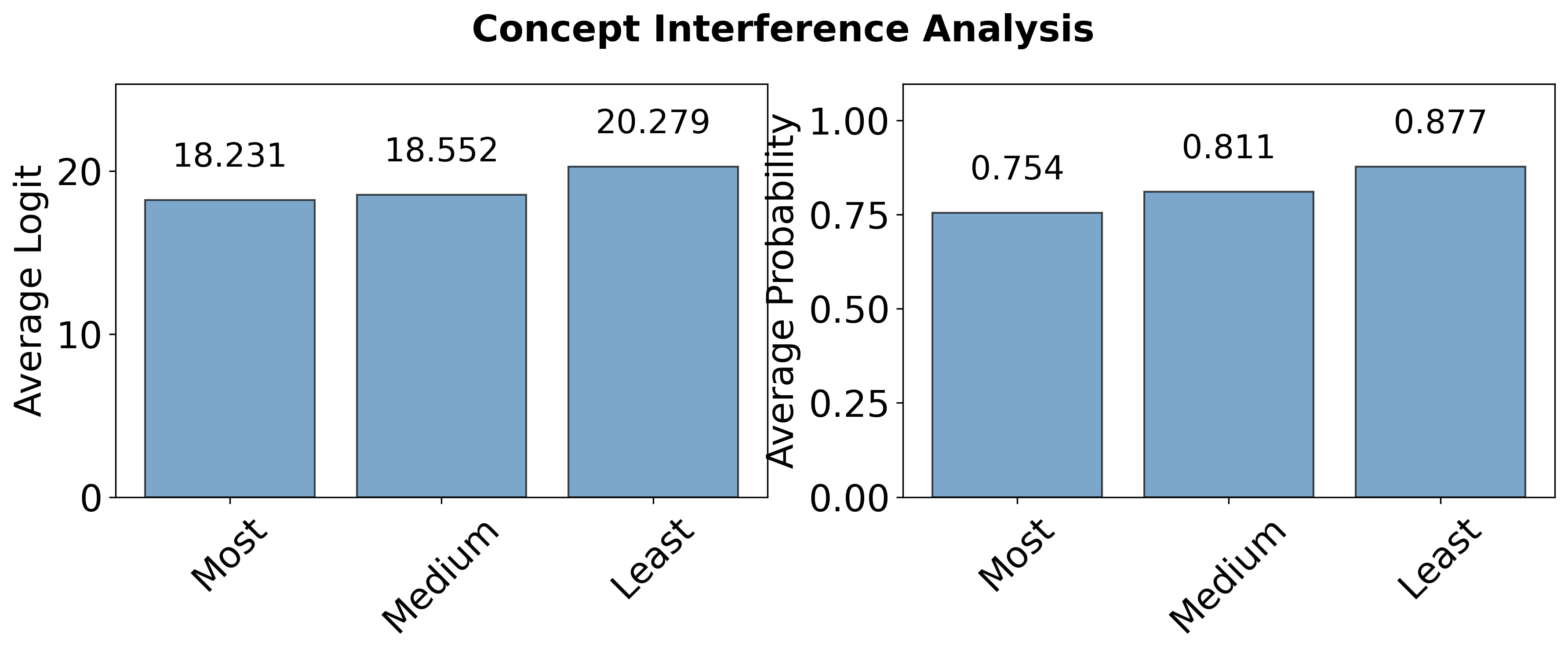}
	\caption{Concept-level interference under joint training.}
	\label{fig:concept_interference_100_llama}
\end{figure}

\begin{figure}[htbp]
	\centering
	\includegraphics[width=1\linewidth]{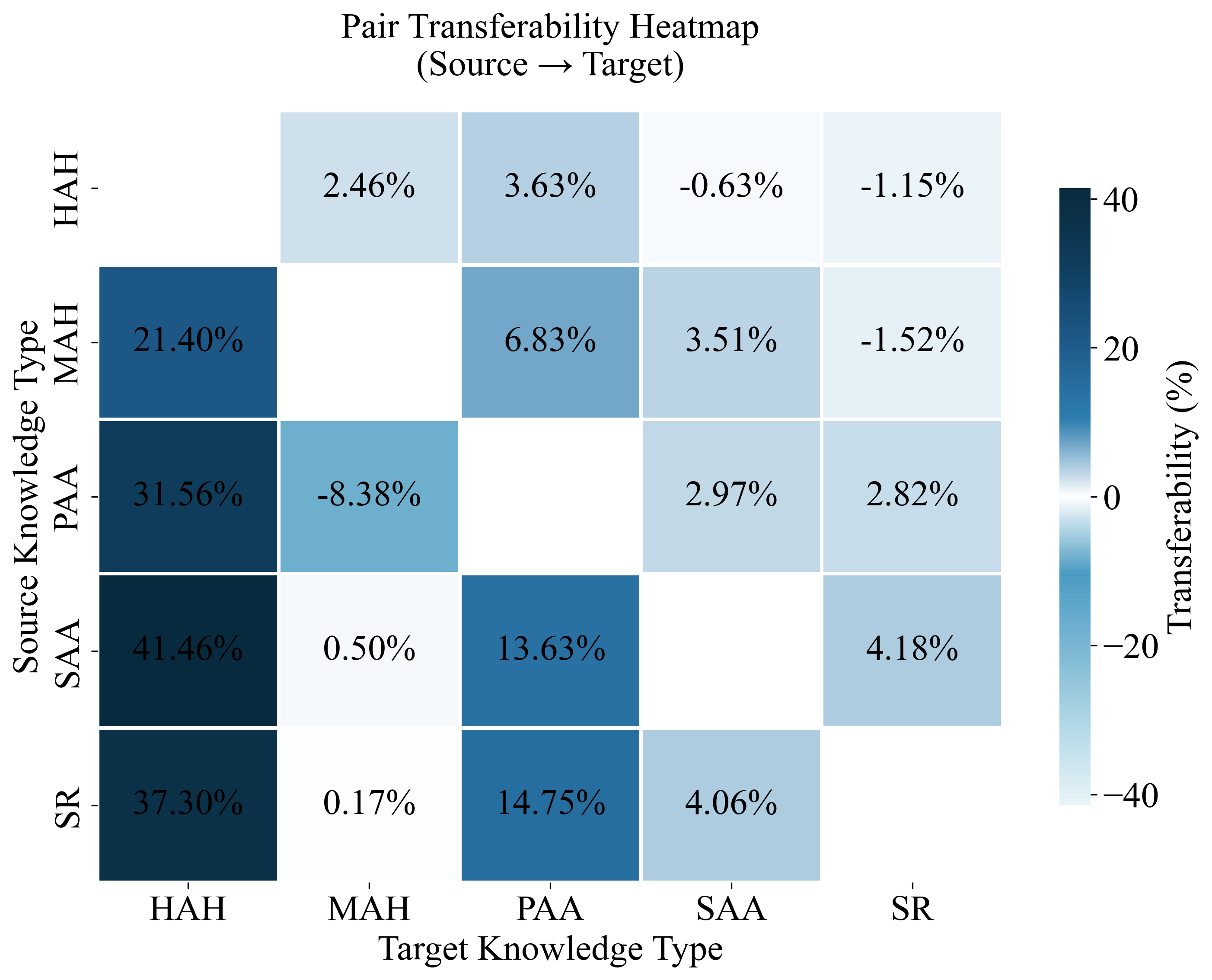}
	\caption{Pairwise dependency between knowledge types}
	\label{fig:knowledge_dependency_llama}
\end{figure}
Figure~\ref{fig:concept_interference_100_llama} shows concept interference under joint training for LLaMA-3.2-1B, where semantically related concepts induce stronger interference than weakly related concepts. Figure~\ref{fig:knowledge_dependency_llama} further shows that transferability across knowledge types remains highly heterogeneous and directional, indicating that concept synergy and interference patterns generalize beyond GPT-2 Large.

\subsection{LLaMA-3.1-8B Results}

\subsubsection{RQ1 Results} 
\begin{figure}[htbp]
	\centering
	\includegraphics[width=1\linewidth]{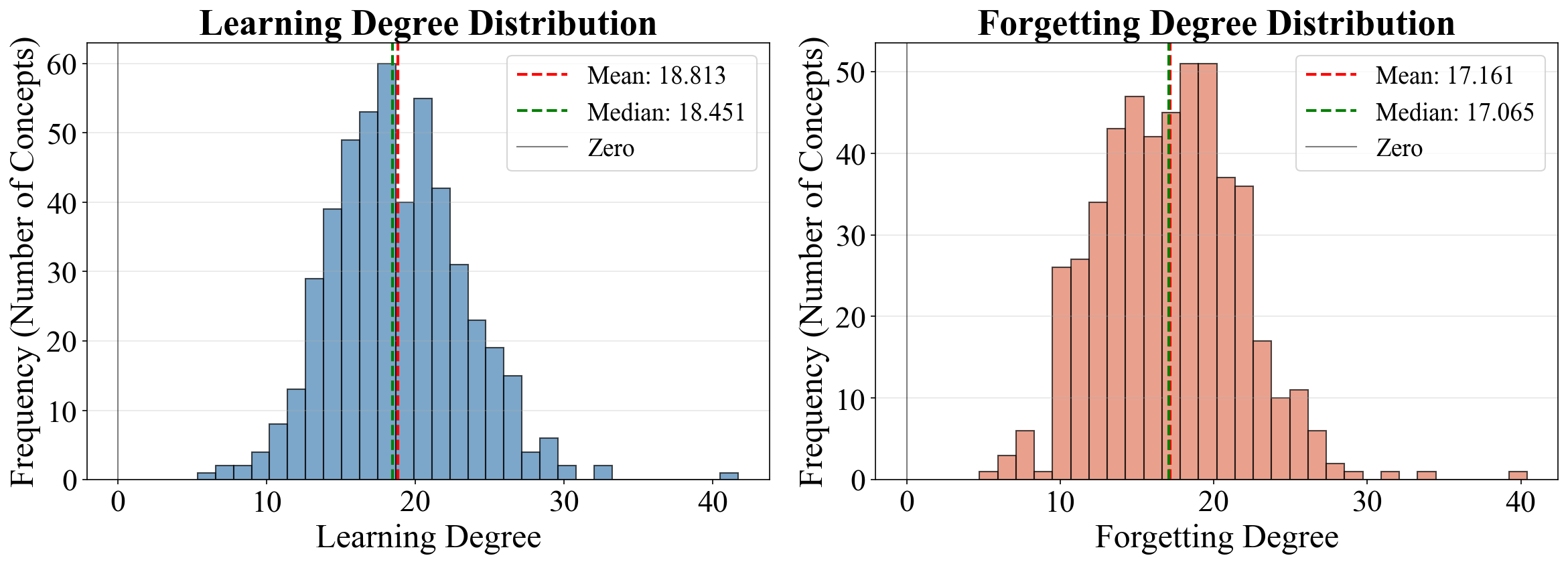}
	\caption{Distribution of learning and forgetting degree across concepts for LLaMA.}
	\label{fig:concept_learn_hist_llama_8b}
\end{figure}

\begin{figure}[t]
    \centering
    \includegraphics[width=1\linewidth]{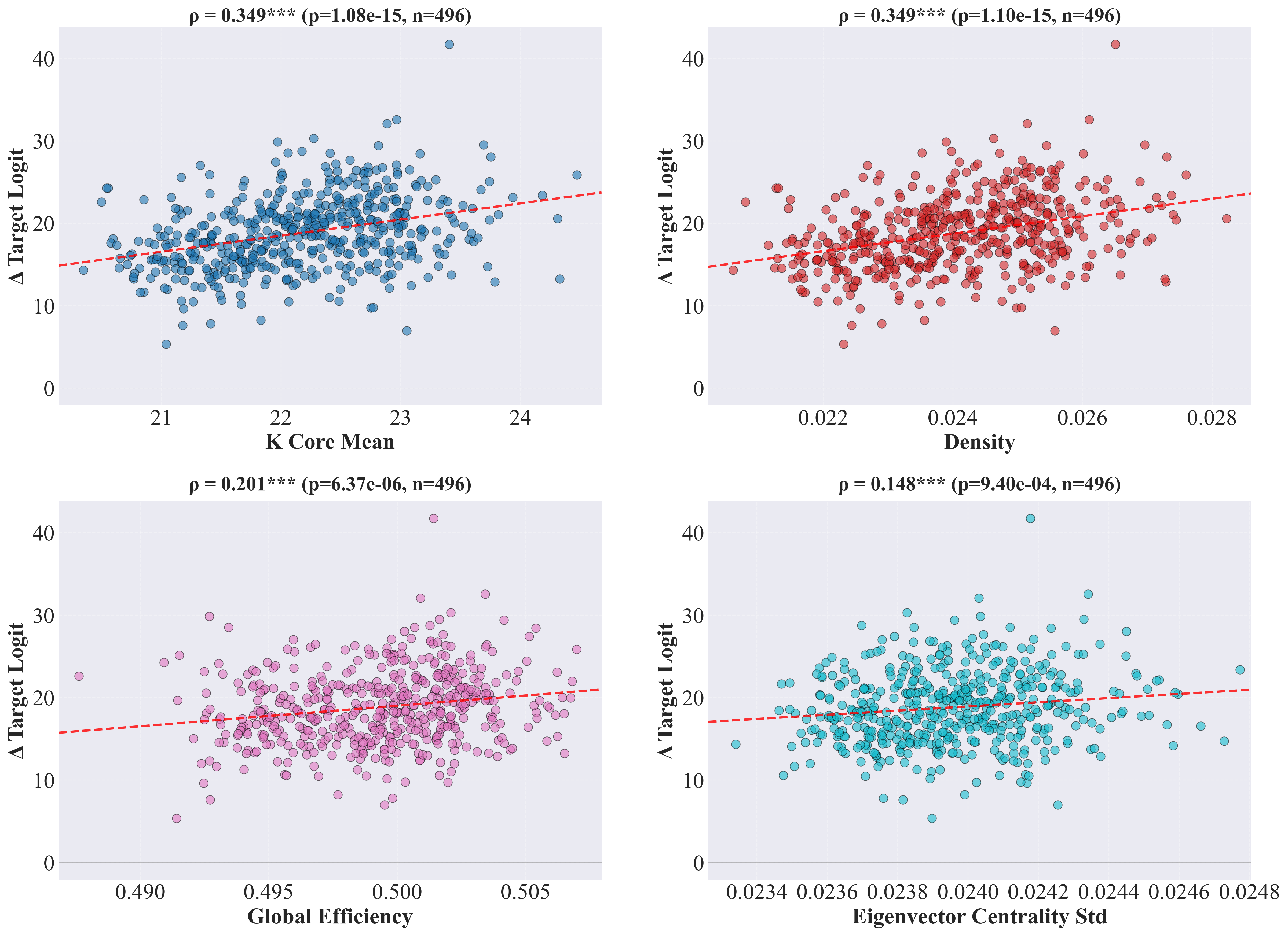}
    \caption{Correlation between learning degree and LLM circuit graph scores. }
    \label{fig:learn_correlation_final_llama_8b}
\end{figure}

\begin{figure}[htbp]
	\centering
	\includegraphics[width=1\linewidth]{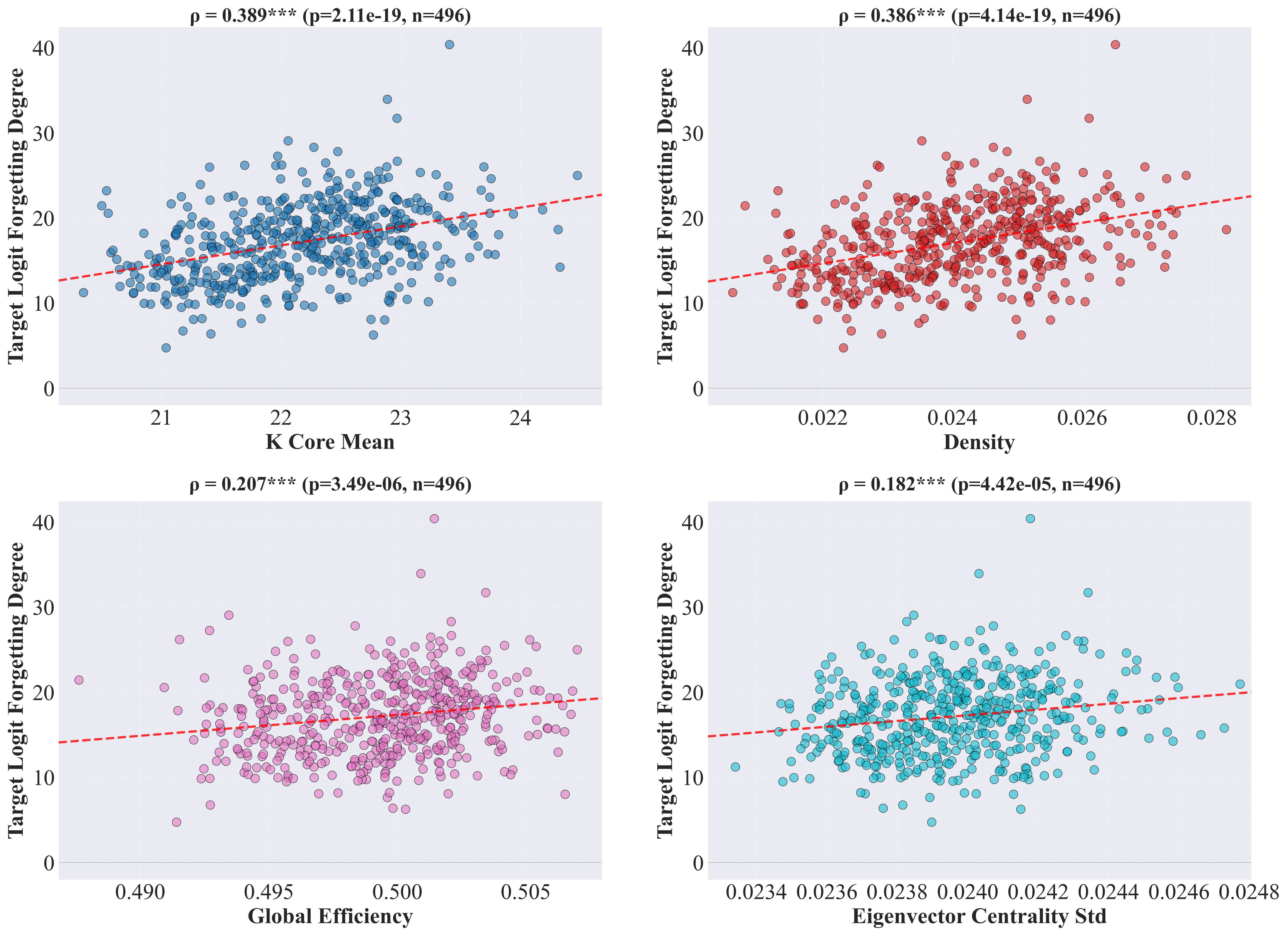}
\caption{Correlation between forgetting degree and LLM circuit graph scores.}
	\label{fig:forget_correlation_llama_8b}
\end{figure}
Figure~\ref{fig:concept_learn_hist_llama_8b} shows the distribution of concept learning and forgetting degree for LLaMA-3.1-8B, again indicating substantial cross-concept variability. Figure~\ref{fig:learn_correlation_final_llama_8b} and Figure~\ref{fig:forget_correlation_llama_8b} show that circuit topology metrics remain meaningfully associated with concept learning and forgetting degree, respectively, reproducing the same overall trend observed in smaller models and in the GPT-2 Large results; the strongest metrics, density and $k$-core, reach $\rho\approx0.35$ for learning and $\rho\approx0.39$ for forgetting (Table~\ref{tab:correlation}).

\subsubsection{RQ2 Results}
 
\begin{figure}[htbp]
	\centering
	\includegraphics[width=1\linewidth]{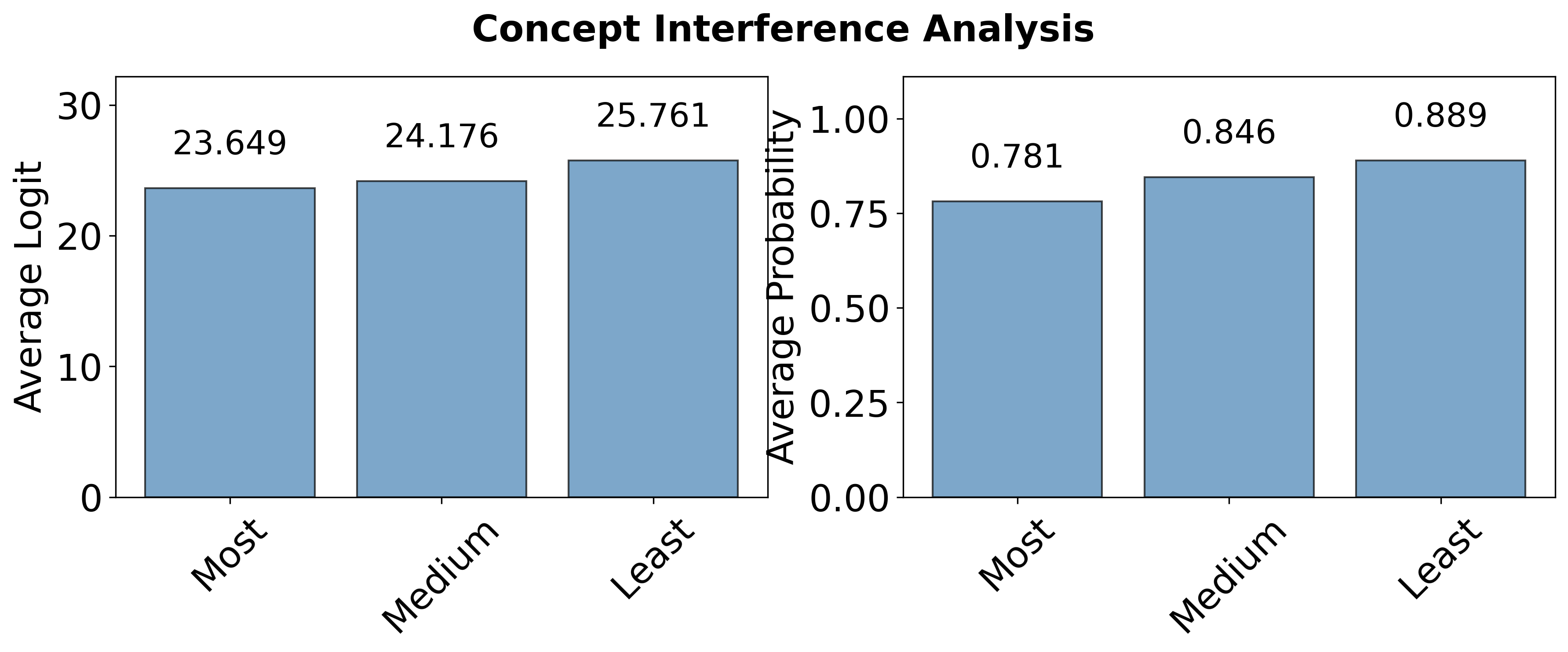}
	\caption{Concept-level interference under joint training on LLaMA-3.1-8B with sharpened top-20 co-training groups (final-step evaluation).}
	\label{fig:concept_interference_top20_llama_8b}
\end{figure}

\begin{figure}[htbp]
	\centering
	\includegraphics[width=1\linewidth]{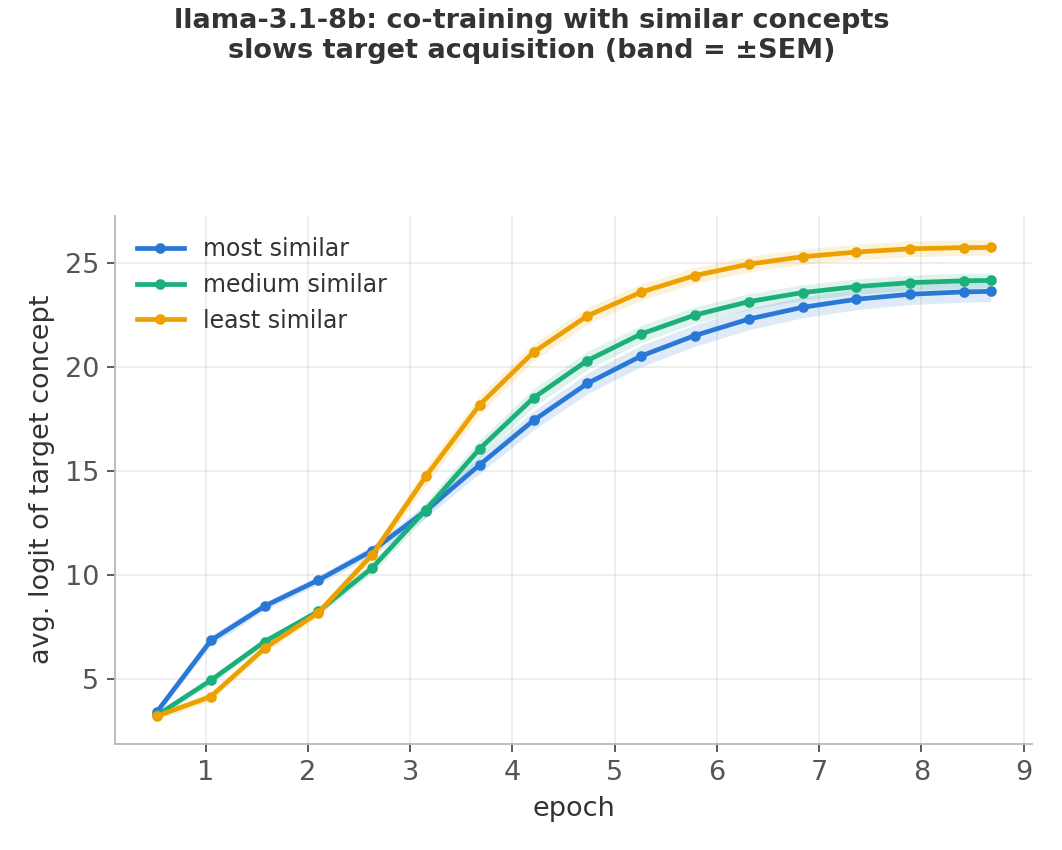}
	\caption{Acquisition curves (average logit of the target concept) for LLaMA-3.1-8B under the three co-training conditions (band $=\pm$SEM). The x-axis shows training epochs over the co-training corpus. The interference gap emerges early in acquisition, peaks mid-training, and narrows gradually as all conditions approach the performance ceiling.}
	\label{fig:concept_interference_curves_llama_8b}
\end{figure}

\begin{figure}[htbp]
	\centering
	\includegraphics[width=1\linewidth]{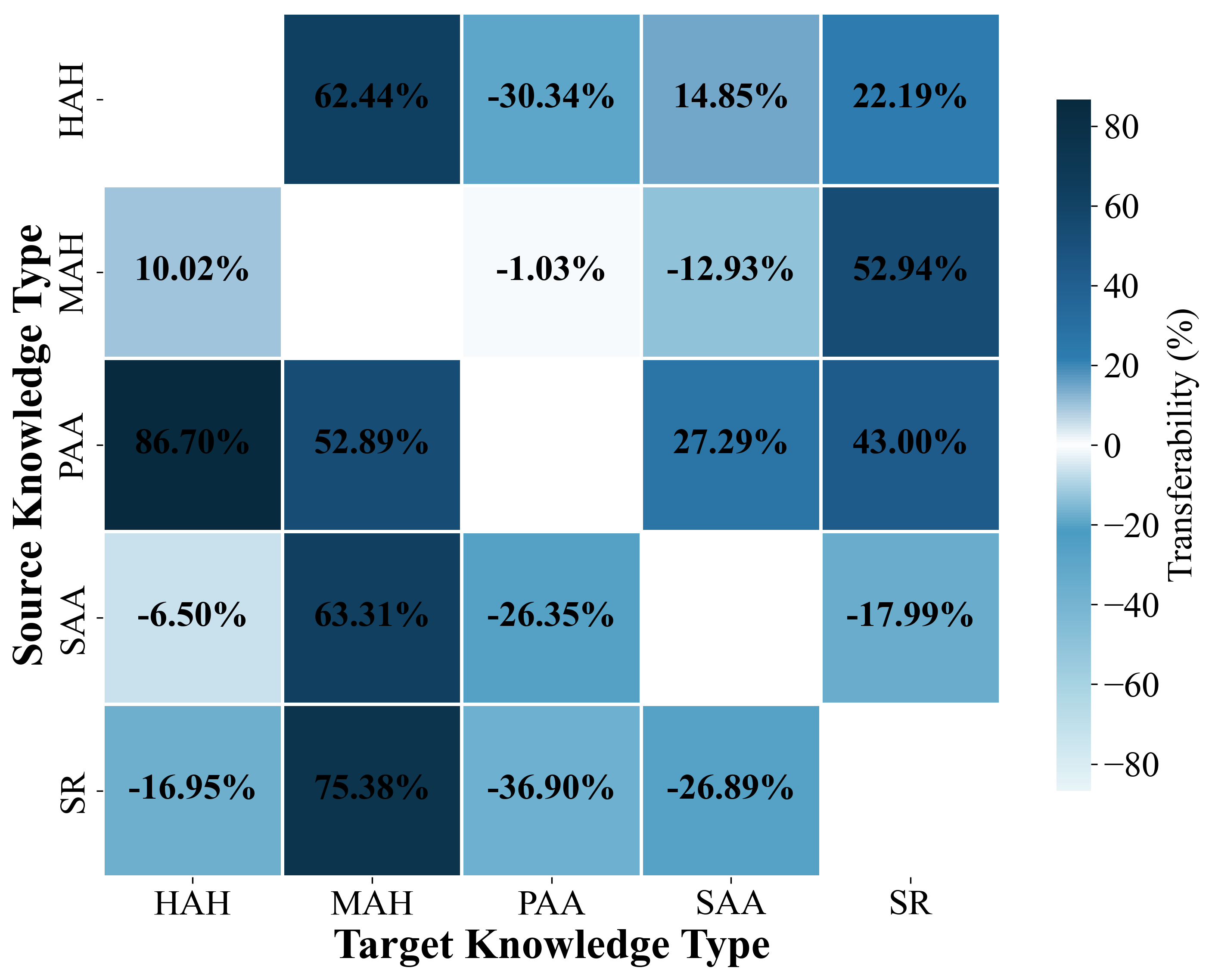}
	\caption{Pairwise dependency between knowledge types}
	\label{fig:knowledge_dependency_llama_8b}
\end{figure}

Since LLaMA-3.1-8B is a more powerful model, the original interference setting saturates: with $K=100$ co-training groups, the most, medium, and least similar co-training groups reach 85.1\%, 85.8\%, and 86.7\% average target probability (average target logits of 23.55, 23.94, and 24.59), respectively, a \emph{ceiling effect} in which the task is too easy for a model of this capacity to separate the conditions, not evidence that interference disappears at scale. We therefore leverage a sharpened top-20 setting for LLaMA-3.1-8B: each of the same target concepts is co-trained with its top-20 / middle-20 / bottom-20 most-similar concepts, upsampled so that the training-set size is unchanged (with the same batch size as the original experiment). The final results are shown in Figure~\ref{fig:concept_interference_top20_llama_8b}, recovering the clear interference ordering. To make the mechanism explicit, we further trace full acquisition curves, evaluating each target's held-out samples throughout training (Figure~\ref{fig:concept_interference_curves_llama_8b}): the gap emerges early in acquisition and peaks mid-training, then narrows gradually as all conditions approach the ceiling. At the 8B scale, interference among semantically similar concepts thus manifests as \emph{slower acquisition} rather than significantly lower final attainment. Figure~\ref{fig:knowledge_dependency_llama_8b} further shows that transferability across knowledge types remains asymmetric and uneven, suggesting that concept-level synergy and interference patterns are robust across model scale.

\subsection{Qwen3-4B-Base Results}\label{sec:qwen}

\subsubsection{RQ1 Results}

\begin{figure}[htbp]
	\centering
	\includegraphics[width=1\linewidth]{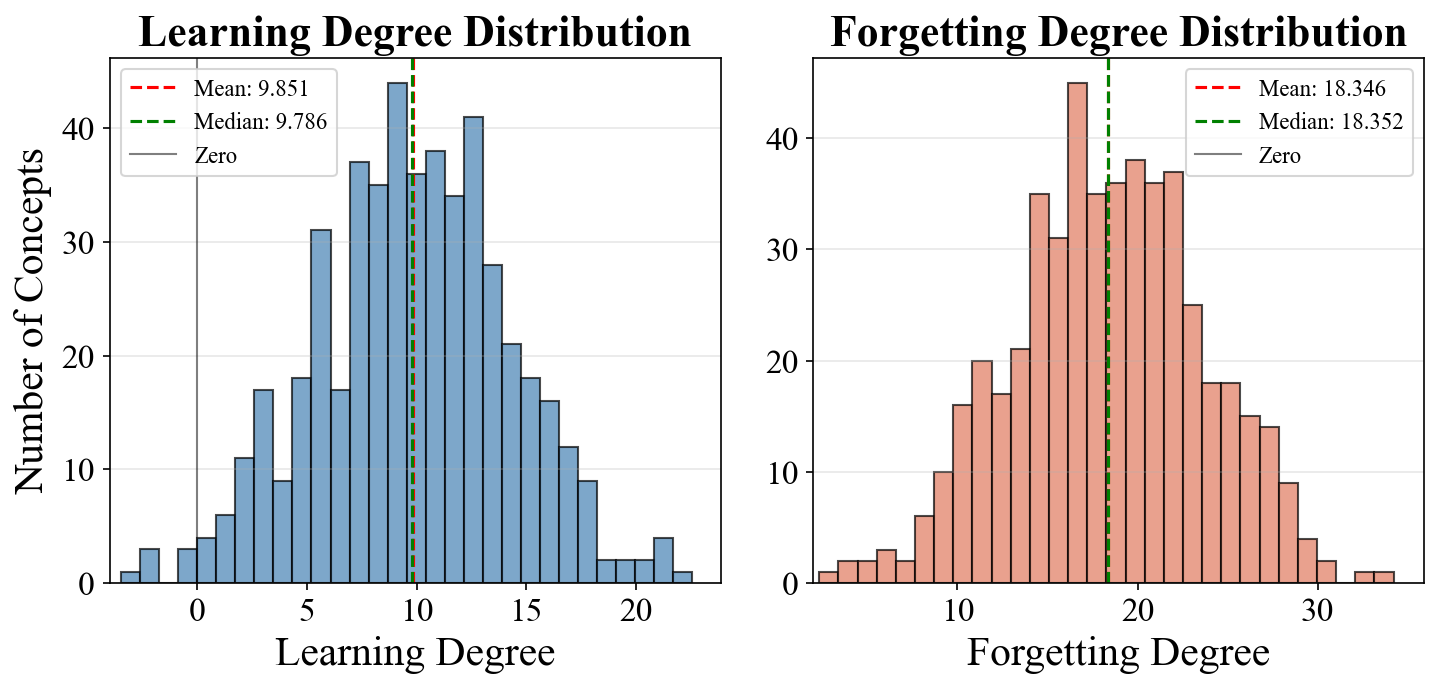}
	\caption{Distribution of learning and forgetting degree across concepts for Qwen3-4B-Base.}
	\label{fig:concept_learn_hist_qwen}
\end{figure}

\begin{figure}[htbp]
	\centering
	\includegraphics[width=1\linewidth]{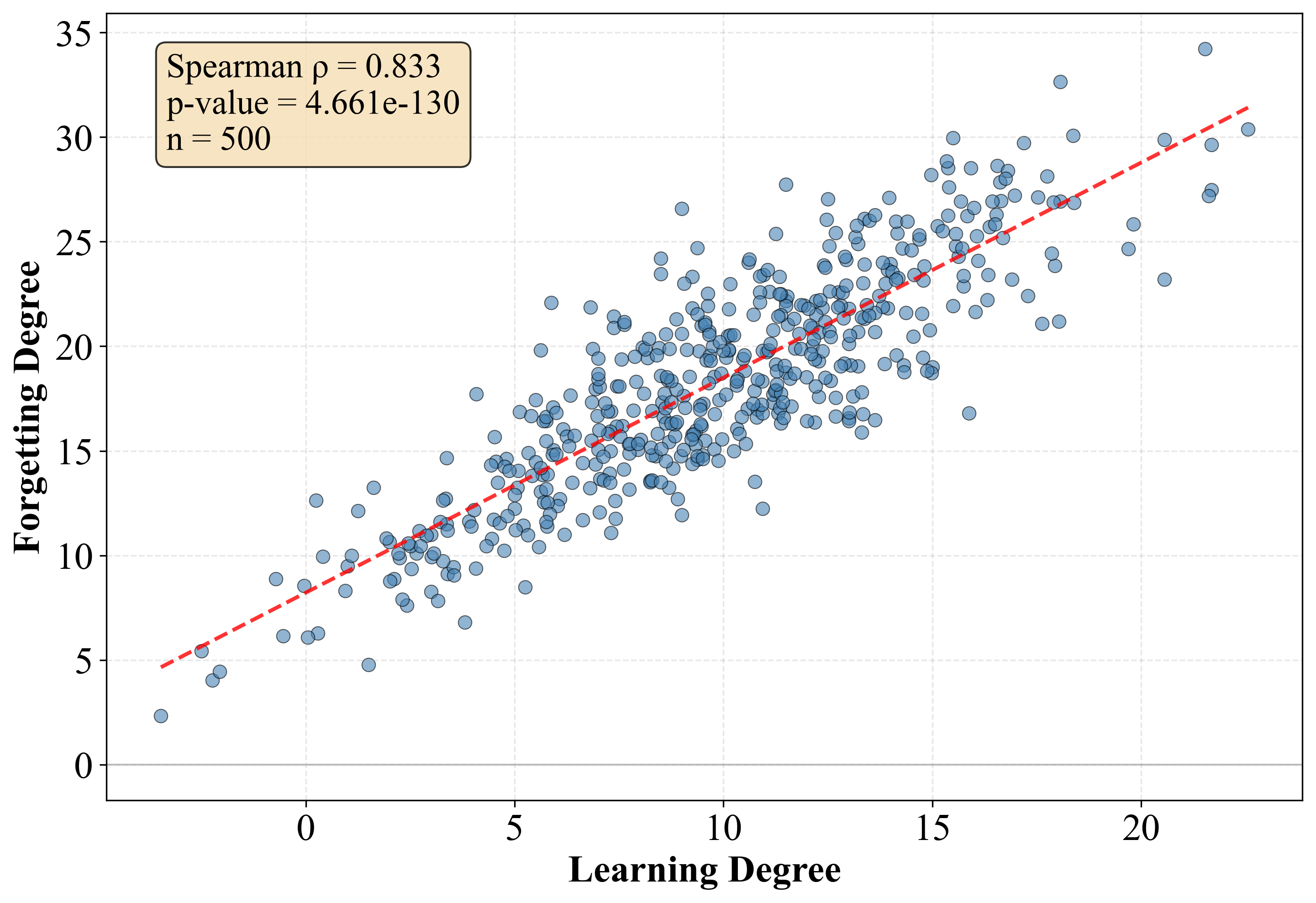}
	\caption{Correlations between learning and forgetting of concepts for Qwen3-4B-Base.}
	\label{fig:concept_corr_qwen}
\end{figure}

To further validate our findings on modern architectures, we replicate the full RQ1--RQ2 protocol on Qwen3-4B-Base, a recent LLM from a third model family. Figure~\ref{fig:concept_learn_hist_qwen} shows the distribution of concept learning and forgetting degree for Qwen3-4B-Base, again revealing substantial variability across concepts. As shown in Table~\ref{tab:correlation}, circuit topology metrics remain correlated with both concept learning and forgetting degree, reproducing the same qualitative trend observed for GPT-2 Large and LLaMA. Finally, Figure~\ref{fig:concept_corr_qwen} shows a positive correlation between learning degree and forgetting degree, confirming that concepts learned more strongly also tend to be forgotten more strongly under subsequent training.

\subsubsection{RQ2 Results}

\begin{figure}[htbp]
	\centering
	\includegraphics[width=1\linewidth]{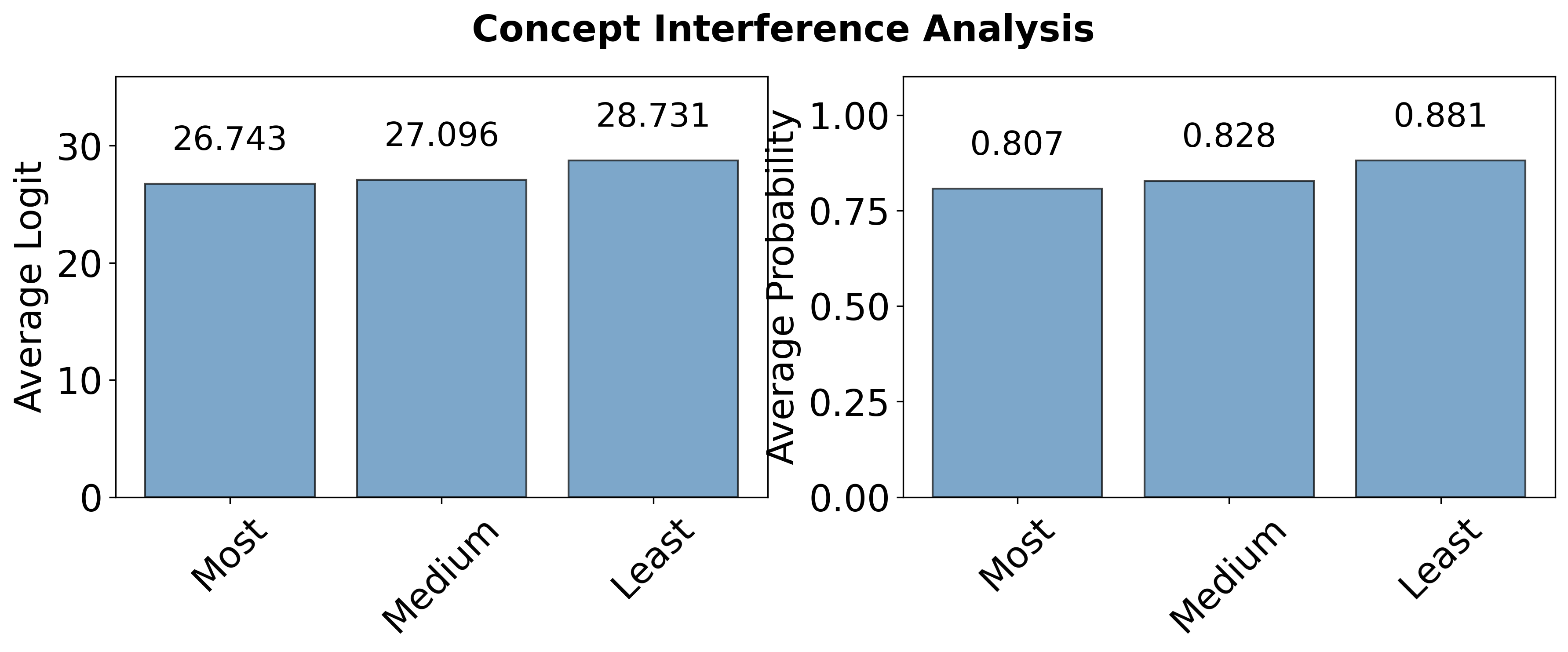}
	\caption{Concept-level interference under joint training on Qwen3-4B-Base.}
	\label{fig:concept_interference_qwen}
\end{figure}

\begin{figure}[htbp]
	\centering
	\includegraphics[width=1\linewidth]{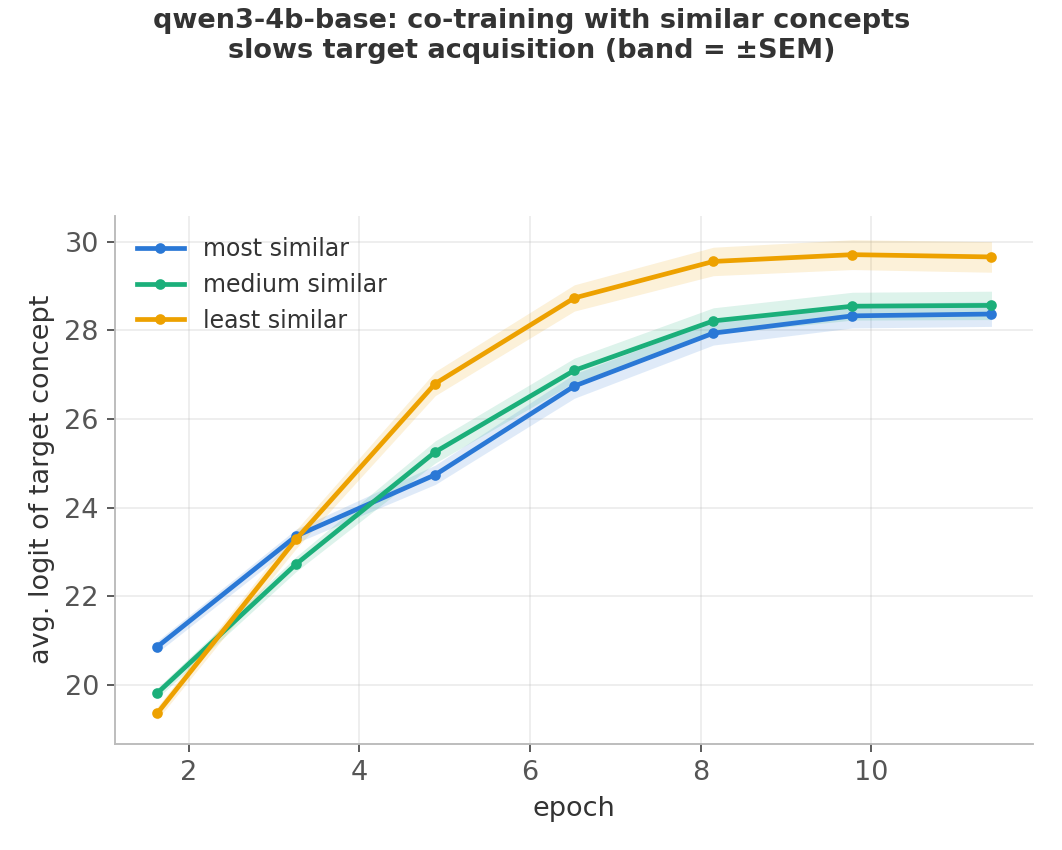}
	\caption{Acquisition curves (average logit of the target concept) for Qwen3-4B-Base under the three co-training conditions (band $=\pm$SEM). The x-axis shows training epochs over the co-training corpus. The interference gap emerges early in acquisition, peaks mid-training, and narrows as all conditions approach the performance ceiling.}
	\label{fig:concept_interference_curves_qwen}
\end{figure}

\begin{figure}[htbp]
	\centering
	\includegraphics[width=1\linewidth]{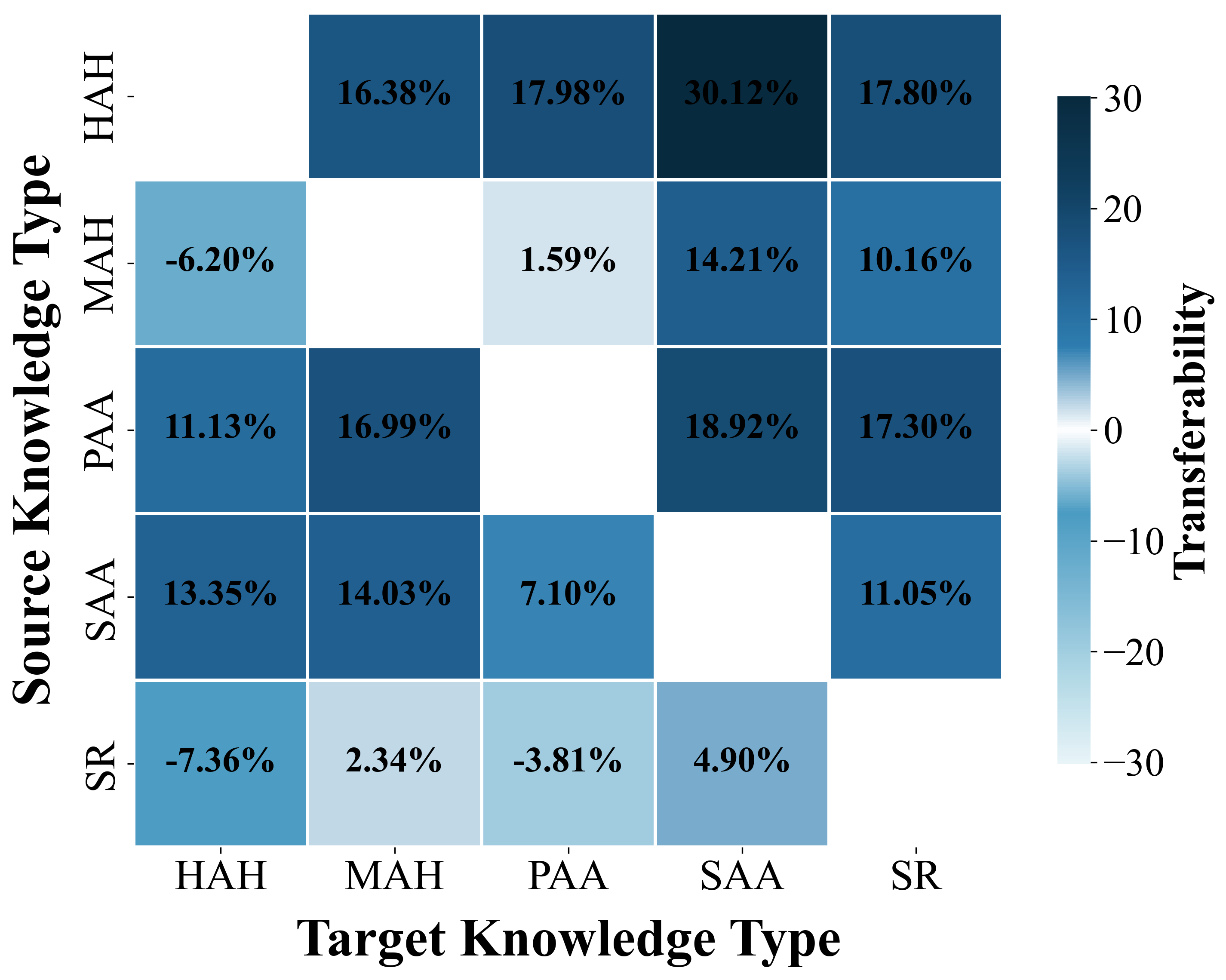}
	\caption{Pairwise dependency between knowledge types on Qwen3-4B-Base.}
	\label{fig:knowledge_dependency_qwen}
\end{figure}

We further run the concept-interference experiment on Qwen3-4B-Base with the same grouping as the main experiments and trace full acquisition curves (Figure~\ref{fig:concept_interference_curves_qwen}). Figure~\ref{fig:concept_interference_qwen} shows the three co-training conditions: training with weakly related concepts achieves the highest performance (average target probability $0.881$, vs.\ $0.828$ for moderately and $0.807$ for highly related groups), reproducing the interference ordering observed on the other models. Figure~\ref{fig:knowledge_dependency_qwen} further shows that transferability across knowledge types remains asymmetric and uneven on Qwen3-4B-Base, indicating that concept-level synergy and interference patterns are robust across model families.

\section{Experiment Details}\label{sec:experiment_setting}
We conduct experiments on $4\times40$GB A40 GPUs. We use learning rate as $5 \times 10^{-5}$ and batch size as $128$. In our experiments for RQ1, the training data is automatically reshuffled before each training epoch. Since each experiment runs for multiple epochs, this ensures that our analysis is independent of the order in which the concepts are presented to the model, addressing any sensitivity to concept presentation order.

\section{GPT-2 Large Results for Log Probability}
\label{sec:log_probability}

To verify that our findings are not specific to logit-based evaluation, we repeat the RQ1 analyses using \textbf{log probability}~\cite{heimersheim2024use,hanna2024have} as the behavioral metric. 

\begin{figure}[htbp]
	\includegraphics[width=1\linewidth]{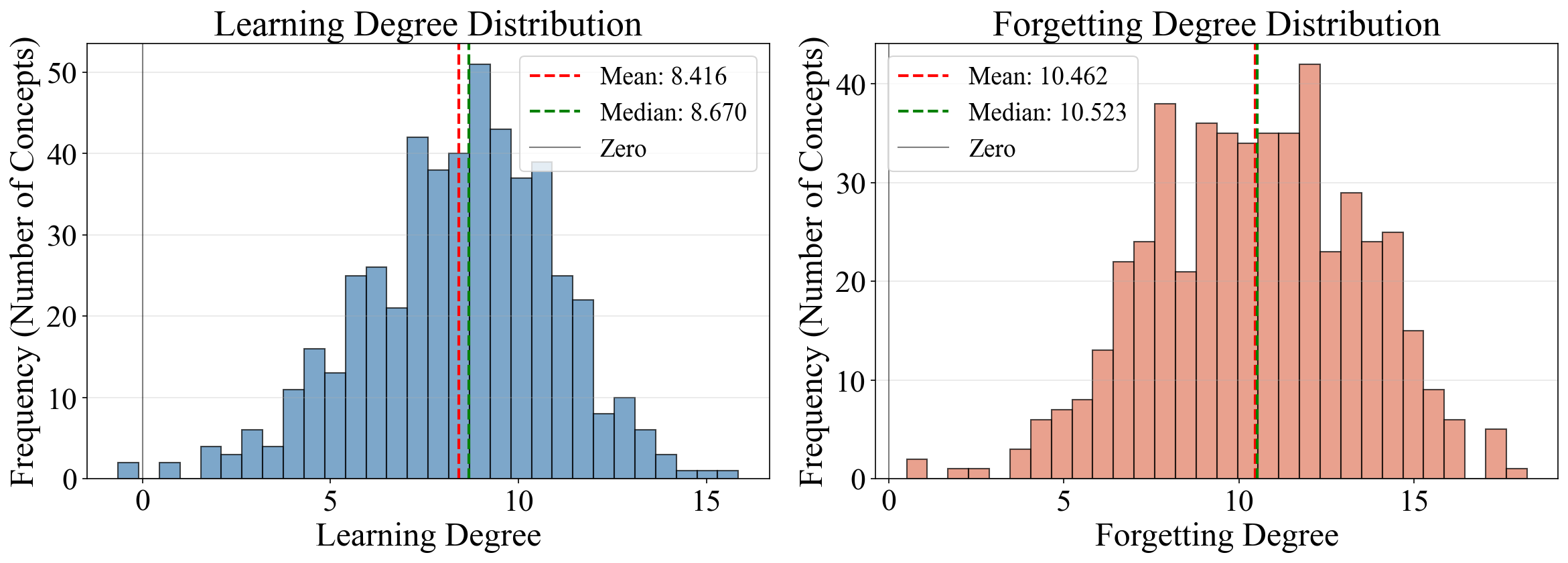}
	\caption{Distribution of learning and forgetting degree across concepts.}
	\label{fig:concept_learn_forget_hist_log_probability}
    \vspace{-1pt}
\end{figure}

Figure~\ref{fig:concept_learn_forget_hist_log_probability} shows the distribution of concept learning and forgetting degree under log probability. Similar to the logit-based results in the main paper, we observe substantial variability across concepts, indicating that both acquisition and forgetting remain highly heterogeneous.

\begin{figure}[t]
    \centering
    \begin{subfigure}[b]{\linewidth}
        \centering
        \includegraphics[width=\linewidth]{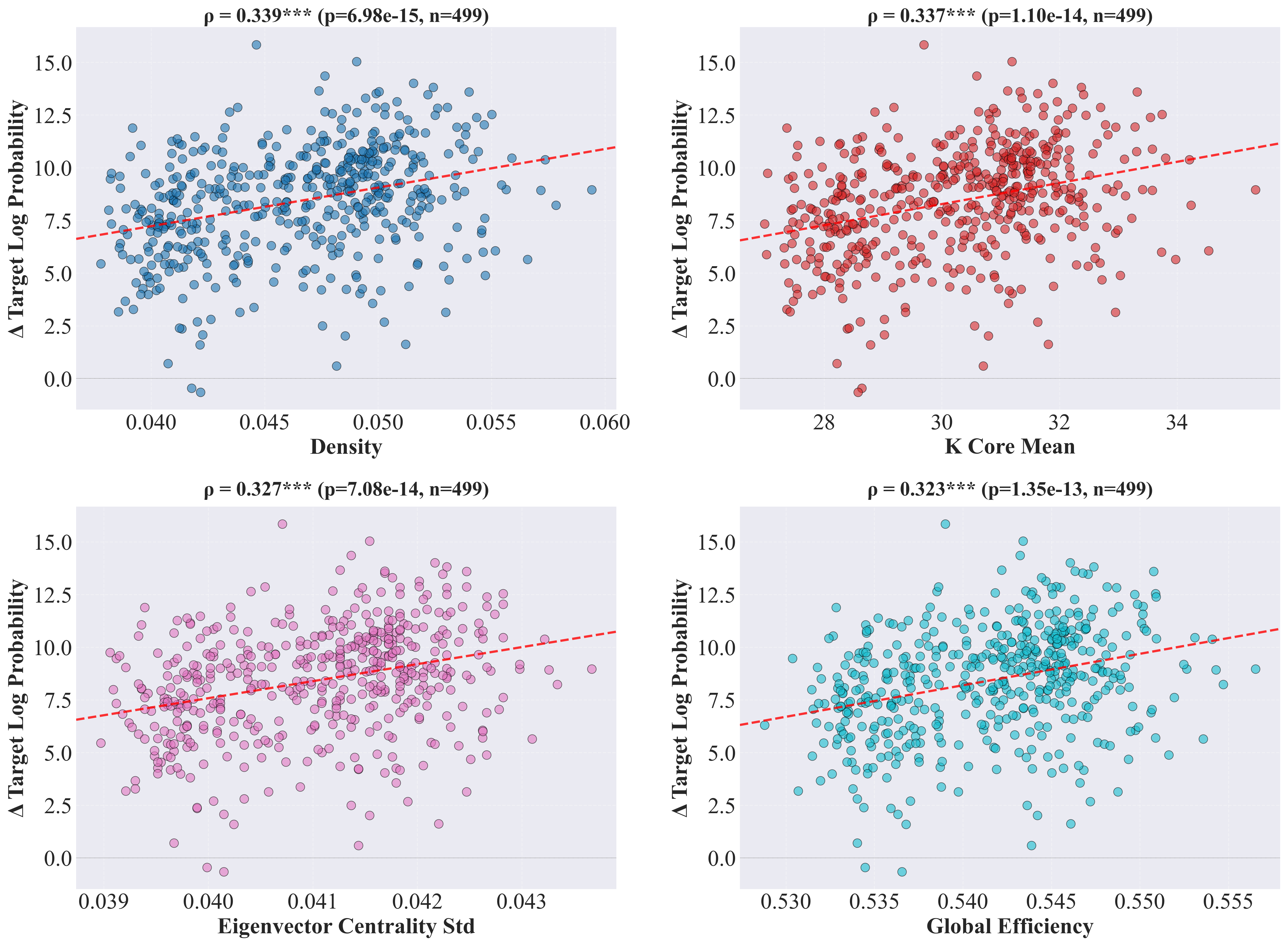}
    \caption{Correlation between learning degree and LLM circuit pattern. }
    \label{fig:learn_correlation_final_merge_log_probability}
    \end{subfigure}
    \hfill
    \begin{subfigure}[b]{\linewidth}
        \centering
        \includegraphics[width=1\linewidth]{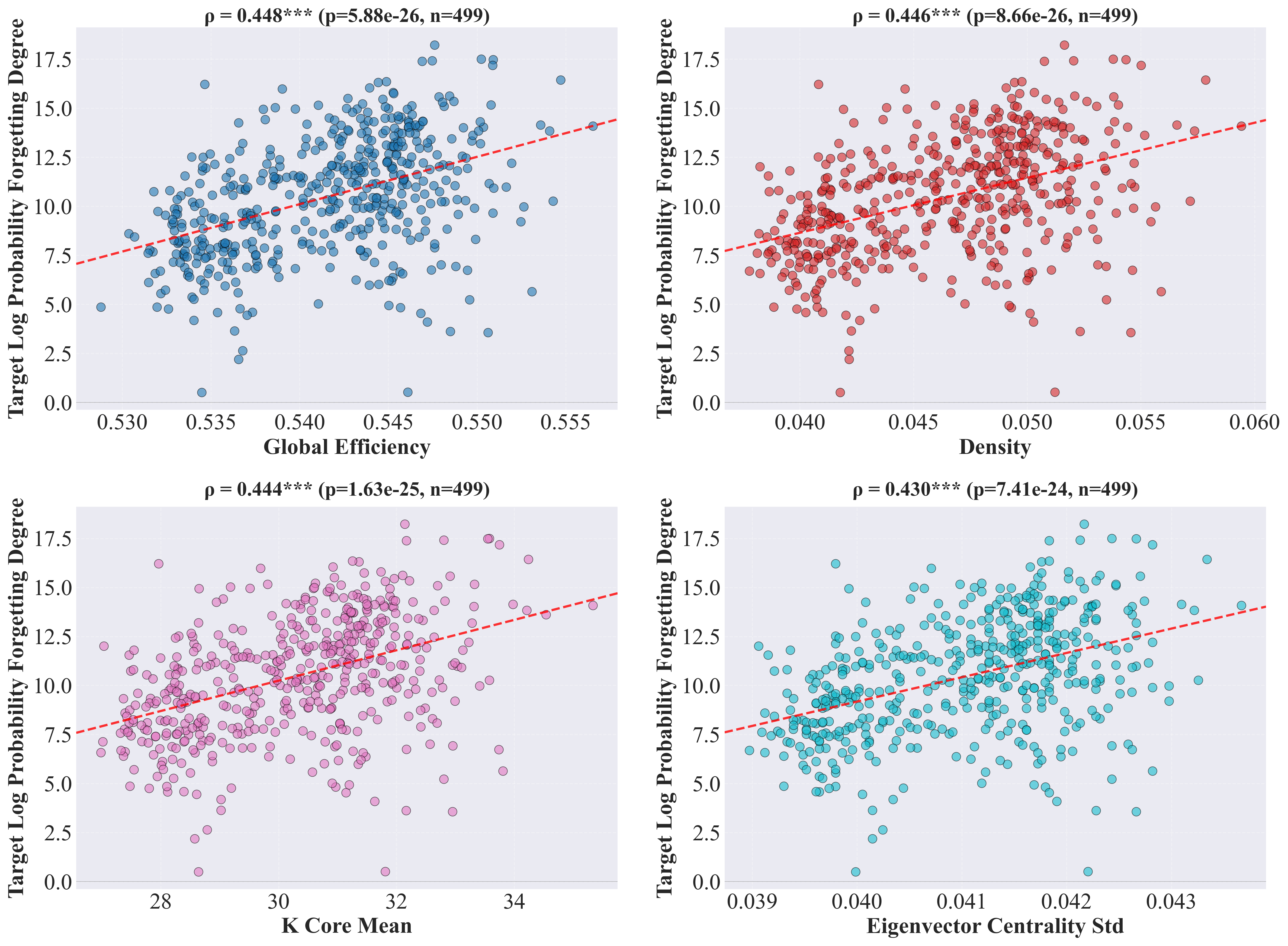}
        \caption{Correlation between forgetting degree and LLM circuit pattern.}
    	\label{fig:forget_correlation_merge_log_probability}
    \end{subfigure}

    \caption{Correlation between learning dynamics and LLM circuit pattern.}

\end{figure}

Figure~\ref{fig:learn_correlation_final_merge_log_probability} and Figure~\ref{fig:forget_correlation_merge_log_probability} show that circuit topology metrics remain positively correlated with concept learning and forgetting degree, respectively. In particular, metrics such as eigenvector centrality and $k$-core continue to exhibit consistent associations, suggesting that the relationship between circuit structure and learning/forgetting dynamics is robust to the choice of behavioral metric.

\begin{figure}[htbp]
	\centering
	\includegraphics[width=1\linewidth]{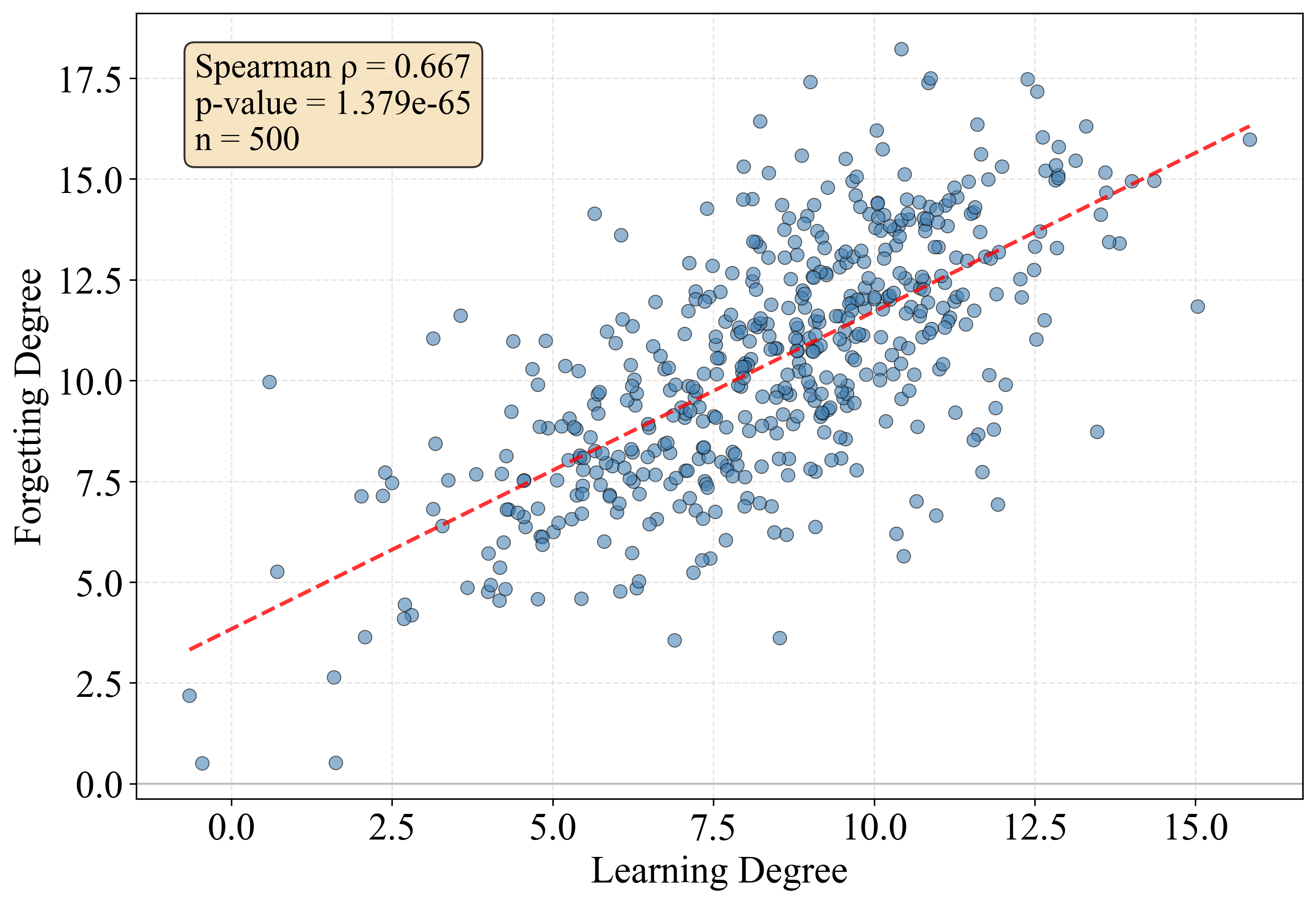}
	\caption{Spearman Correlations between learning and forgetting of concepts}
	\label{fig:concept_corr_log}
\end{figure}

Finally, Figure~\ref{fig:concept_corr_log} shows that concepts with larger learning gains also tend to exhibit greater forgetting under subsequent training, confirming the learn-forget trade-off observed in the main paper. Overall, these results demonstrate that our key findings remain consistent when measured using log probability.


\begin{table}[t]
\centering
\resizebox{\linewidth}{!}{
\setlength{\tabcolsep}{6pt}
\renewcommand{\arraystretch}{0.92}
\begin{tabular}{l l c c c c}
\toprule
Type & Metric & Learning & Learning $p$ & Forgetting & Forgetting $p$  \\
\midrule
Circuit Topology (70\%) & Density & 0.4474 & 6.16e-26 & 0.4585 &2.64e-27 \\
Circuit Topology (70\%) & $k$-core & 0.4468 & 7.45e-26 & 0.4587 &  2.46e-27\\
Circuit Topology (70\%) & Eigenvector Centrality & 0.4403 &4.40e-25 &  0.4459 & 9.60e-26 \\
Circuit Topology (70\%) & Global Efficiency & 0.4085 &1.73e-21 &  0.4322 &3.95e-24  \\
\midrule
Circuit Topology (80\%) & Density &  0.4723 & 4.34e-29 &  0.4785  &6.54e-30 \\
Circuit Topology (80\%) & $k$-core &  0.4475 &  6.04e-26&  0.4610 &1.25e-27 \\
Circuit Topology (80\%) & Eigenvector Centrality & 0.3235  &1.27e-13 & 0.3314  &2.95e-14  \\
Circuit Topology (80\%) & Global Efficiency & 0.4662 &2.71e-28 &  0.4747 & 2.09e-29  \\
\midrule
Activation & Mean EAP-IG & 0.2680 & 1.18e-09
           & 0.2784 & 2.47e-10 \\
Activation & STD EAP-IG  & 0.2398 & 5.87e-08
           & 0.2494 & 1.64e-08 \\
Activation & Max EAP-IG  & 0.1575 & 4.13e-04
           & 0.1342 & 2.67e-03 \\

\midrule

Random Topology & Density & -0.0078 &  0.86  & -0.0255 & 0.57 \\
Random Topology & $k$-core & -0.0458 & 0.31 & -0.0437 &0.33  \\
Random Topology & Eigenvector Centrality & 0.0473 & 0.29 & 0.0737 & 0.10 \\
Random Topology & Global Efficiency & -0.0559 & 0.21 & -0.0769  &0.09 \\
\bottomrule
\end{tabular}
}
\caption{Correlation between learning/forgetting degree and circuit metrics (Spearman $\rho$);  $p$ is the corresponding $p$-value testing the null hypothesis of zero correlation.}
\label{tab:baseline_complete}
\vspace{-0.8em}
\end{table}

\section{Complete Results}~\label{sec:complete_result}
Table ~\ref{tab:baseline_complete} shows complete correlation results with $p$-values, where $p$ is the corresponding $p$-value testing the null hypothesis of zero correlation.

\section{Baselines Against Circuit Topology}\label{sec:baseline}
To further evaluate the indicative power of circuit graph topology, we compare our method against two baselines:
(1) \textbf{Random graph topology}, where we randomly sample the same number of edges to form a graph and compute its topology metrics; and
(2) \textbf{edge-level EAP-IG statistics}~\cite{hanna2024have}, including the mean, maximum, and standard deviation of EAP-IG scores. EAP-IG is a first-order approximation of each edge's indirect effect and serves as a proxy for both gradient norm and activation magnitude at the circuit level.

As shown in Table~\ref{tab:baseline}, circuit graph metrics exhibit strong correlations with concept learning and forgetting degree (Spearman $\rho \approx 0.41$--$0.47$ and $0.43$--$0.46$, respectively, at the 70\% threshold).


\section{Correlation between Learning and Forgetting}\label{sec:learn_forget_correlation}

\begin{figure}[htbp]
	\centering
	\includegraphics[width=1\linewidth]{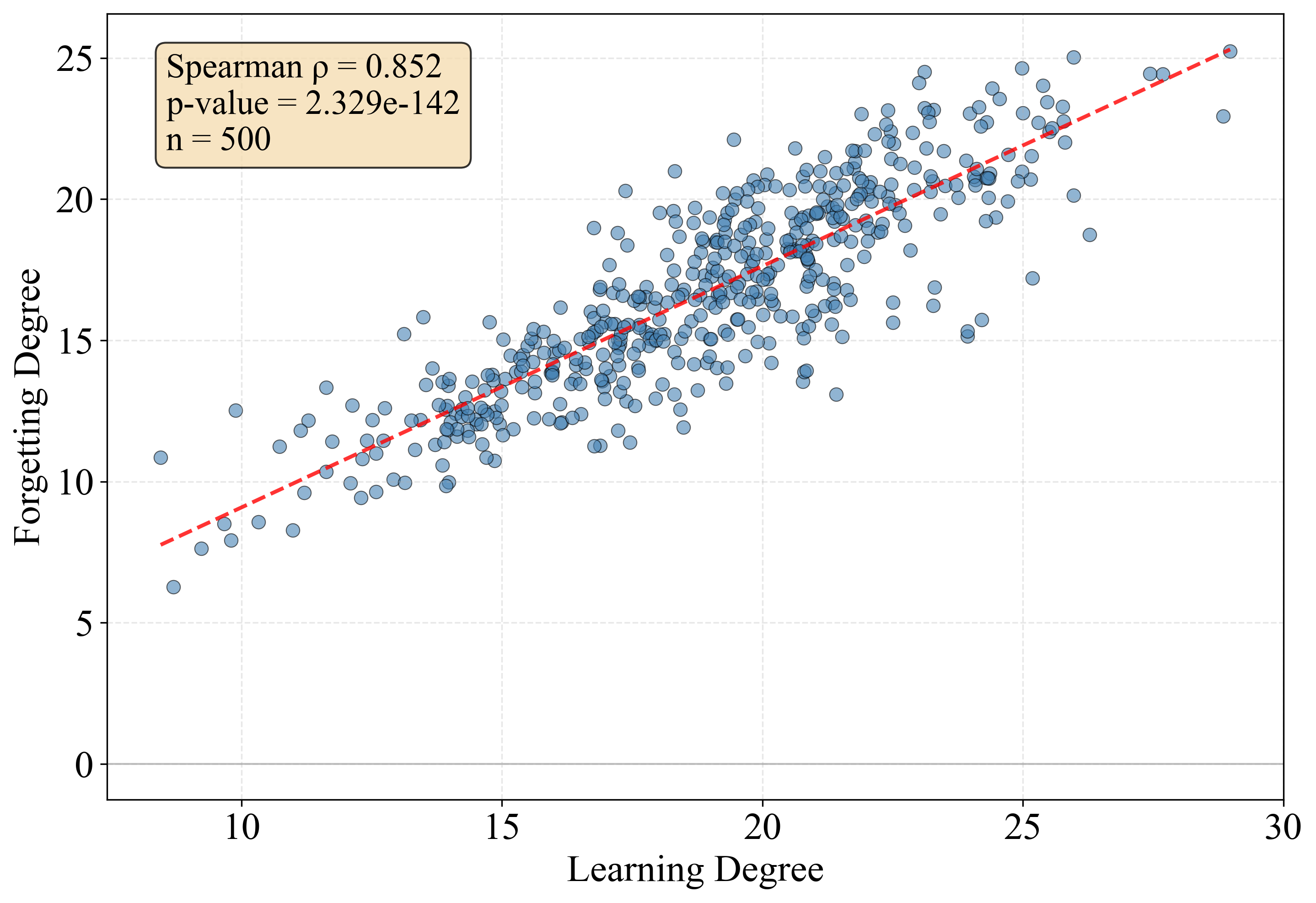}
	\caption{Spearman Correlations between learning and forgetting of concepts}
	\label{fig:concept_corr}
\end{figure}

\noindent\textit{Concepts with larger learning degree tend to exhibit larger forgetting degree during subsequent training, suggesting a trade-off between acquisition strength and subsequent stability.}
\paragraph{Correlation between Concept Learning and Forgetting.} Figure~\ref{fig:concept_corr} shows a positive correlation between learning degree and forgetting degree under continual training, indicating that concepts learned more strongly also tend to be forgotten more strongly. Together with the circuit correlation results in Section~\ref{finding:finding_circuit_allocation}, this pattern is consistent with a trade-off between acquisition strength and subsequent stability.

\begin{figure}[htbp]
	\centering
	\includegraphics[width=1\linewidth]{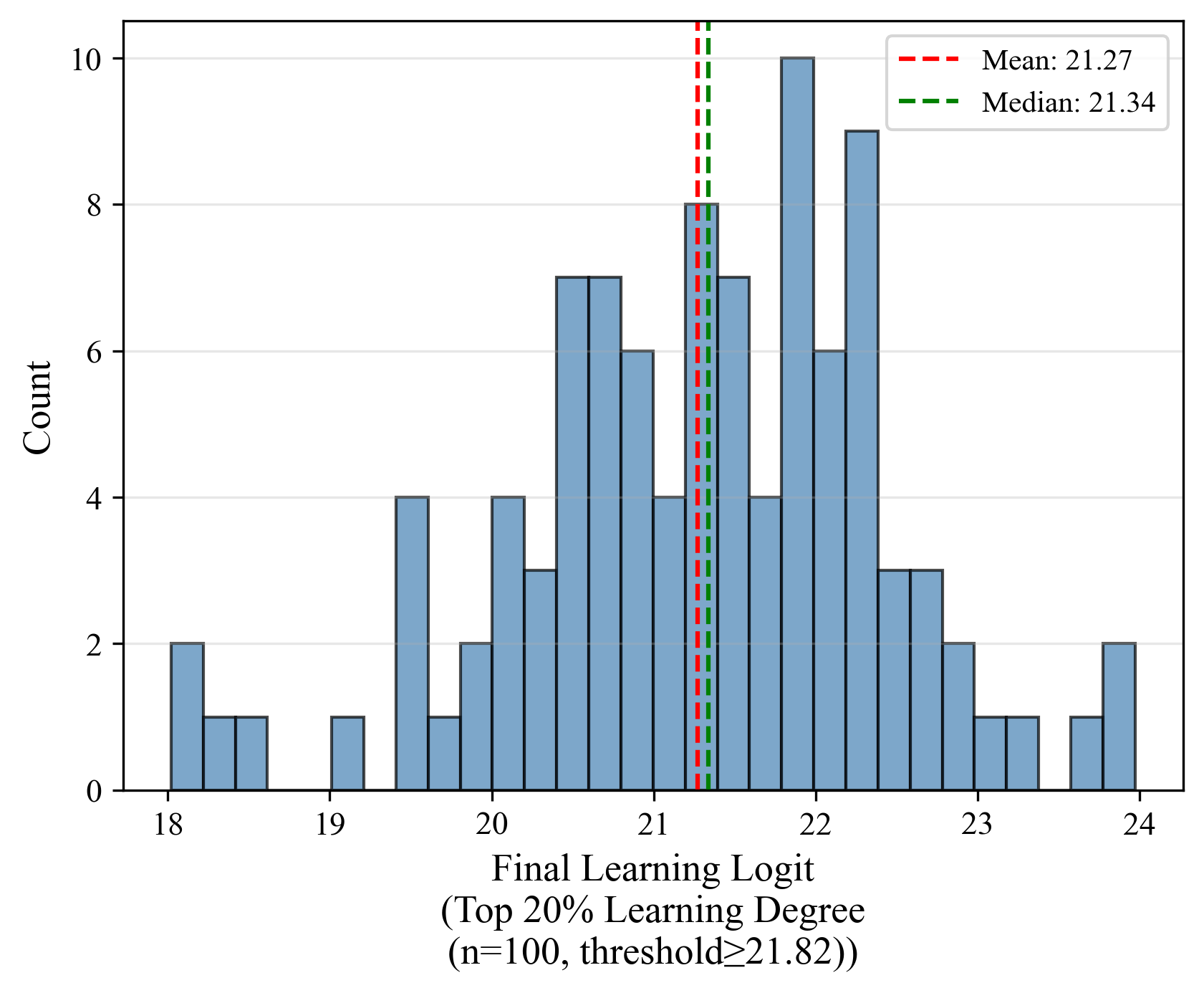}
	\caption{Distribution of final learning logit.}
	\label{fig:concept_corr_logit}
\end{figure}

\begin{figure}[htbp]
	\centering
	\includegraphics[width=1\linewidth]{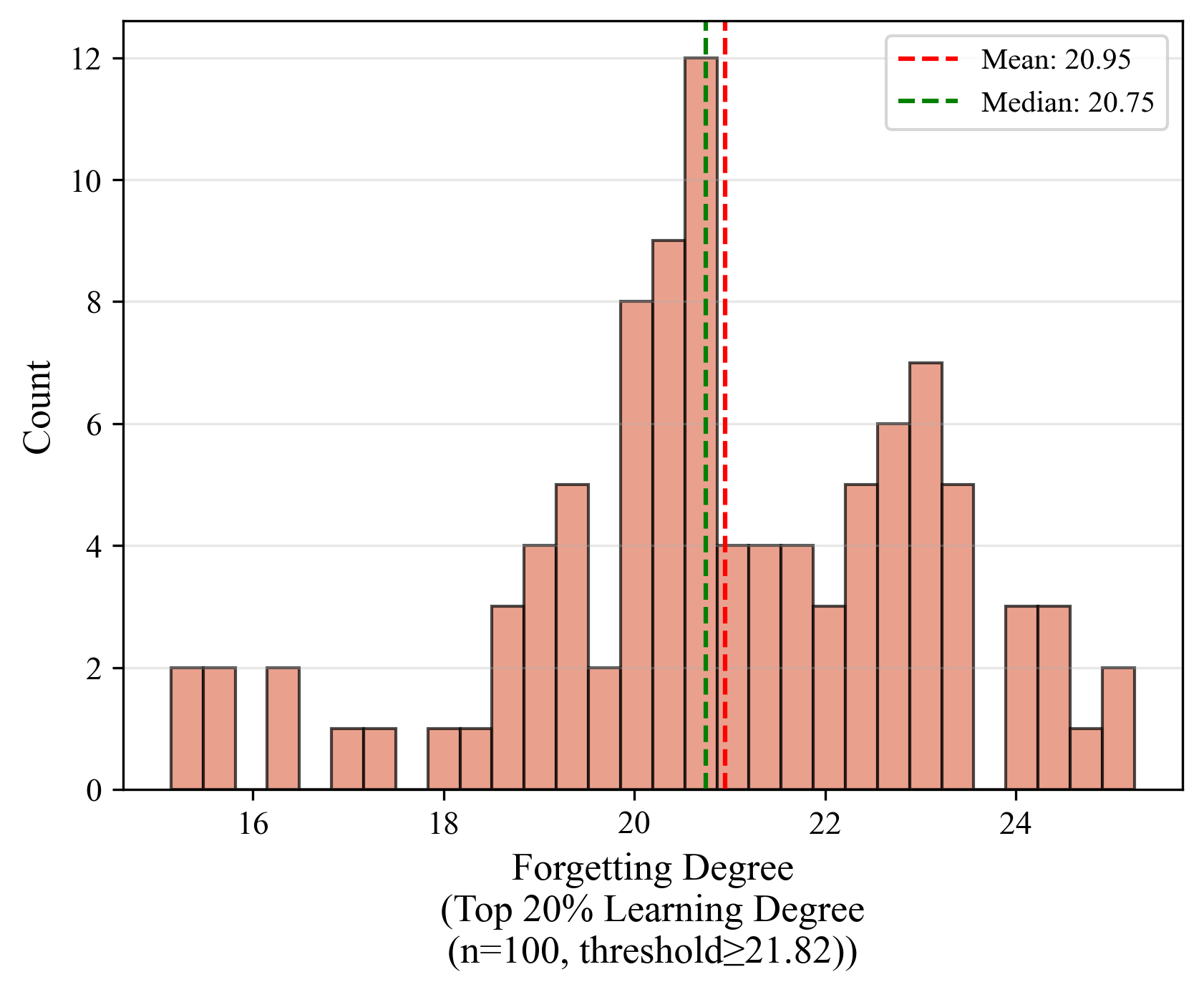}
	\caption{Distribution of forgetting degree}
	\label{fig:concept_corr_forget}
\end{figure}

To shed light on the nature of the learning--forgetting relationship, we further analyze the 20\% of concepts with the highest learning degrees. Figure~\ref{fig:concept_corr_logit} and Figure~\ref{fig:concept_corr_forget} show the distributions of their final learning logit and forgetting degree. Two observations are notable. First, while these high-learning-degree concepts do tend to reach elevated final logit values (mean = 21.27), their final logits span a wide range (18--24), indicating that high learning degree does not simply equate to a uniformly high logit ceiling. Second, and more importantly, these concepts also exhibit substantially higher forgetting degrees (mean = 20.95) but span a wide range (14--26). Together, these distributions suggest that the correlation between learning degree and forgetting degree reflects a genuine property of how concepts are encoded during training.




\section{Evolution of All Circuit Topology Metrics}\label{sec:metric_evolution}

\begin{figure}[htbp]
	\centering
	\includegraphics[width=1\linewidth]{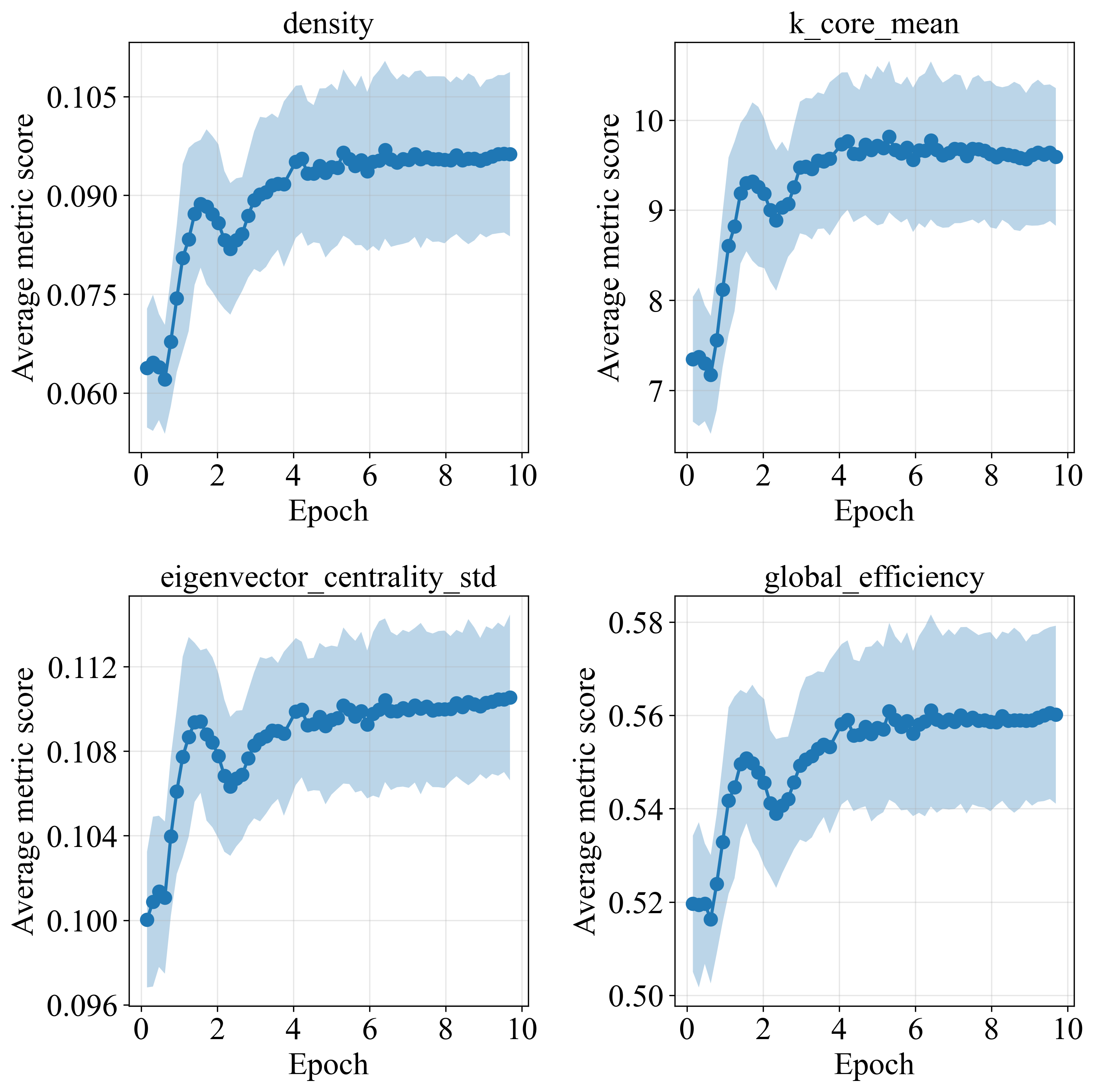}
	\caption{Evolution of all four circuit graph metrics during the learning stage (mean over concepts). The x-axis shows training epochs; the full learning stage spans 10 epochs.}
	\label{fig:rq1_learn_over_steps_all_metrics}
\end{figure}

\begin{figure}[htbp]
	\centering
	\includegraphics[width=1\linewidth]{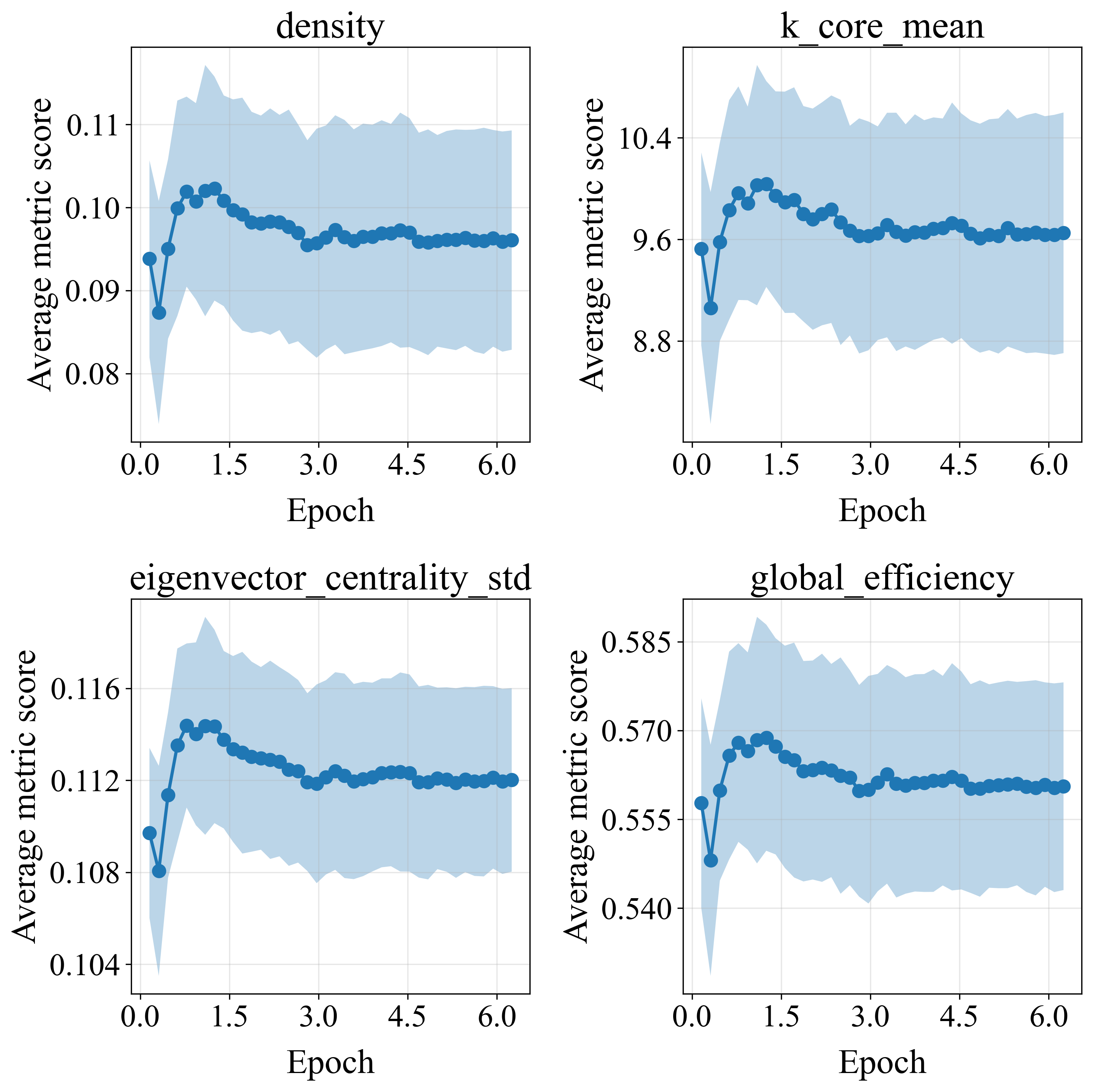}
	\caption{Evolution of all four circuit graph metrics during the forgetting stage (mean over concepts).}
	\label{fig:rq2_forget_over_steps_all_metrics}
\end{figure}

Figure~\ref{fig:rq1_learn_over_steps_all_metrics} and Figure~\ref{fig:rq2_forget_over_steps_all_metrics} show the temporal evolution of all four circuit topology metrics (density, mean $k$-core, eigenvector centrality, and global efficiency) during the learning and forgetting stages, complementing the two representative metrics in the main paper. All four metrics follow the same qualitative pattern within each stage.

\section{Controlled Interference Experiment: Relatedness vs.\ Surface Overlap}\label{sec:controlled_interference}

In the interference experiment of Section~\ref{sec:concept_interference_exp}, highly related co-training groups may also share surface forms (object strings, relation types) with the target, so the observed gap could in principle reflect surface-level overlap rather than representational competition. We quantify the covariates of the original design and find that relation-type composition is essentially constant across conditions (mean relation-histogram Jaccard 0.178--0.185, $p=0.20$), while object-string overlap (19.3\% coverage for the highly related group vs.\ 0.3\% for the weakly related group) do differ.

We therefore run a controlled experiment that removes these channels. For each of the same 100 target concepts, we construct three co-training conditions of exactly $K=75$ auxiliary concepts: high relatedness with natural overlap retained (\textit{hi-ovl}), high relatedness with auxiliary object strings \emph{and} object content-word tokens fully disjoint from the target's (\textit{hi-noovl}), and low relatedness with the same disjointness (\textit{lo-noovl}). Auxiliary texts are subsampled and upsampled so that all three conditions have identical relation histograms, while preserving the relatedness contrast. Training follows the original protocol (GPT-2 Large).

\begin{table}[htbp]
\centering
\small
\resizebox{\linewidth}{!}{
\begin{tabular}{l c c}
\toprule
Contrast (avg.\ prob.) & $\Delta$ & $p$ \\
\midrule
lo-noovl $-$ hi-noovl & $+11.6$\,pp & $6.1\times10^{-7}$ \\
hi-noovl $-$ hi-ovl & $-2.2$\,pp & $0.015$ \\
lo-noovl $-$ hi-ovl & $+9.4$\,pp & $6.5\times10^{-6}$ \\
\bottomrule
\end{tabular}
}
\caption{Controlled interference contrasts. The first row isolates semantic relatedness at exactly zero lexical overlap; the second isolates the effect of shared object strings; the third is the protocol-matched total.}
\label{tab:controlled_interference}
\end{table}

Table~\ref{tab:controlled_interference} shows the paired contrasts. The key comparison is the first row, \textit{lo-noovl} vs.\ \textit{hi-noovl}: both conditions share zero object strings with the target, so they differ only in semantic relatedness. Even under this condition, targets co-trained with highly related but no-object overlap concepts, \textit{hi-noovl}, reach an average target probability $11.6$\,pp lower than targets co-trained with weakly related concepts, \textit{lo-noovl}, (logit contrast $+0.81$, $p=7.6\times10^{-6}$; 79\% of targets show the effect). This pure-relatedness effect accounts for roughly 60\% of the $+18.5$\,pp least-vs-most gap in the original, uncontrolled design, indicating that the majority of the observed interference is attributable to semantic relatedness itself rather than to surface-level overlap. Removing shared objects slightly \emph{hurts} performance ($-2.2$\,pp), indicating that object overlap by itself slightly helps learning rather than causing the interference. Overall, the interference in Finding~4 is genuinely representational: it survives the removal of all object-string overlap.

\section{Computational Overhead of Circuit Scoring}\label{sec:overhead}

\begin{table}[htbp]
\centering
\small
\resizebox{\linewidth}{!}{
\begin{tabular}{l c c}
\toprule
Scoring method & Single GPU & 10-way parallel \\
\midrule
Gradient scoring (per-sample gradient norm) & 11.1 min & -- \\
Circuit extraction (EAP-IG) & 83.4 min & $\approx$8 min \\
\bottomrule
\end{tabular}
}
\caption{One-time scoring cost for replay-concept selection, measured on GPT-2 Large with all 7,140 AntiLeak concepts.}
\label{tab:overhead}
\end{table}

We measure one-time scoring costs on GPT-2 Large with all 7,140 AntiLeak concepts (Table~\ref{tab:overhead}). Gradient scoring (one backward pass per sample on a single GPU) takes 11.1 minutes. EAP-IG circuit extraction requires only a fixed number of forward/backward passes per concept and takes 83.4 minutes on a single GPU; since extraction is fully parallelizable across concepts, this reduces to $\approx$8.3 minutes wall-clock with 10-way data parallelism in our experiments. Notably, our replay strategy does not require extraction at every checkpoint: circuits are extracted only once, at the final checkpoint of the learning stage. The scoring cost is therefore a fixed one-time overhead, while the retention gains it enables persist for all subsequent deployment, yielding a favorable cost-benefit trade-off.

\begin{table}[t]
\centering
\small
\begin{tabular}{l c c}
\toprule
Quintile (dispersion) & $n$ & Answers lost (\%) \\
\midrule
Q1  & 173 & 67.6 \\
Q2 & 172 & 61.9 \\
Q3 & 172 & 58.4 \\
Q4 & 172 & 53.2 \\
Q5 & 173 & 50.9 \\
\bottomrule
\end{tabular}
\caption{Downstream answer loss by quintile of pre-forgetting eigenvector-centrality dispersion (an answer is lost if it is correct before but incorrect after the forgetting stage).}
\label{tab:mcqa_quintiles}
\end{table}

\section{Metrics for AntiLeak Experiments}\label{sec:metrics}
We report three evaluation metrics on the AntiLeak test set: (1) \textbf{Logit}, the average logit assigned to the target object; (2) \textbf{Accuracy}, the proportion of examples for which the top-1 predicted token matches the ground-truth object; and (3) \textbf{Hit@5}, the proportion of examples for which the ground-truth object appears in the top-5 predicted tokens.

\section{Downstream MCQA Evaluation on AntiLeak}\label{sec:downstream_mcqa}

To verify that the forgetting tracked by our logit-based metrics translates into downstream task degradation, we build a closed-book multiple-choice QA evaluation~\cite{yao-etal-2025-error} from AntiLeakBench's official multi-choice questions~\cite{wu2025antileakbench}: 994 single-hop questions whose gold answers match facts in our AntiLeak training set. Since GPT-2 Large is a base model, we use the standard likelihood-based protocol: each of the four options is scored by its log-likelihood as a continuation of the question, and the model's answer is the highest-scoring option.

MCQA accuracy is 92.2\% after concept learning and drops to 40.0\% after the forgetting stage. Circuit topology computed at the \emph{pre-forgetting} checkpoint predicts this degradation at the level of individual concepts: taking the concepts the model answered correctly before the forgetting stage, ranking them by the eigenvector-centrality score of their pre-forgetting circuits, and splitting them into five equal-sized quintiles, the fraction of answers lost (answered correctly before but incorrectly after the forgetting stage) decreases monotonically across the quintiles, from 67.6\% to 50.9\%, as shown in Table~\ref{tab:mcqa_quintiles}.

 \section{Use Of AI Assistants}
We used AI assistants to help with writing and editing the manuscript. The assistants were used to refine wording, improve clarity, and enhance overall presentation.

\end{document}